\documentclass{article}

\usepackage{microtype}
\usepackage{graphicx}
\usepackage{subfigure}
\usepackage{booktabs} 

\usepackage{hyperref}

\usepackage[accepted]{icml2025}
\usepackage{amsmath}
\usepackage{amssymb}
\usepackage{mathtools}
\usepackage{amsthm}

\usepackage{multirow}

\def\e{\varepsilon{}}
\def\N{\mathcal{N}}
\def\E{\mathbb{E}}
\def\cost{\mathcal{L}}

\definecolor{ourgreen}{RGB}{20, 165, 105}

\def\our{$\beta$-CFG}

\usepackage[capitalize,noabbrev]{cleveref}

\theoremstyle{plain}

\theoremstyle{definition}

\theoremstyle{remark}

\usepackage[textsize=tiny]{todonotes}

\icmltitlerunning{Classifier-free Guidance with Adaptive Scaling}

\begin{document}

\twocolumn[
\icmltitle{Classifier-free Guidance with Adaptive Scaling}



\icmlsetsymbol{equal}{*}

\begin{icmlauthorlist}
\icmlauthor{Dawid Malarz}{equal,yyy}
\icmlauthor{Artur Kasymov}{equal,yyy}
\icmlauthor{Maciej Zieba}{comp}
\icmlauthor{ Jacek Tabor}{yyy}
\icmlauthor{Przemys\l{}aw Spurek}{yyy}
\end{icmlauthorlist}

\icmlaffiliation{yyy}{Jagiellonian University}
\icmlaffiliation{comp}{ University of Science and Technology Wrocław}

\icmlcorrespondingauthor{Przemys\l{}aw Spurek}{przemyslaw.spurek@uj.edu.pl}

\icmlkeywords{Machine Learning, ICML}

\vskip 0.3in
]



\printAffiliationsAndNotice{\icmlEqualContribution} 

\begin{abstract}
Classifier-free guidance (CFG) is an essential mechanism in contemporary text-driven diffusion models. In practice, in controlling the impact of guidance we can see the trade-off between the quality of the generated images and correspondence to the prompt. When we use strong guidance, generated images fit the conditioned text perfectly but at the cost of their quality. Dually, we can use small guidance to generate high-quality results, but the generated images do not suit our prompt. 
In this paper, we present \our{} ($\beta$-adaptive scaling in Classifier-Free Guidance), which controls the impact of guidance during generation to solve the above trade-off. First, \our{} stabilizes the effects of guiding by gradient-based adaptive normalization. Second, \our{} uses the family of single-modal ($\beta$-distribution), time-dependent curves to dynamically adapt the trade-off between prompt matching and the quality of samples during the diffusion denoising process. Our model obtained better FID scores, maintaining the text-to-image CLIP similarity scores at a level similar to that of the reference CFG.
\end{abstract}

\begin{figure}[h]
    \centering

    \begin{tabular}{cc}
        \includegraphics[width=0.43\linewidth]{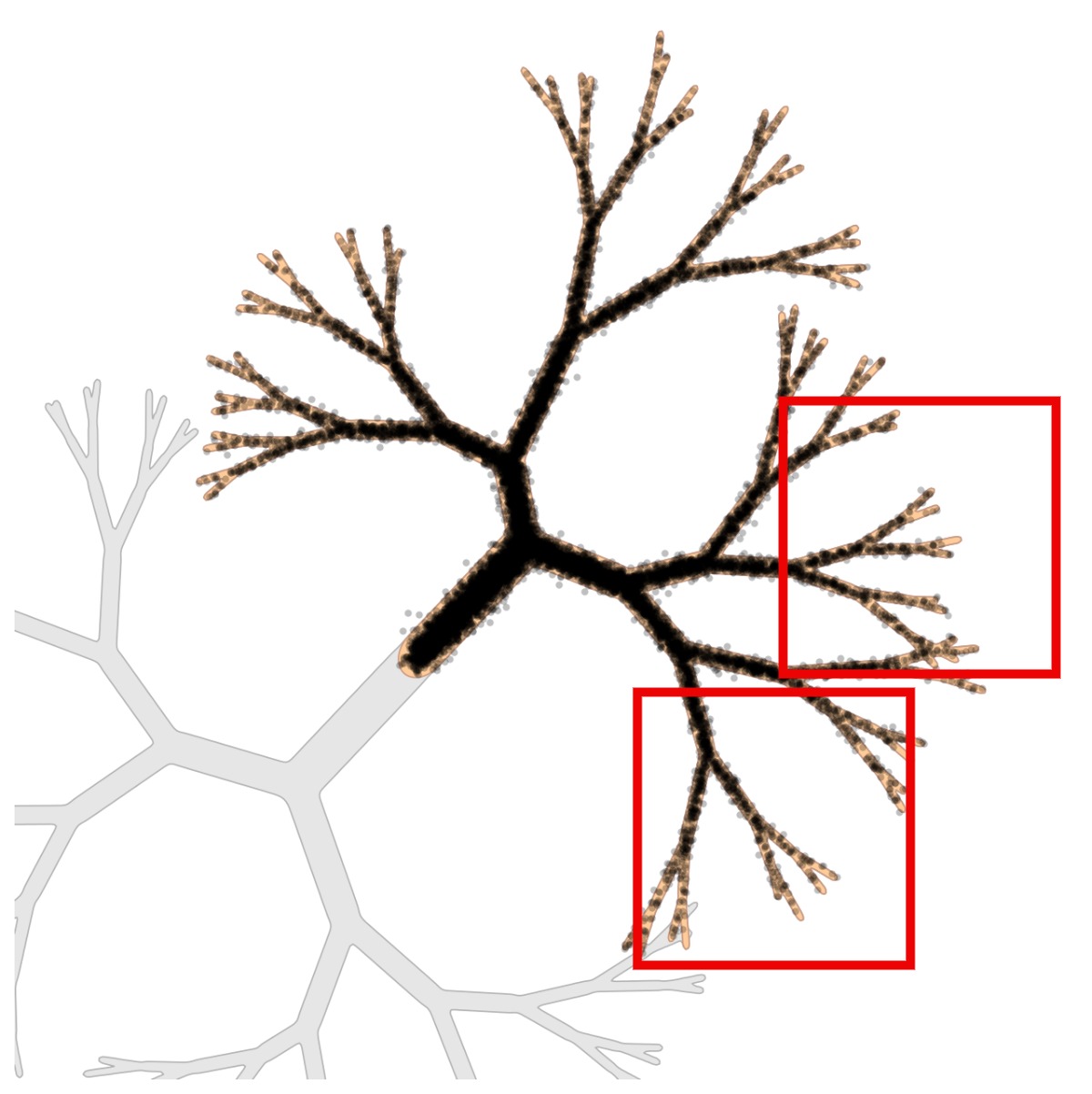} & 
        \includegraphics[width=0.43\linewidth]{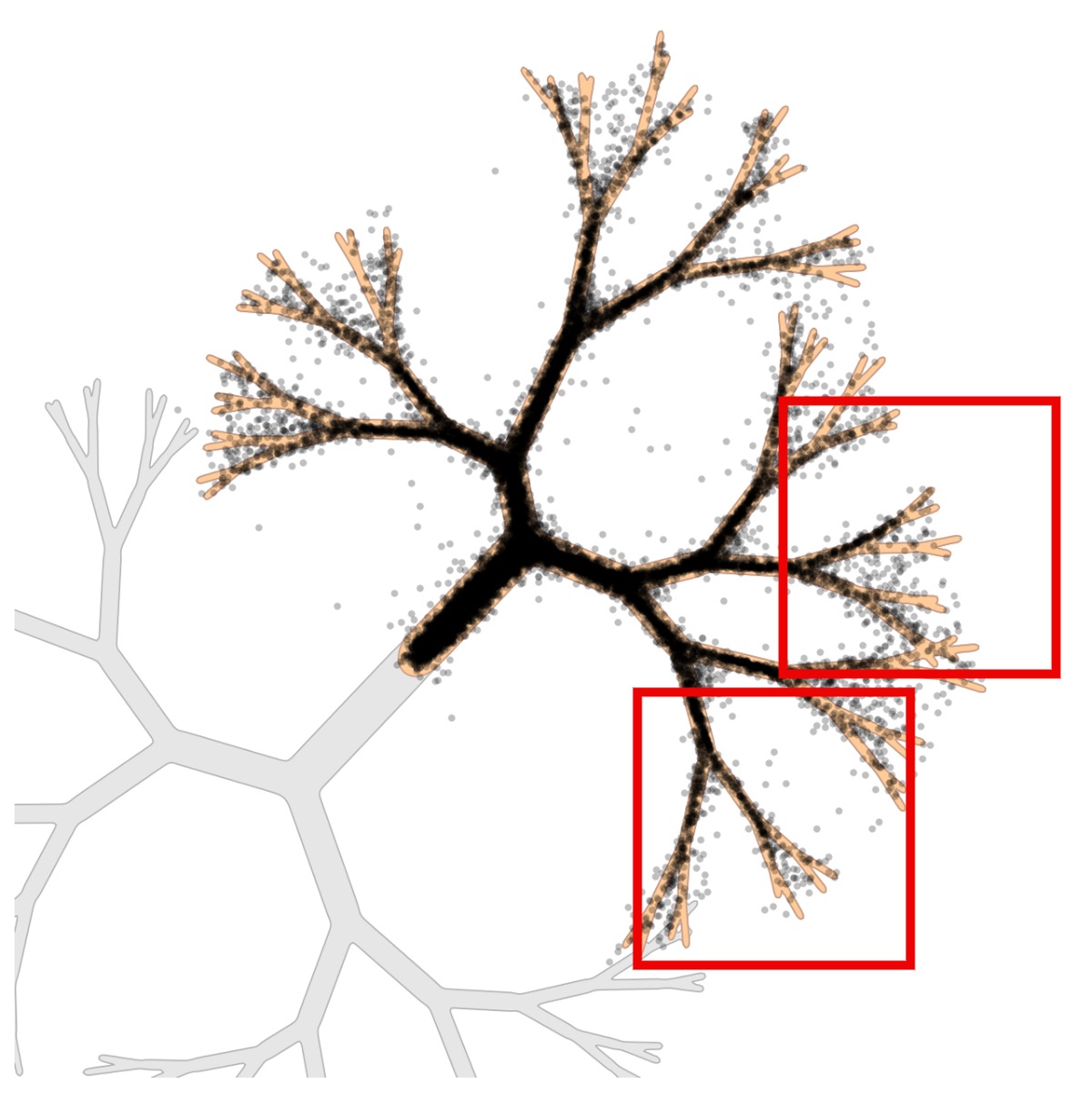} \\ 
        (a) Ground truth & (b) No guidance \\
        \includegraphics[width=0.43\linewidth]{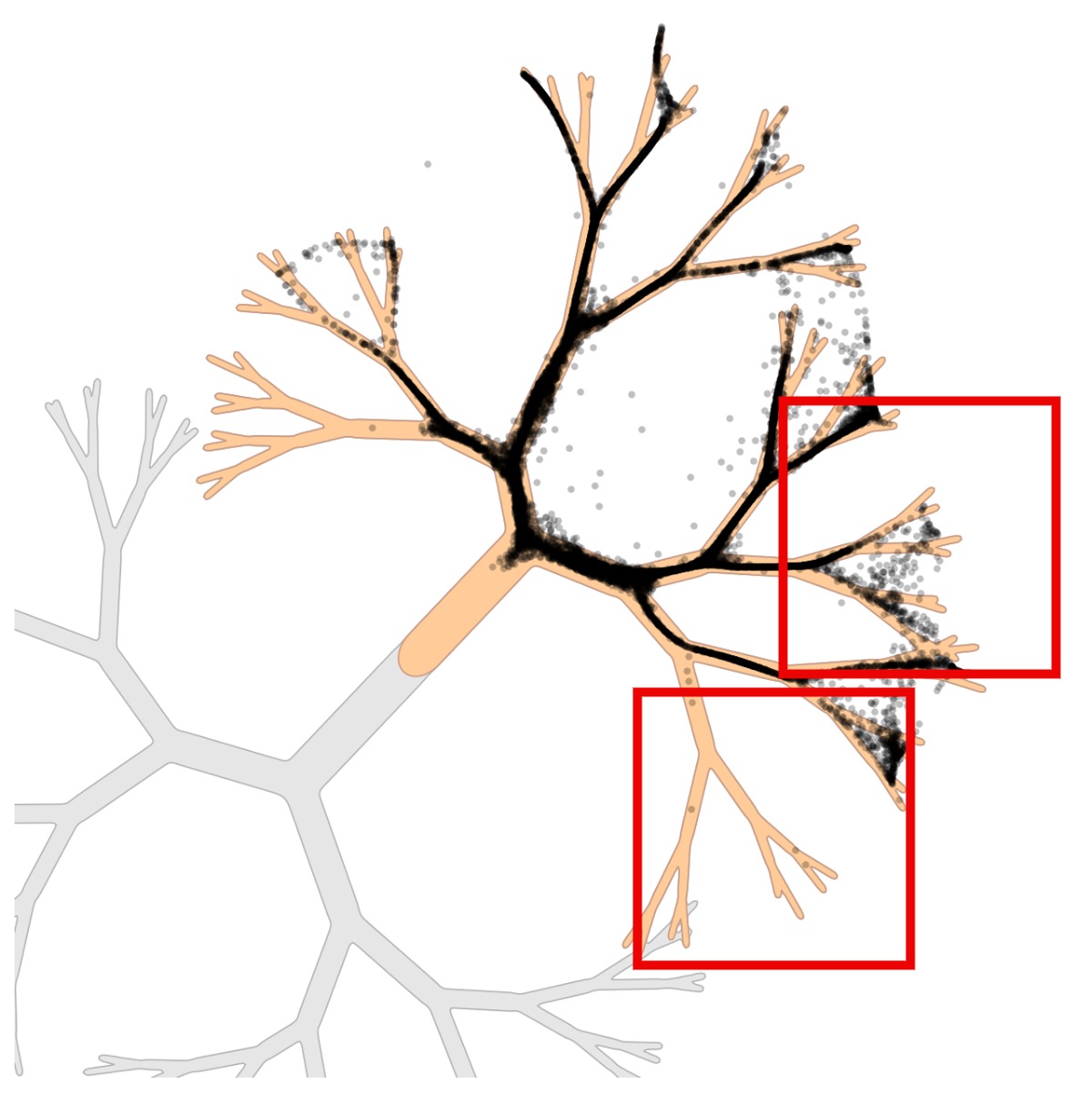} & 
        \includegraphics[width=0.43\linewidth]{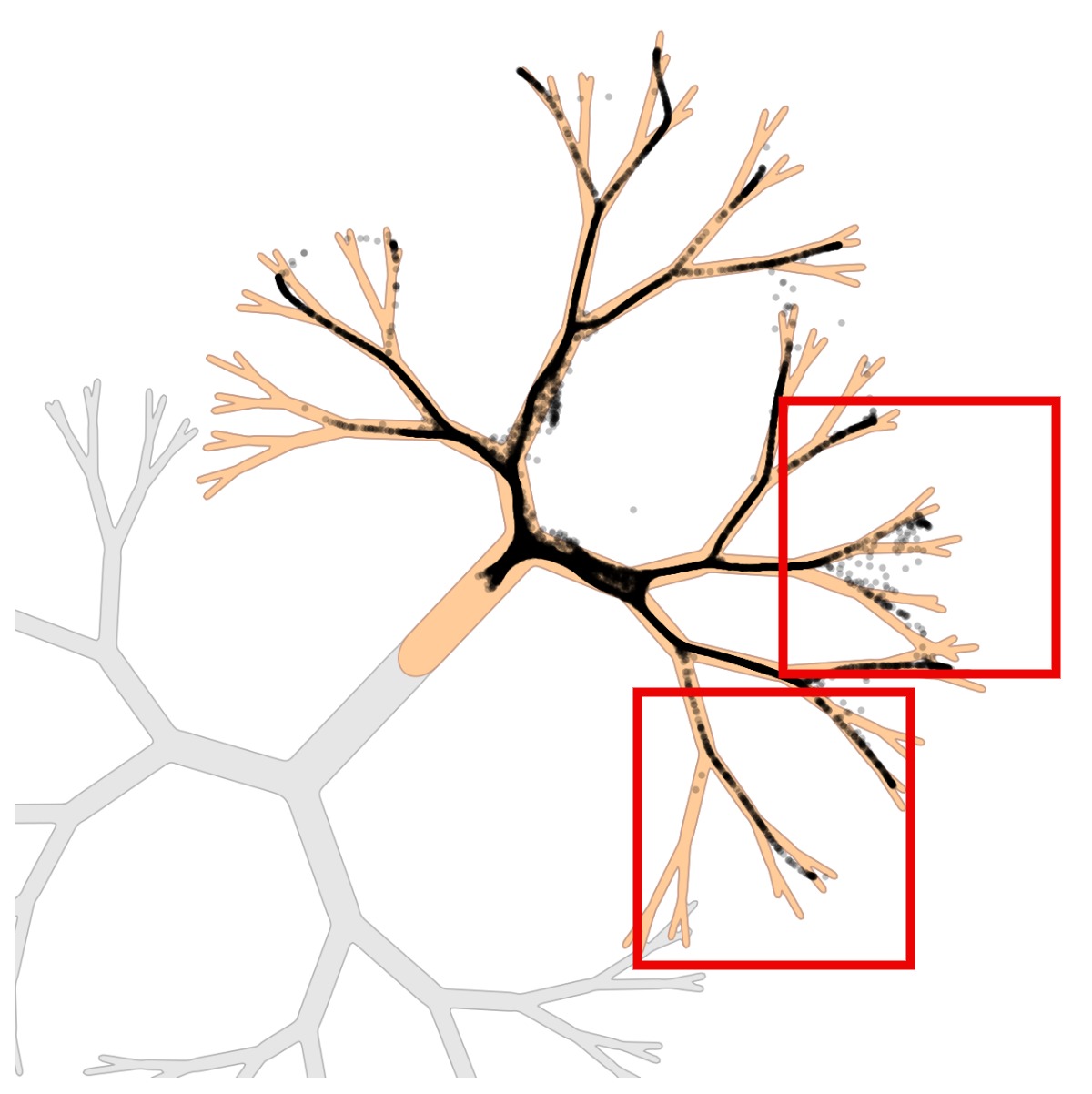} \\
        (c) CFG & (d) \our{}
    \end{tabular}

    \caption{
    A two-dimensional distribution featuring two classes represented by gray and orange regions. {\bf(a)} Ground truth samples from the orange class. {\bf(b)} Conditional sampling with no additional guidance techniques. {\bf(c)} Classifier-free guidance decreases sample diversity to achieve outlier removal {\bf (d) \our{}} preserves the diversity of the samples while still achieving the objective of outlier removal.
    }
    \vspace{-0.5cm}
    \label{fig:toy-example}    
\end{figure}

\section{Introduction}
\label{sec:intro}

Diffusion models \cite{dhariwal2021diffusion,rombach2022high,croitoru2023diffusion} are regarded as one of the leading techniques for image generation, especially due to their ability to be easily conditioned with text prompts. Classifier-free guidance (CFG) \citep{ho2022classifier} is a crucial component in modern diffusion models used for generating content based on text prompts. This method aims to balance diversity and consistency relative to the conditioning factor by employing a mix of constrained and unconstrained diffusion models. In practice, a trade-off \cite{kynkaanniemi2024applying,chung2024cfg++} must be made between the quality of generated elements and their alignment with the prompt. Employing strong guidance results in images that match the conditioned text but of compromised quality. Conversely, using limited guidance yields high-quality results at the expense of alignment with the prompt. 
\begin{figure}
    \centering
        \includegraphics[width=0.95\linewidth, trim=0 10 0 0 , clip]{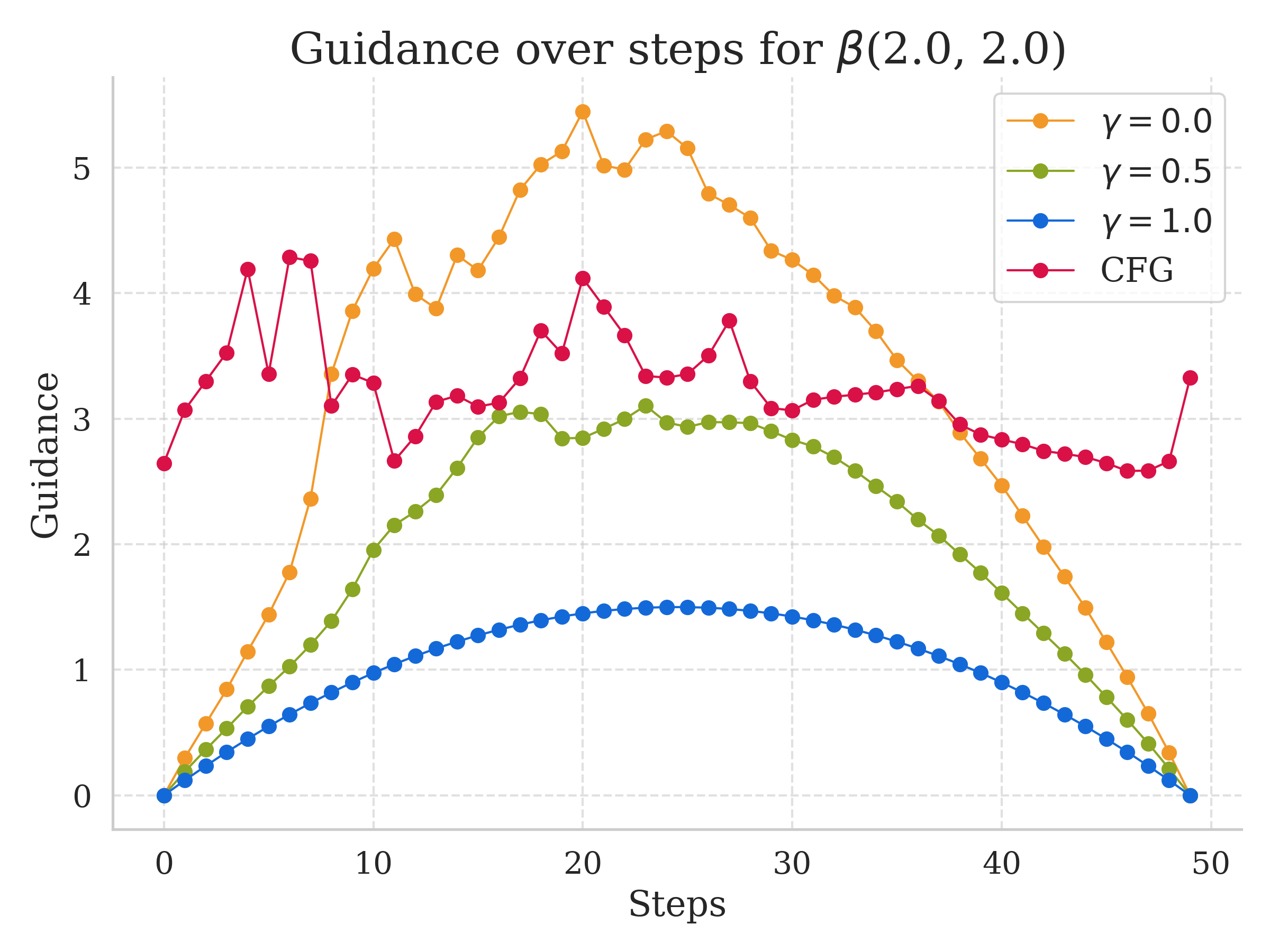}
        \vspace{-0.5cm}
    \caption{  Norm values of the modification factor applied at each iteration of the classifier-free guided diffusion sampling backward process. We compare classical CFG and our solution \our{}. We model such trajectory by $\beta $-distribution and parameter $\gamma$. $\beta $-distribution gives the general trend of a diffusion process. For $\gamma=1$ we have a pure Gamma curve while by going with Gamma to zero, add local perturbation from pure CFG. Thanks to the $\beta$-distribution, we have no guidance at the beginning and at the end of trajectory.}.
    \vspace{-0.9cm}
    \label{fig:curves}
\end{figure}

Using the same guidance for every sampling step isn't optimal because CFG functions uniquely at high, medium, and low noise levels. In \cite{kynkaanniemi2024applying} the authors analyze such three phases. Strong guidance restricts sampling to a few average (template) images in the initial steps. The middle stage is crucial, where guidance modifies important high-order features. In such a part, the guidance can change the sampling trajectory without significantly losing the quality of the renders. In CFG++\cite{chung2024cfg++}, authors introduce a simple modification of CFG, which keeps the trajectory closer to the data manifold.  The last part of diffusion sampling is only denoising, and conditioning can only destroy this process~\cite{poleski2024geoguide}.

This paper introduces \our{}\footnote{The
source code is available at \url{https://github.com/gmum/beta-CFG}} ($\beta$-distribution Classifier-Free Guidance), which controls the impact of guidance during generation to solve the above trade-off. \our{} use a family of single-modal curve families ($\beta$-distribution) to model the strength of guidance. Instead of scores between a conditional and an unconditional diffusion model we normalize such value. A similar strategy was used in Classifier Guided Diffusion~\cite{poleski2024geoguide} where fixes constant classification guidance weight was used.
In Classifier-Free Guidance, we need to control the impact of conditioning dynamically, so we use the additional parametric function. Thanks to this approach, we can dynamically change the trade-off between prompt matching and sample quality. 
The single modal $\beta$-distribution allows the data manifolds to remain at the beginning and end of the sampling trajectory. 
Furthermore, we use additional $\beta$ parameters that control the middle stage of the diffusion process. 

Due to this adjustment, we can more accurately represent the data distribution. This is demonstrated in a 2D illustration; refer to Fig.~\ref{fig:toy-example}. As observed, the traditional CFG fails to draw samples from the data distribution, evidenced by the bottom right branch in Fig.~\ref{fig:toy-example}~(c). Conversely, \our{} aligns more closely with the training data distribution, avoiding outlier generation, see Fig.~\ref{fig:toy-example}~(c).

Concluding, the main contributions of the paper are the following:
\begin{itemize}
    \item we propose \our{} a model which solves the tartrate-of between prompt fitting and quality of generated objects
\vspace{-0.3cm}    
    \item \our{} is easy to implement and controls the norm of the guidance (see~Fig.~\ref{fig:curves}),
\vspace{-0.3cm}    
    \item \our{} surpasses the traditional CFG in terms of FID score while maintaining a constant CLIP value.  
\vspace{-0.3cm}    
\end{itemize}

\section{Related works}

\paragraph{Diffusion models}

The idea of diffusion models was first presented in \cite{sohl2015deep}. These models leverage Stochastic Differential Equations (SDEs) to progressively transform a simple initial distribution (e.g., a normal distribution) into a more complex target distribution through a series of manageable diffusion steps. The evolving advances, including the decrease in the trajectory steps~\cite{bordes2017learning}, have created more efficient diffusion models.

\begin{figure*}[!ht]
    \centering
    \setlength{\tabcolsep}{0.8pt}    
    \centering
    \begin{tabular}{@{}c@{}c@{}c@{}c@{}c@{}c@{}}
        \rotatebox{90}{ \qquad CFG} &
        \includegraphics[width=0.19\linewidth]{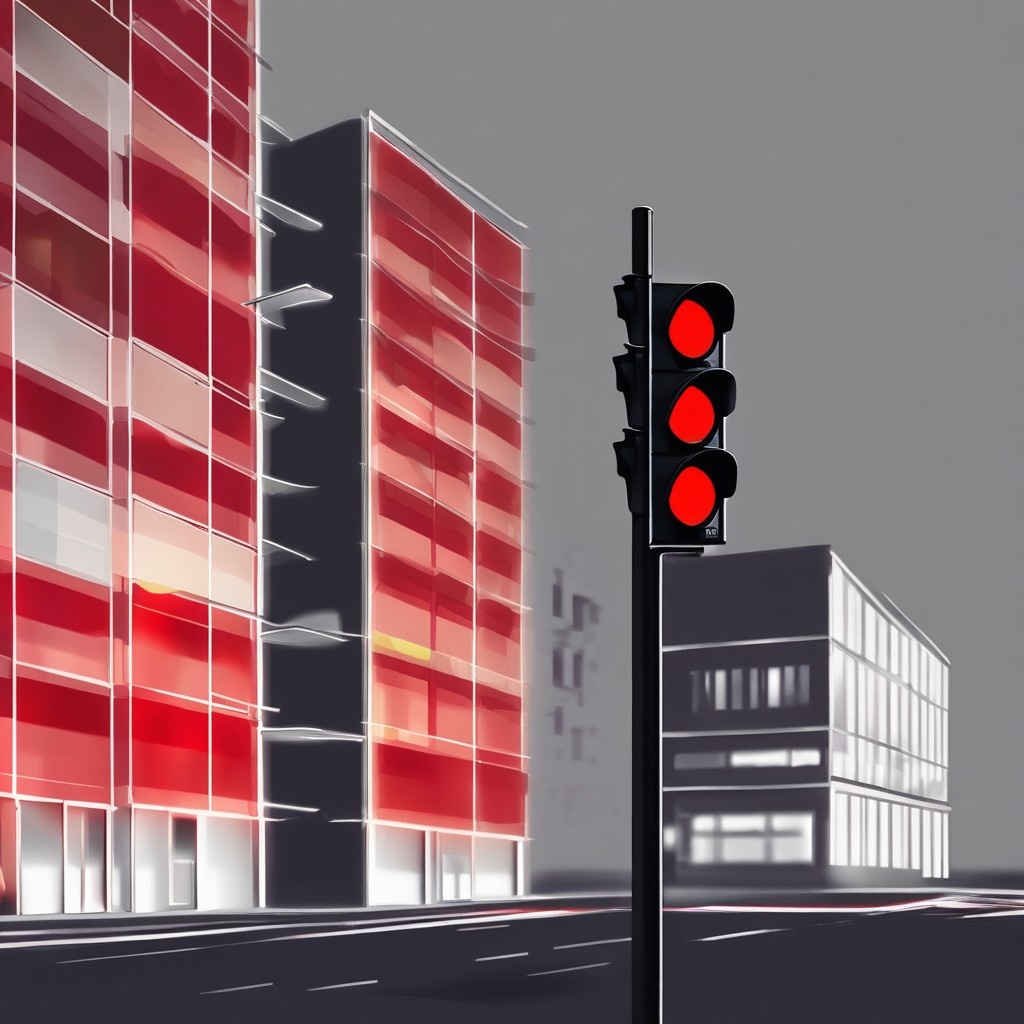} & 
        \includegraphics[width=0.19\linewidth]{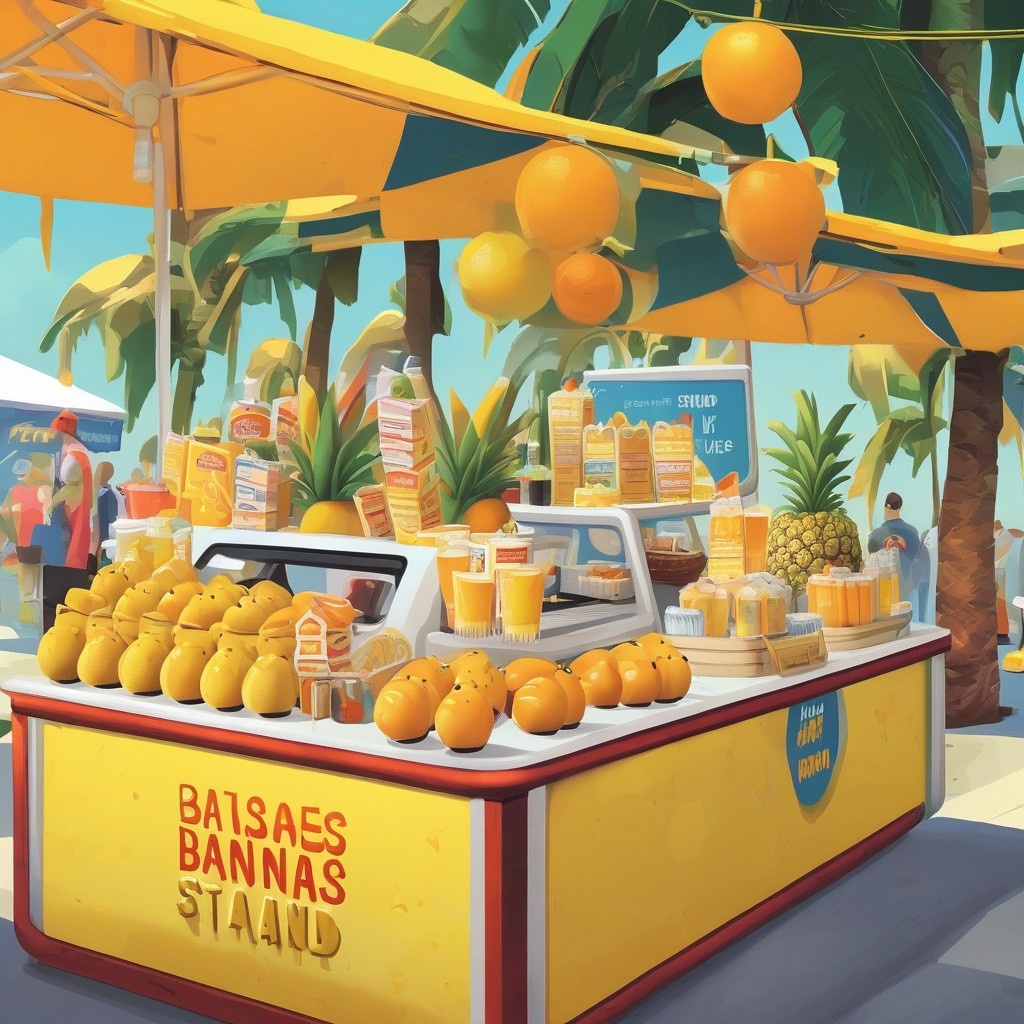} & 
        \includegraphics[width=0.19\linewidth]{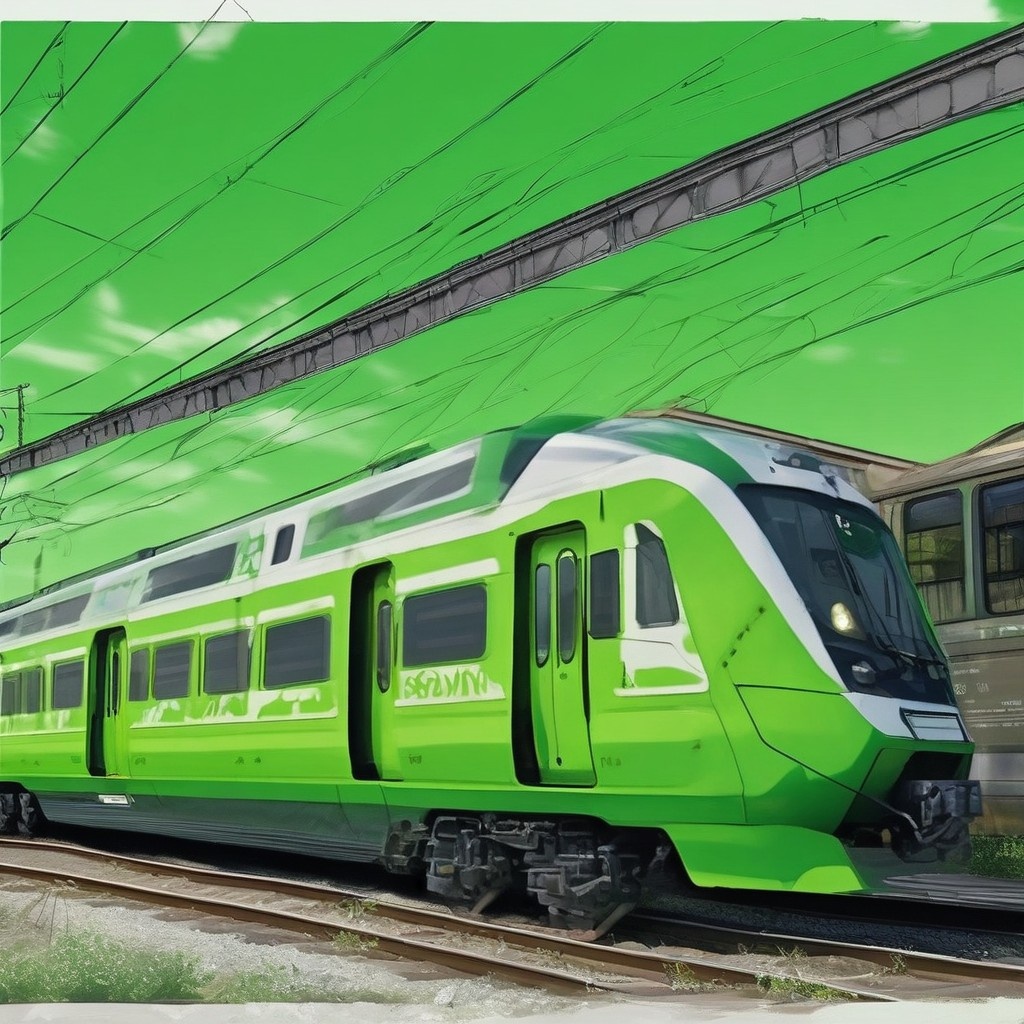} & 
        \includegraphics[width=0.19\linewidth]{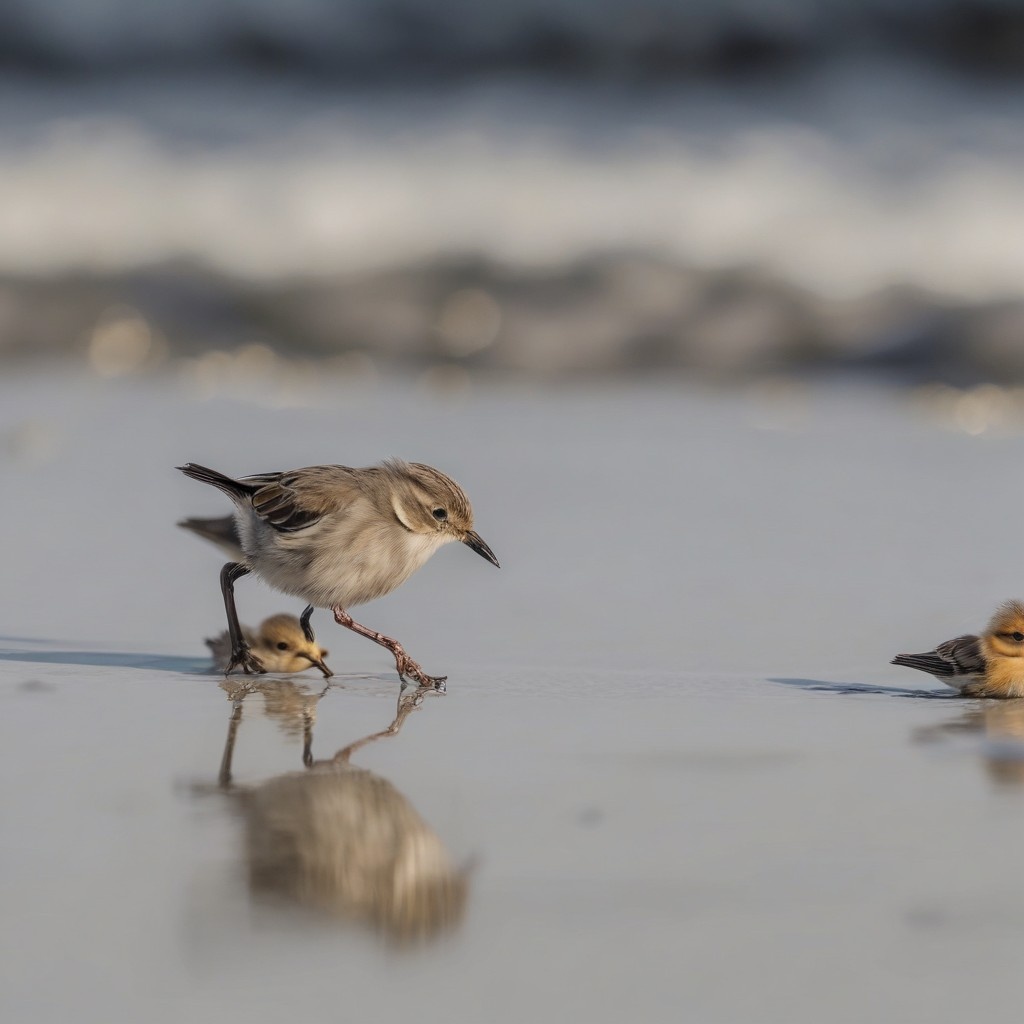} & 
        \includegraphics[width=0.19\linewidth]{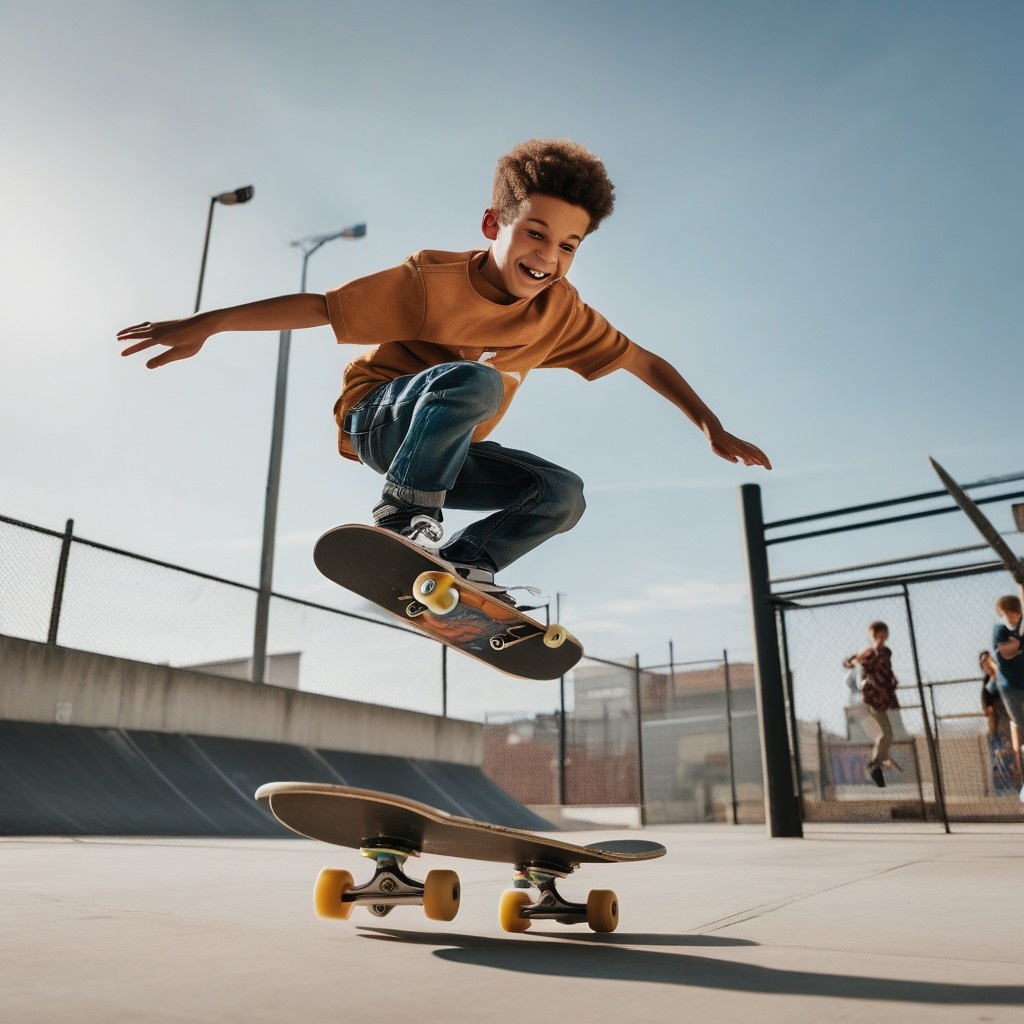} 
        \\
        \rotatebox{90}{ \qquad \our{}} &
        \includegraphics[width=0.19\linewidth]{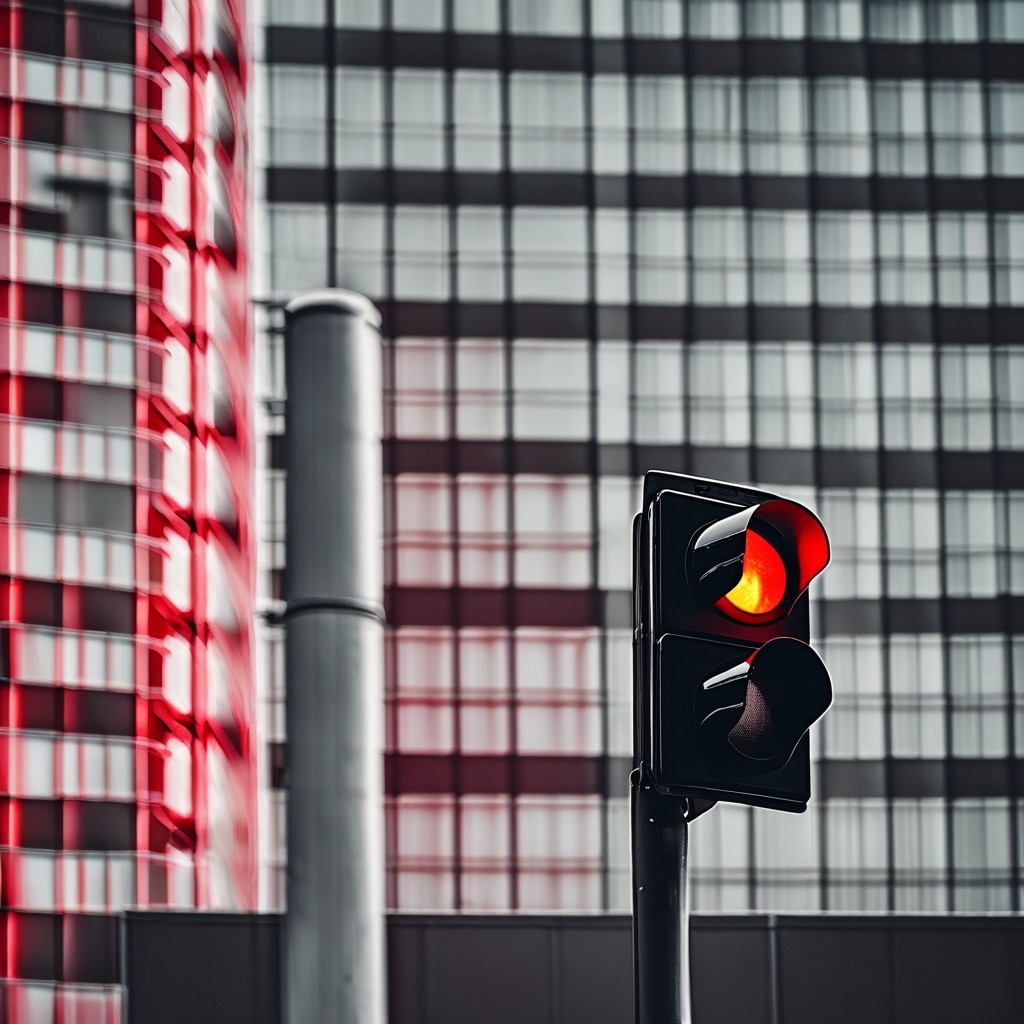} & 
        \includegraphics[width=0.19\linewidth]{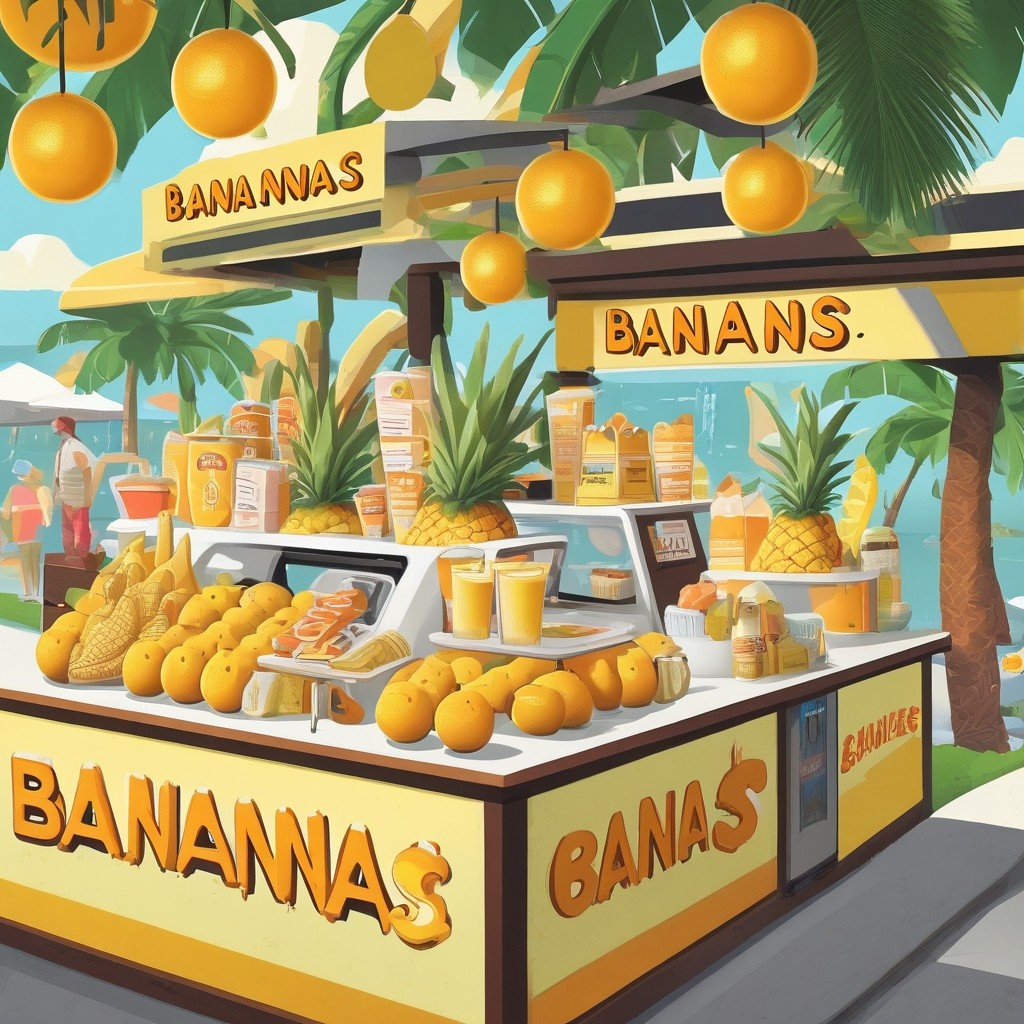} & 
        \includegraphics[width=0.19\linewidth]{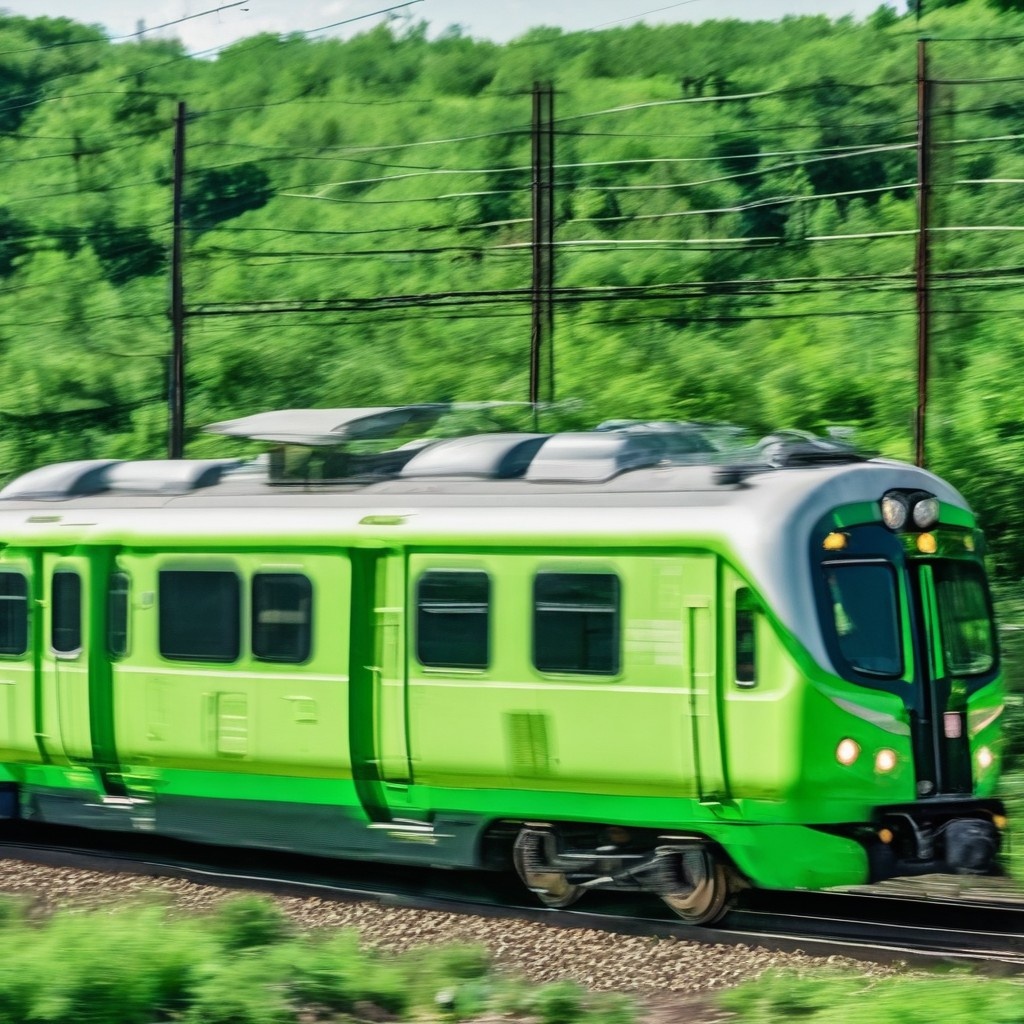} &  
        \includegraphics[width=0.19\linewidth]{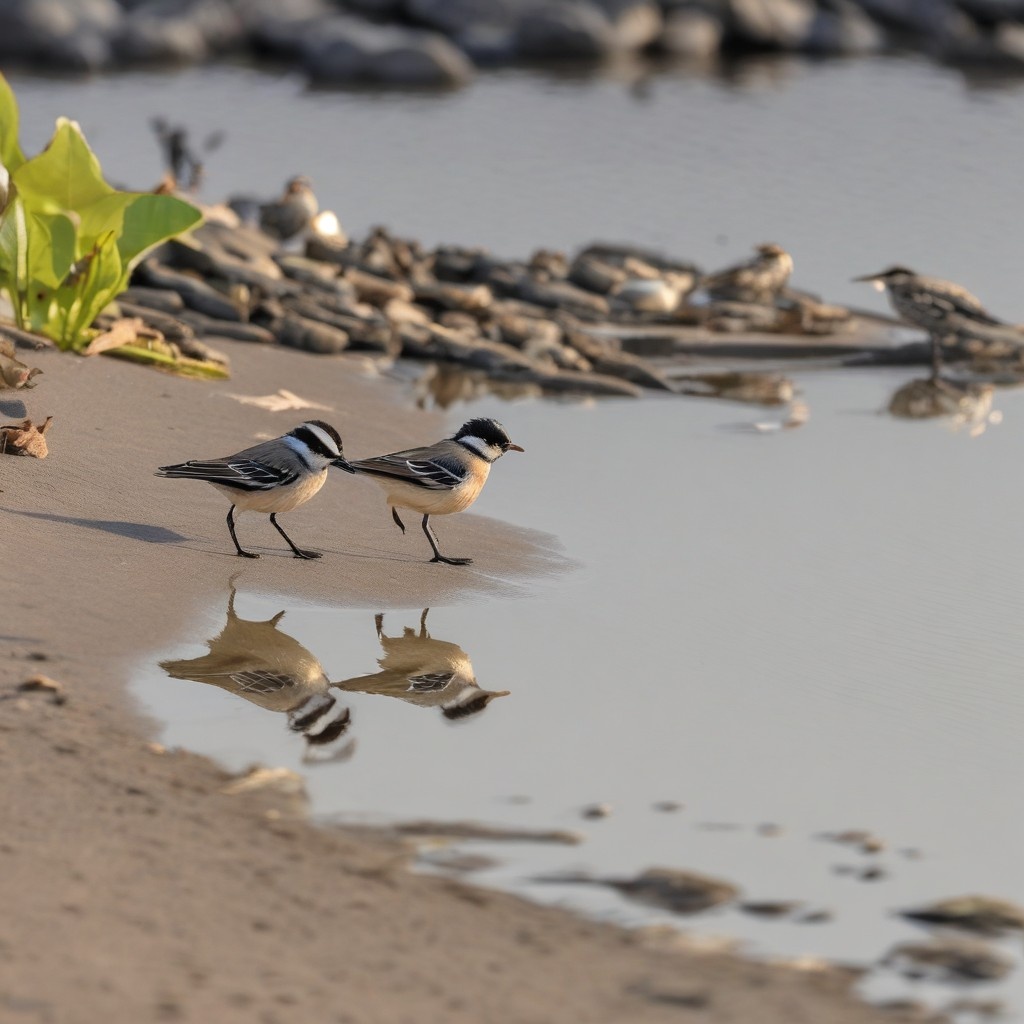} & 
        \includegraphics[width=0.19\linewidth]{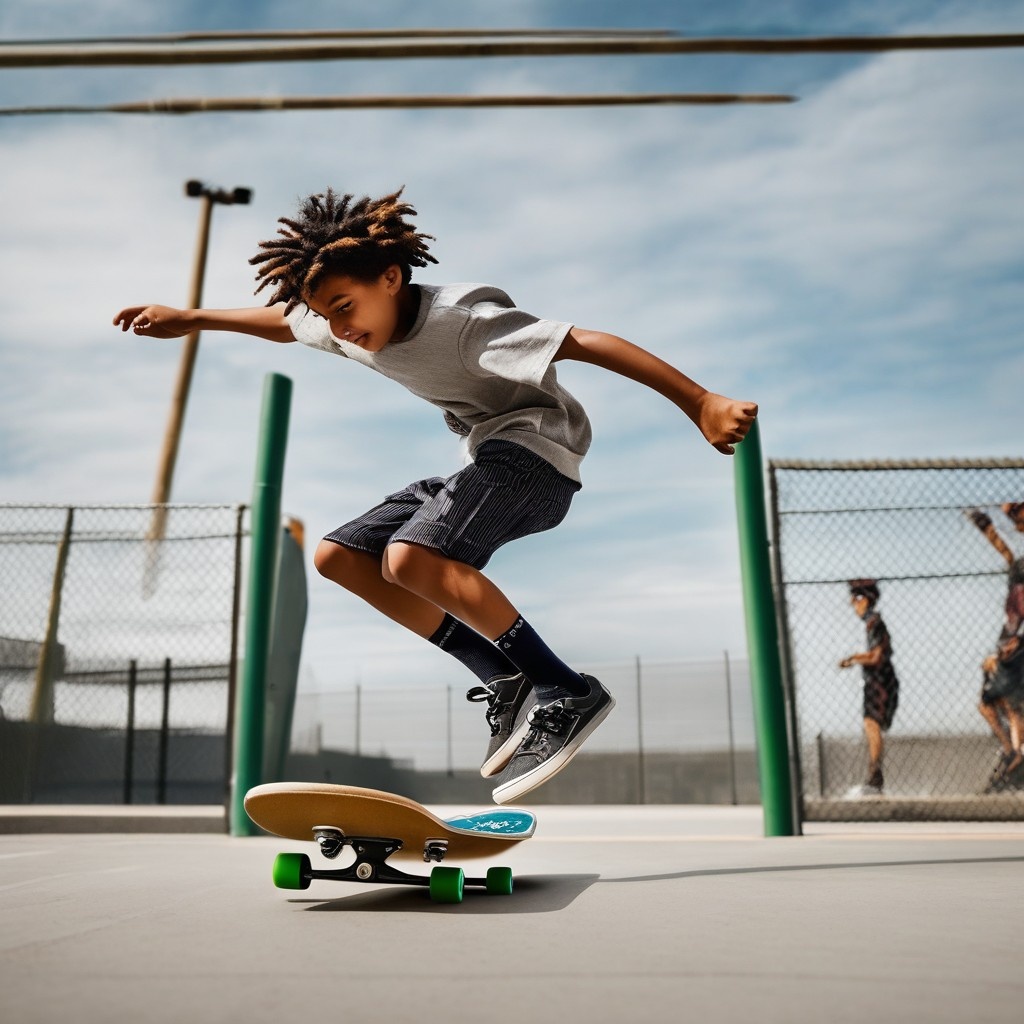}  
        \\[0.08cm]
        &
        \multicolumn{1}{p{0.16\linewidth}}{\centering \small  A traffic light sitting next to a building with its red blurred.
        }& 
        \multicolumn{1}{p{0.16\linewidth}}{\centering \small 
        There are bananas, pineapples, oranges, sandwiches, and drinks at the stand.
        }& 
        \multicolumn{1}{p{0.16\linewidth}}{\centering \small 
        a green train is coming down the tracks
        }& 
        \multicolumn{1}{p{0.16\linewidth}}{\centering \small 
        Small birds are walking along the waters edge
        }& 
        \multicolumn{1}{p{0.16\linewidth}}{\centering \small 
        A boy is jumping a hurdle while on a skateboard.
        }
        
        \\[1.2cm]
        \rotatebox{90}{ \qquad CFG} &
        \includegraphics[width=0.19\linewidth]{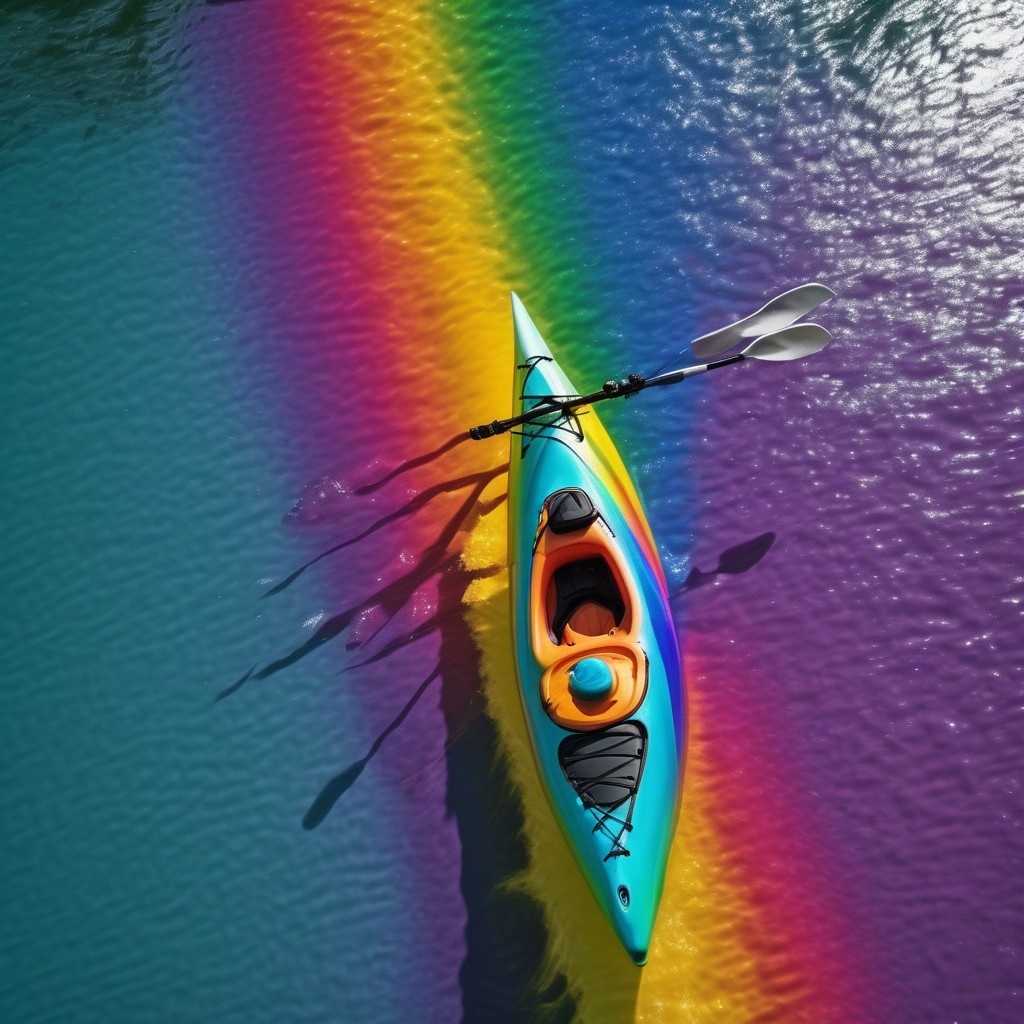} & 
        \includegraphics[width=0.19\linewidth]{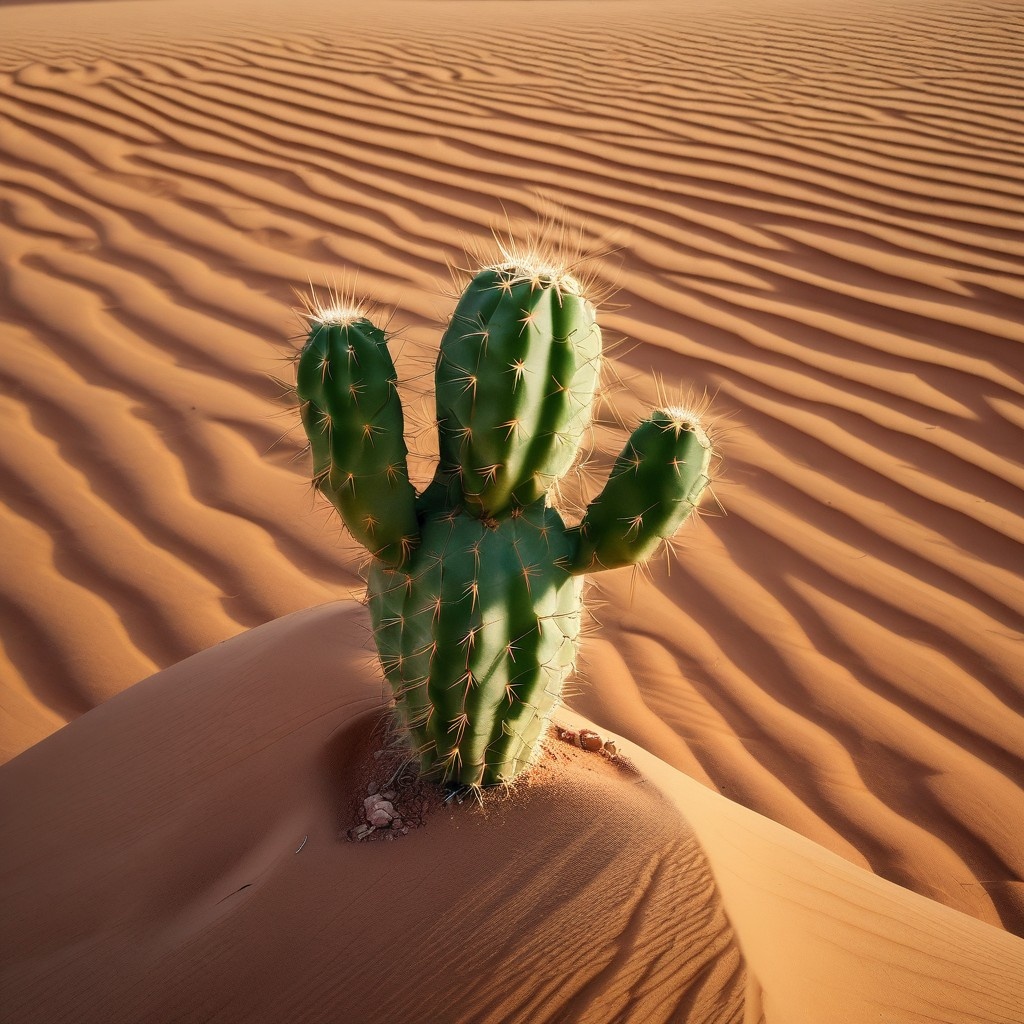} & 
        \includegraphics[width=0.19\linewidth]{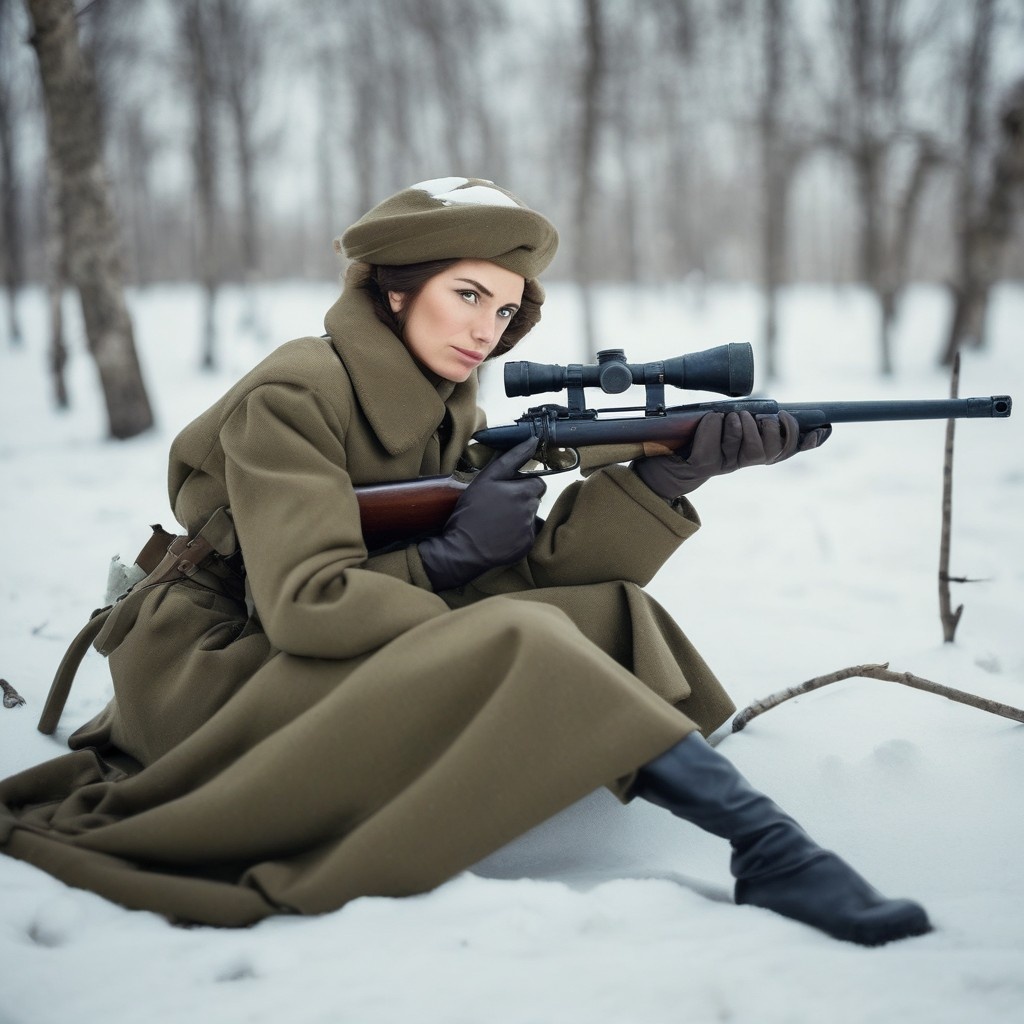} & 
        \includegraphics[width=0.19\linewidth]{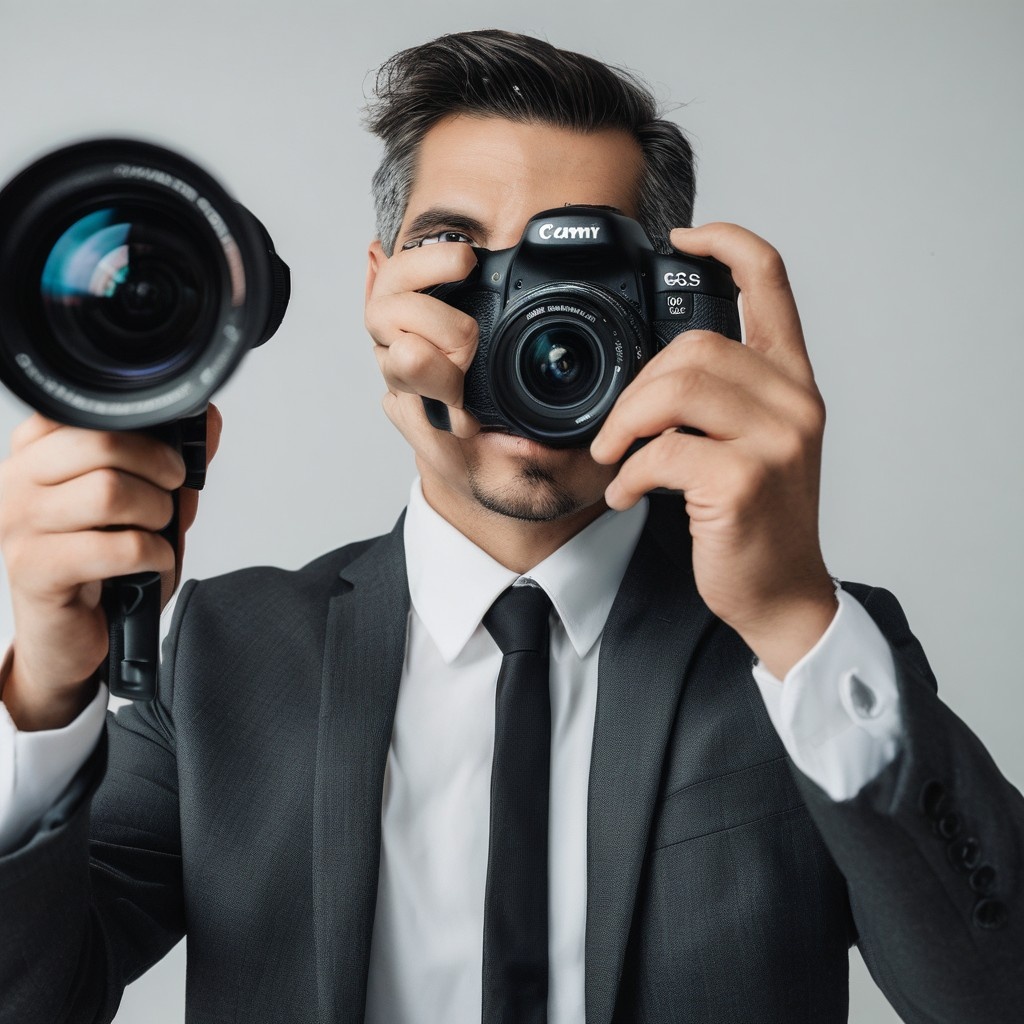} & 
        \includegraphics[width=0.19\linewidth]{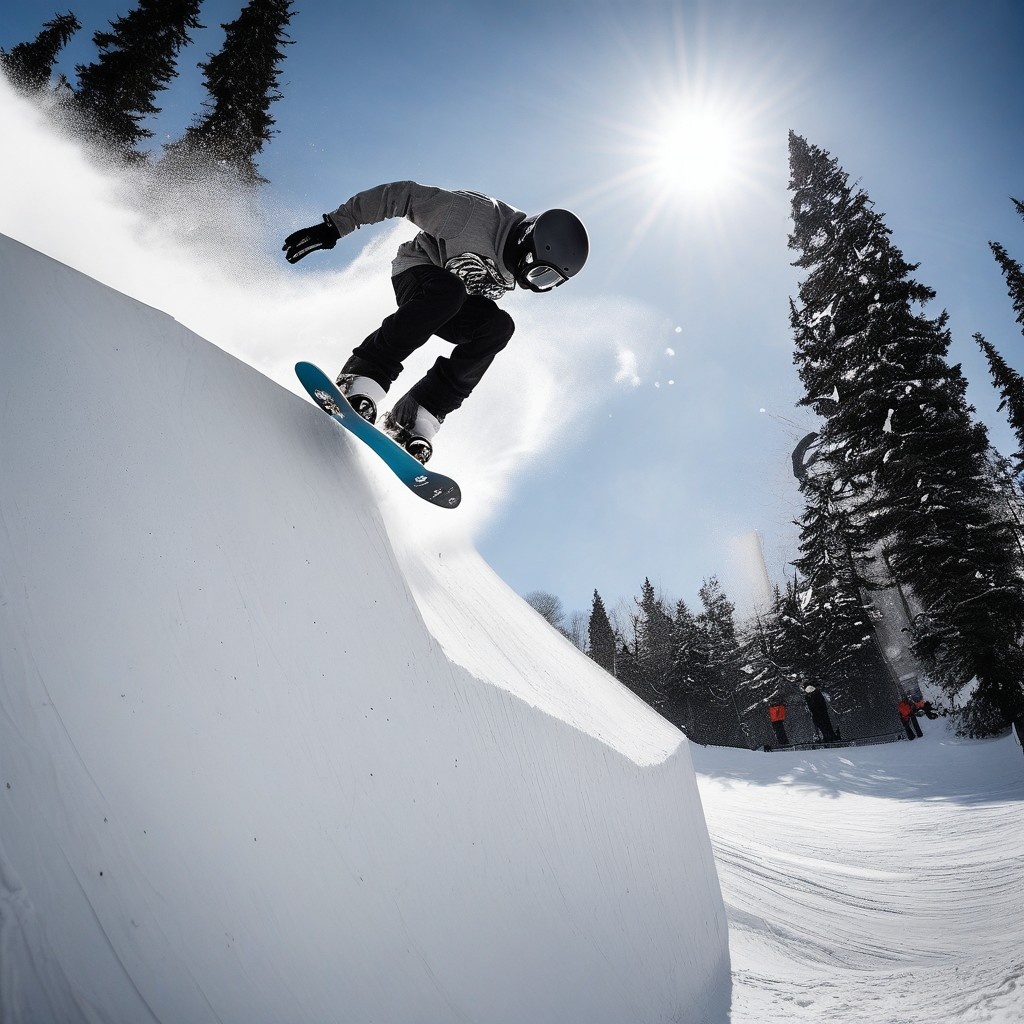} 
        \\
        \rotatebox{90}{ \qquad \our{}} &
        \includegraphics[width=0.19\linewidth]{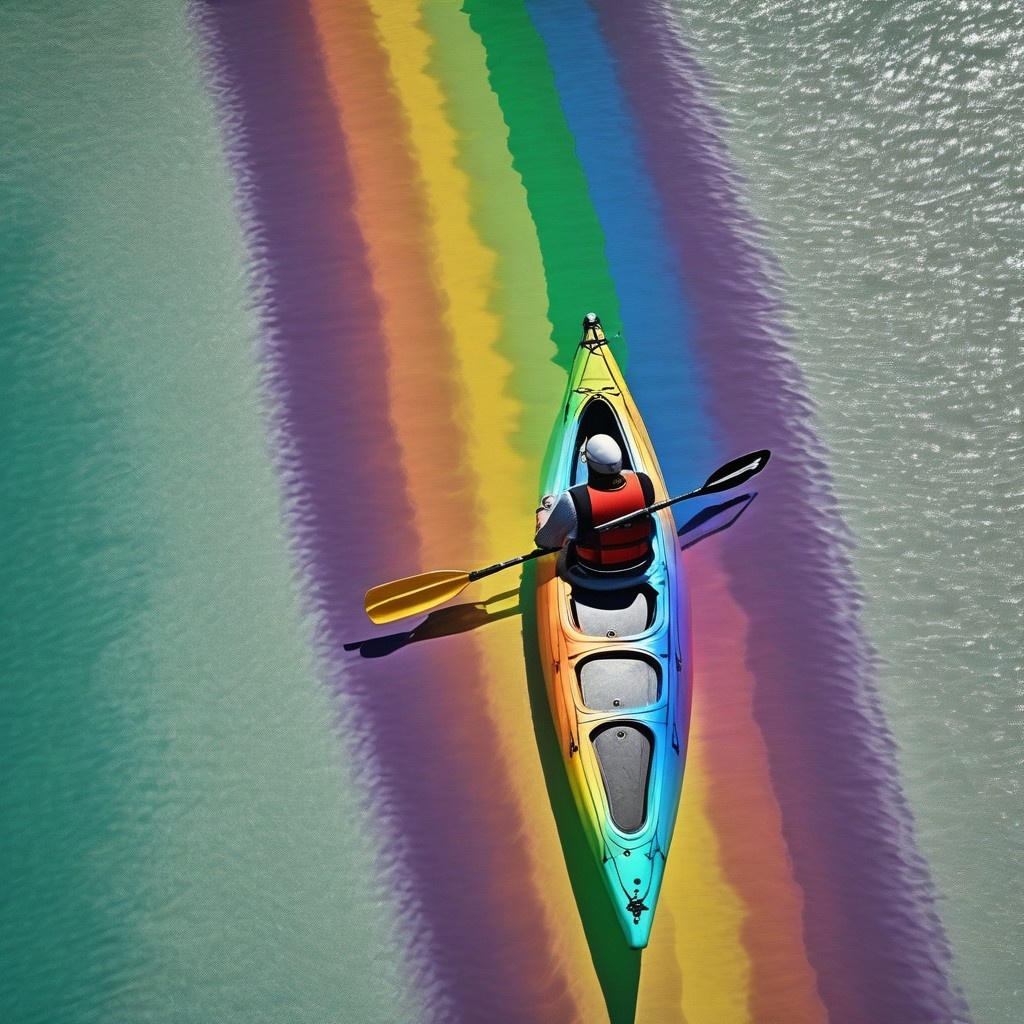} & 
        \includegraphics[width=0.19\linewidth]{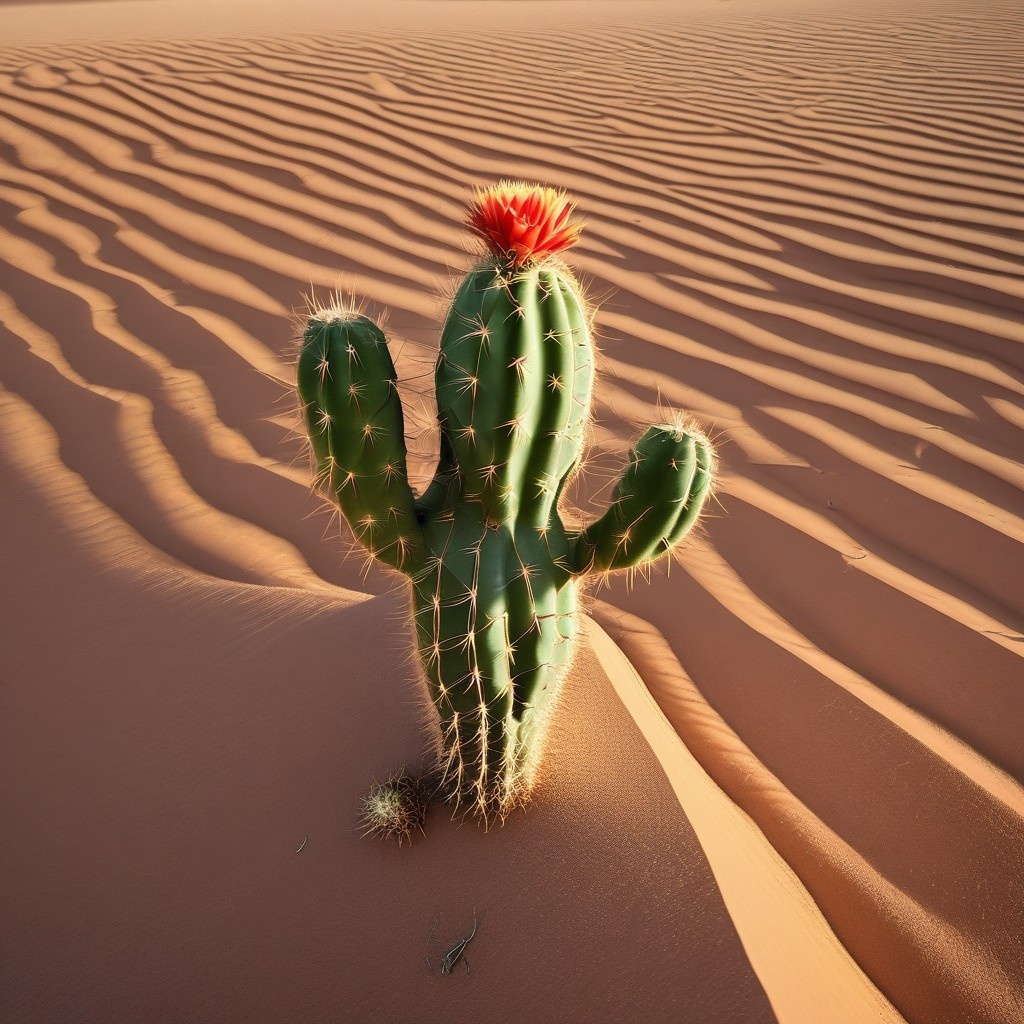} & 
        \includegraphics[width=0.19\linewidth]{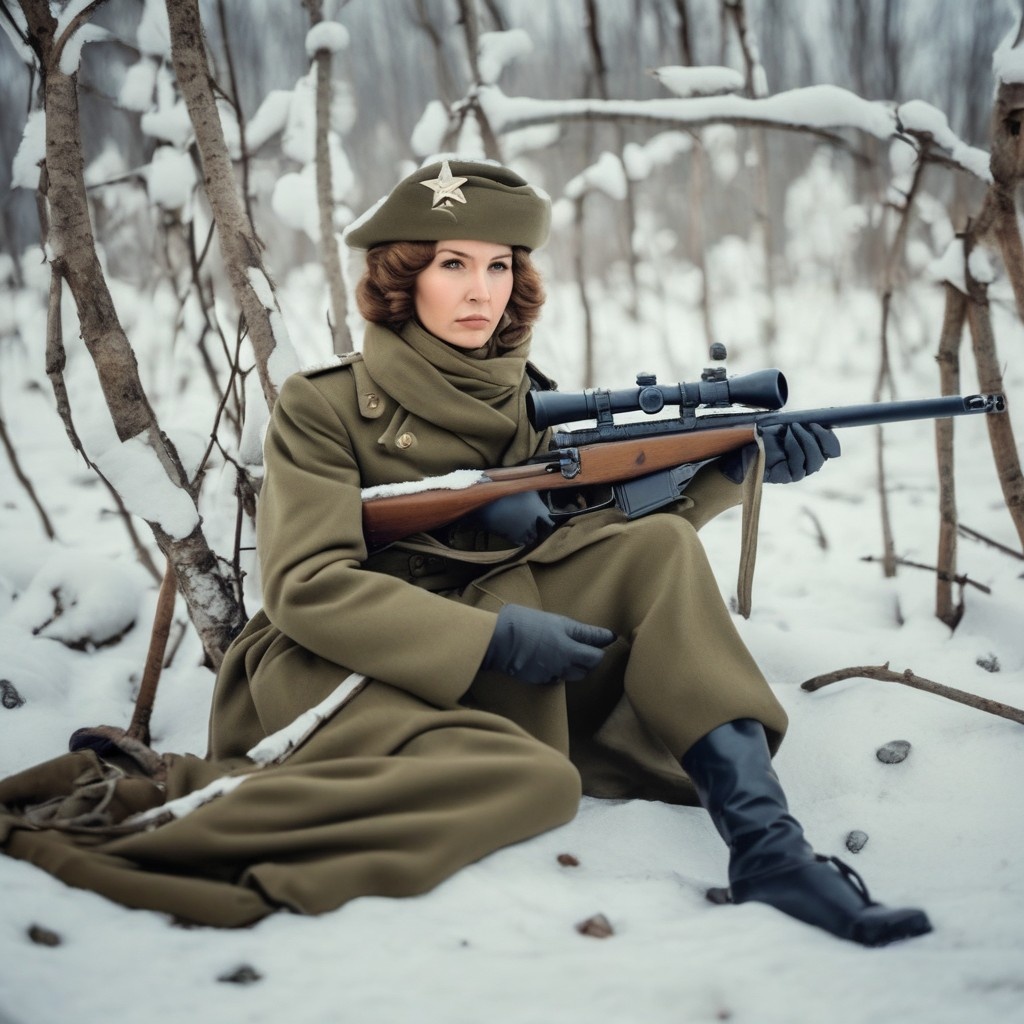} &  
        \includegraphics[width=0.19\linewidth]{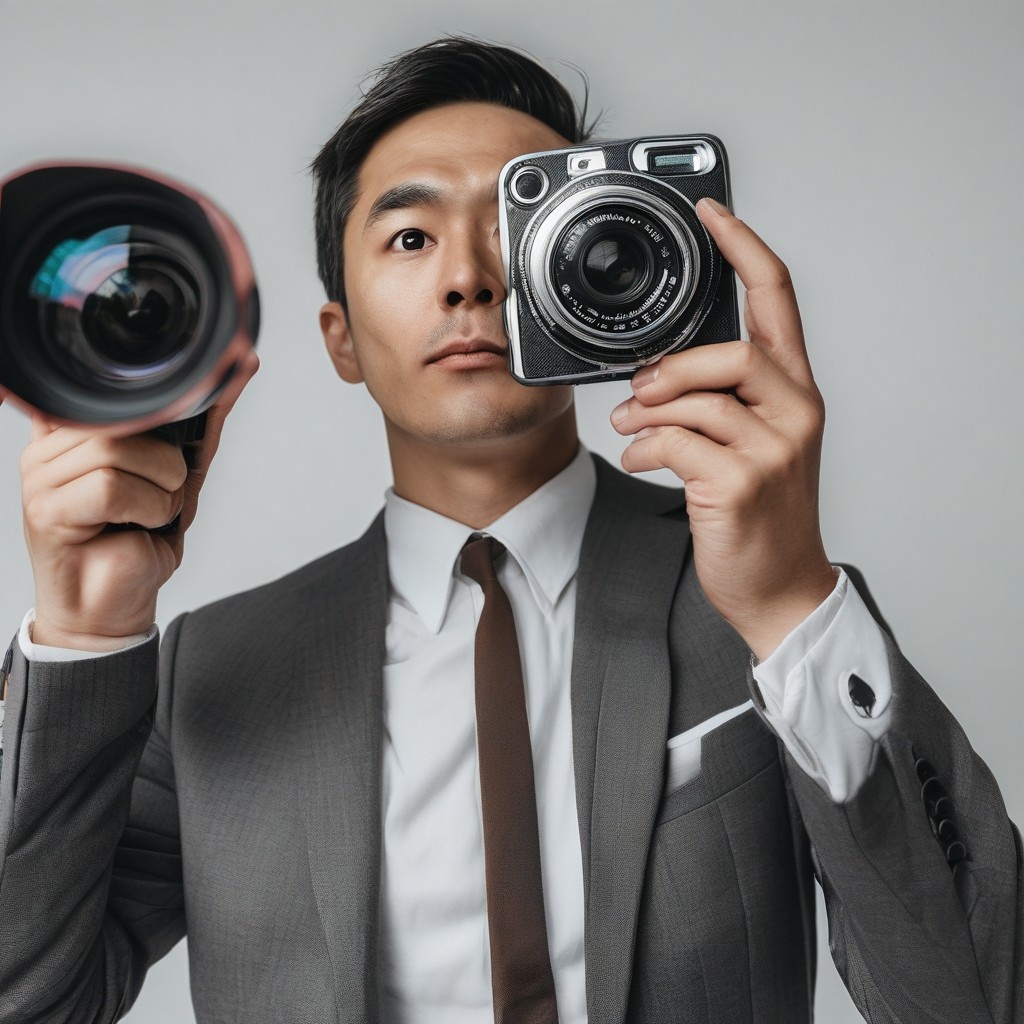} & 
        \includegraphics[width=0.19\linewidth]{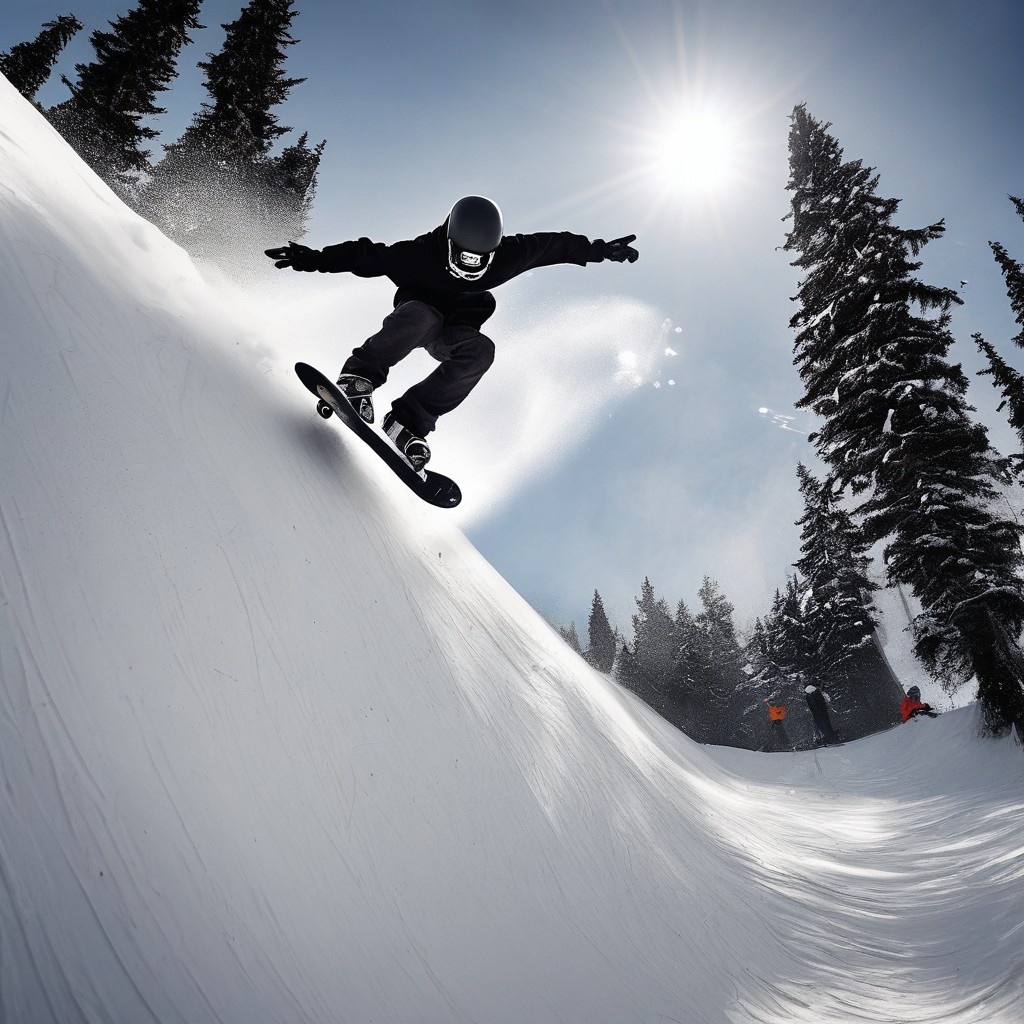}
        \\[0.08cm]
                &
        \multicolumn{1}{p{0.16\linewidth}}{\centering \small kayak in the water, optical color, aerial view, rainbow} & 
        \multicolumn{1}{p{0.16\linewidth}}{\centering \small A small cactus with a happy face in the sahara desert}& 
        \multicolumn{1}{p{0.16\linewidth}}{\centering \small woman sniper, wearing soviet army uniform, in snow ground}& 
        \multicolumn{1}{p{0.16\linewidth}}{\centering \small A man wearing a suit is taking a self portrait with a camera}& 
        \multicolumn{1}{p{0.16\linewidth}}{\centering \small A snowboarder hitting a trick off a giant ramp.}

    \end{tabular}
    \vspace{-0.4cm}
    \caption{Comparison of CFG and \our{}. As we can see, our model produces more realistic images, which is consistent with the numerical results from Tab.~\ref{tab:all}.} 
    \vspace{-0.4cm}
    \label{fig:sampling}
\end{figure*}

Significant progress was made in developing diffusion probabilistic denoising models (DDPM)~\citep{ho2020denoising,dhariwal2021diffusion}. DDPMs utilize a weighted variational bound objective by integrating probabilistic diffusion models with denoising score matching~\citep{song2019generative}. Despite demonstrating excellent generative capabilities and producing high-quality samples, the substantial computational expense of these models presented a significant drawback. Denoising Diffusion Implicit Models (DDIMs)~\citep{song2020denoising} improve scalability, particularly sample efficiency.

Ultimately, Latent Diffusion Models~\citep{rombach2022high} reduced the significant computational demands associated with applying diffusion models to high-dimensional scenarios by suggesting the implementation of diffusion within the low-dimensional latent space of an autoencoder. The practical application of this method is shown in Stable Diffusion~\citep{rombach2022high}. Scalability was enhanced further in models such as SDXL~\citep{podell2023sdxl}, which expanded the potential of latent diffusion models to tackle larger and more intricate tasks.

\paragraph{Guidance of diffusion models}
The Stochastic Differential Equation (SDE) framework is vital in diffusion models. While it facilitates exceptional generative capabilities, better scalability, and expedited training compared to models utilizing Ordinary Differential Equations (ODEs)~\citep{dinh2014nice,rezende2015variational, grathwohl2018ffjord}, the process of stochastic inference still needs \textit{guidance} to generate satisfactory samples.

Various techniques have been designed to guide the generation process in a specific direction. These can largely be classified into different strategies: guidance through classifiers, Langevin dynamics, Markov Chain Monte Carlo (MCMC), external guiding signals, architecture-specific features, etc. Despite their variations, these techniques generally steer the diffusion process toward areas of minimal energy, as inferred by different proxies.

A widely recognized approach is Classifier Guidance~\citep{dhariwal2021diffusion,poleski2024geoguide}, which incorporates an external classifier to infer the class from intermediate noisy diffusion steps. On the other hand, Classifier-Free Guidance~\citep{ho2022classifier} removes the requirement for an external classifier by using a unified model trained in both conditioned and unconditioned modes. Langevin dynamics is frequently used for off-policy guidance, where during each step along a trajectory, the model aligns with the scaled gradient norm toward areas of minimal energy (the maximum \textit{log probability}) \cite{zhang2021path} and \cite{sendera2024diffusion}. Alternatively, MCMC sampling strategies are directly used in diffusion processes \citep{song2023loss,chung2023diffusion}.

Specific techniques integrate external guiding functions to refine the generation path towards the targeted results \cite{bansal2023universal}. Alternatively, some exploit the inherent features of diffusion models, such as leveraging intermediate self-attention maps \citep{hong2023improving} or employing an externally trained discriminator network \citep{kim2022refining}. AutoGuidance \citep{karras2024guiding} enhances classifier-free guidance by substituting the unconditional model with a more compact, less sophisticated version to direct the conditional model.


\section{Background}

\paragraph{Diffusion models}

Diffusion models are generative algorithms that produce new samples through a gradual denoising process. This process begins with an initial Gaussian noise sample, denoted as $x_T$, and progressively refines it through steps that reduce noise, producing samples $x_{T-1}, x_{T-2}, \ldots, x_0$. The end result, $x_0$, lies on the data manifold. At each step $t$, there is a designated noise level, where $x_t$ combines the underlying signal $x_0$ and Gaussian noise $\epsilon$. The parameter $t$ controls the intensity of noise at each step. Training diffusion models involves randomizing over noise levels and time steps to produce a denoised $x_{t-1}$ from $x_t$. This denoising process is often modeled by U-Net ~\cite{ho2020denoising}.

Diffusion models involve two key processes: the forward and the reverse diffusion process. Consider $q(x_0)$ to be the data distribution such that $x_0 \sim q(x_0)$. This forward process introduces small Gaussian noise to the sample over $T$ steps, resulting in a sequence $x_{0}, \ldots, x_{T}$. The process is controlled by parameter $\{ \beta_{t} \in (0,1) \}_{t=1}^T$:
\begin{equation} \label{eq:0}
q(x_t|x_{t-1}) := \N(x_t ; \sqrt{1-\beta_t} x_{t-1},\beta_{t}I).
\end{equation}
Using this formulation, $x_t \sim q(x_t|x_0)$ can be calculated in one step:
\begin{equation}
\label{eq:epsilon}
\begin{split}
q(x_t|x_0) &  = \N(x_t ; \sqrt{ \bar \alpha_t } x_0, (1-\bar \alpha_t) I) \\
& = \sqrt{ \bar \alpha_t } x_0 + \e \sqrt{ 1-\bar \alpha_t }, \mbox{ } \e \sim \N(0,I),
\end{split}
\end{equation}
where $\alpha_t = 1-\beta_t$ and $\bar \alpha_t = \prod_{s=0}^t \alpha_t$.


The calculation of the backward process is more challenging and requires access to the posterior distribution $q(x_{t-1}|x_t,x_0)$, which is Gaussian, with the mean given by $\tilde \mu_t (x_t, x_0)$ and the variance represented by $\tilde \beta_t$:
\begin{equation}
 q(x_{t-1}|x_t,x_0) = \N( x_{t-1}; \tilde \mu_t(x_t,x_0),\tilde \beta_t I ),   
\end{equation}
where $\tilde \mu_t(x_t,x_0) := \frac{\sqrt{\bar \alpha_{t-1}}\beta_t}{1- \bar \alpha_t} x_0 + \frac{\sqrt{\alpha_t}(1-\bar \alpha_{t-1})}{1-\bar \alpha_{t}}x_t $ and $\tilde \beta_t := \frac{1-\bar \alpha_{t-1}}{1-\bar \alpha_{t}} \beta_t$.

Practically, a neural network is applied to approximate conditional probabilities  $q(x_{t-1}|x_t)$.  In \cite{sohl2015deep} the authors show that as $T \to \infty$, $q(x_{t-1}|x_t)$ converges towards a diagonal Gaussian distribution and $\beta_t$ approaches zero. In this context, a neural network is trained to predict both the mean $\mu_{\theta}$ and a diagonal covariance matrix $\gamma_t I$ for the reverse diffusion process:
\begin{equation}
p(x_{t-1}|x_{t}):= \N(x_{t-1};\mu_{\theta}(x_t,t), \gamma_t I).  
\end{equation}
To ensure $p(x_0)$ effectively represents the true data distribution, $q(x_0)$ variational lower bound should be optimized in the training process. In practice, \cite{ho2020denoising} suggests training a model $\e_{\theta}(x_t, t)$ to approximate $\e$ from equation (\ref{eq:epsilon}) instead of modeling mean $\mu_{\theta}$ directly. The training objective is formulated as follows:
\begin{equation}
\cost{}:= \E_{t\sim [1,T],x_t \sim q(x_t|x_0), \e \sim \N (0,I)}  \| \e - \e_{\theta}(x_t,t) \|^2 ,
\end{equation}
where $\gamma_t$ is usually a fixed value, such as $\beta_t$ or $\tilde \beta_t$ representing the maximum and minimum boundaries for the true reverse-step variance, respectively.

For sampling purposes the mean $\mu_{\theta}(x_t, t)$ from $\e_{\theta}(x_t, t)$ can be calculated using the following formula:
\begin{equation}
\mu_{\theta}(x_t,t) = \frac{1}{\sqrt{\alpha_t}} \left( x_t - \frac{1-\alpha_t}{\sqrt{ 1-\bar \alpha_t }} \e_{\theta}(x_t,t)  \right).
\end{equation}
For clarity, we further use $\e(x_t):=\e_{\theta}(x_t,t)$. 


\paragraph{Classifier Guidance.} The process of generating samples from unconditioned, trained model $\e(x_t, t)$ can be guided using classification model $p(y|x_t)$, that predicts class $y$ from intermediate noisy samples $\mathbf{x}_t$ \citep{dhariwal2021diffusion}. The sampling procedure is performed according to modified updates considering the formula below:
\begin{equation}
    \hat{\epsilon}(x_t) = \epsilon(x_t) - \sqrt{1 - \bar{\alpha}_t} w \nabla_{x_t} \log p(y \vert x_t),
\end{equation}   
where $w$ is a scaling factor that adjusts the strength of the classifier’s influence, and $\bar{\alpha}_t = \prod_{i=1}^t 1 - \beta_i$. While effective, classifier-based guidance introduces several downsides, such as added complexity, the need of training an additional classifier and potential inaccuracies due to classifier errors.

\begin{figure*}[t]
    \centering
    \renewcommand{\arraystretch}{0}
    \setlength{\tabcolsep}{0.4pt}
    \begin{tabular}{c@{}c@{}c@{}c@{}c@{}c@{}}
        
        \includegraphics[width=0.15\linewidth]{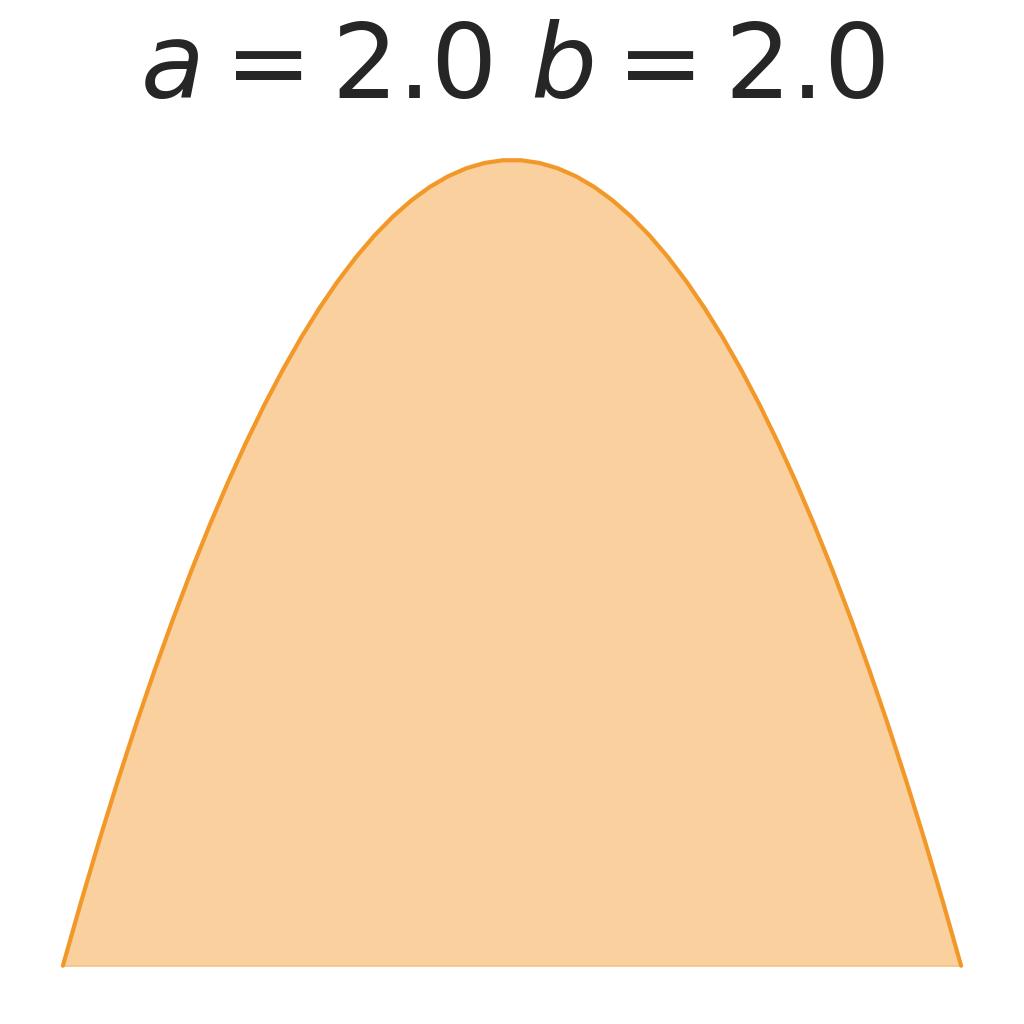} &
        \includegraphics[width=0.15\linewidth]{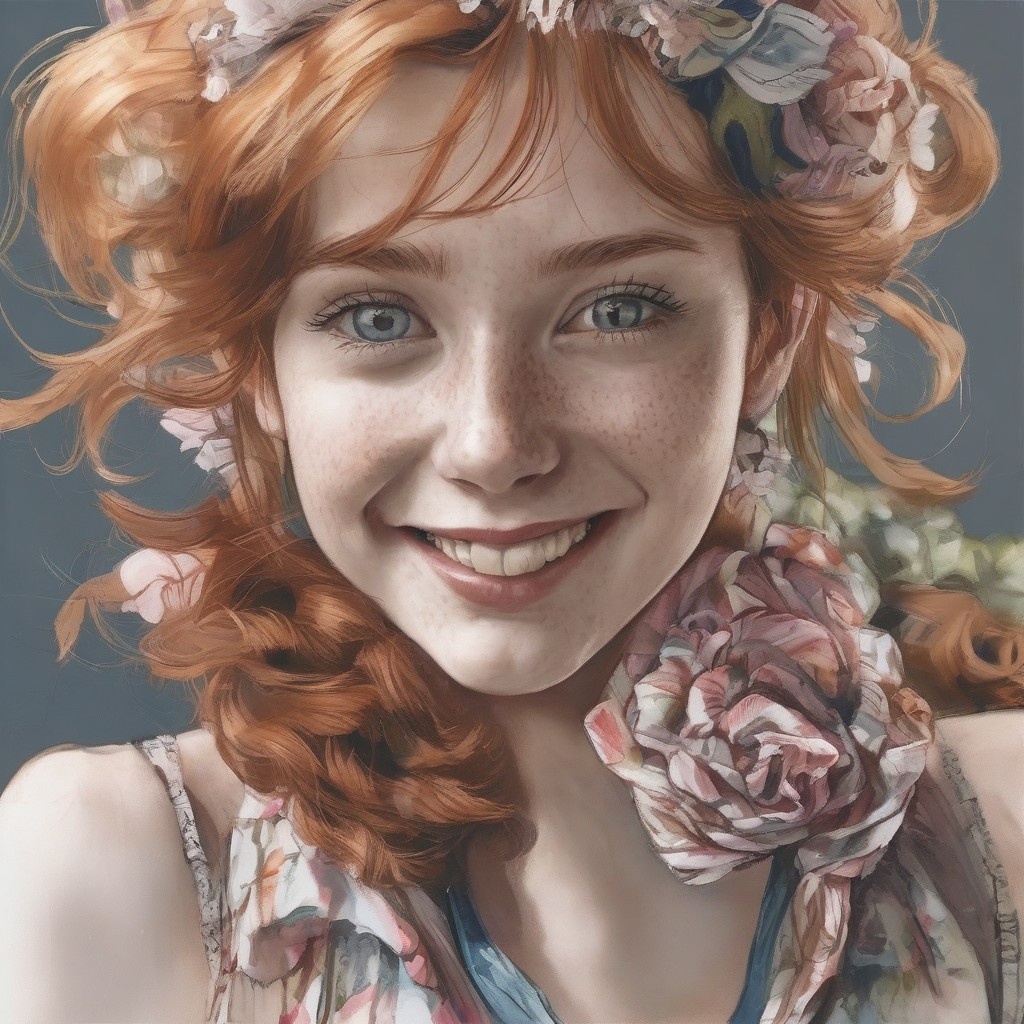} & 
        \includegraphics[width=0.15\linewidth]{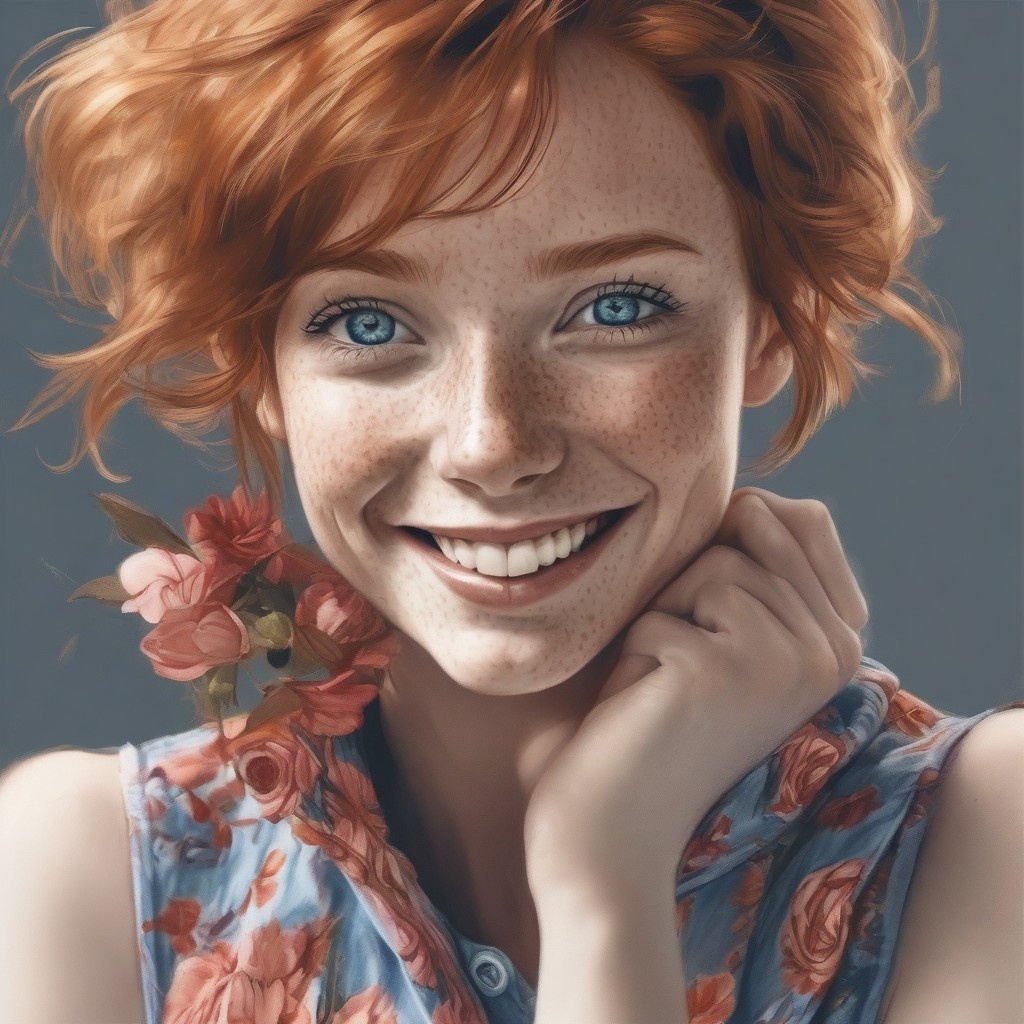} & 
        \includegraphics[width=0.15\linewidth]{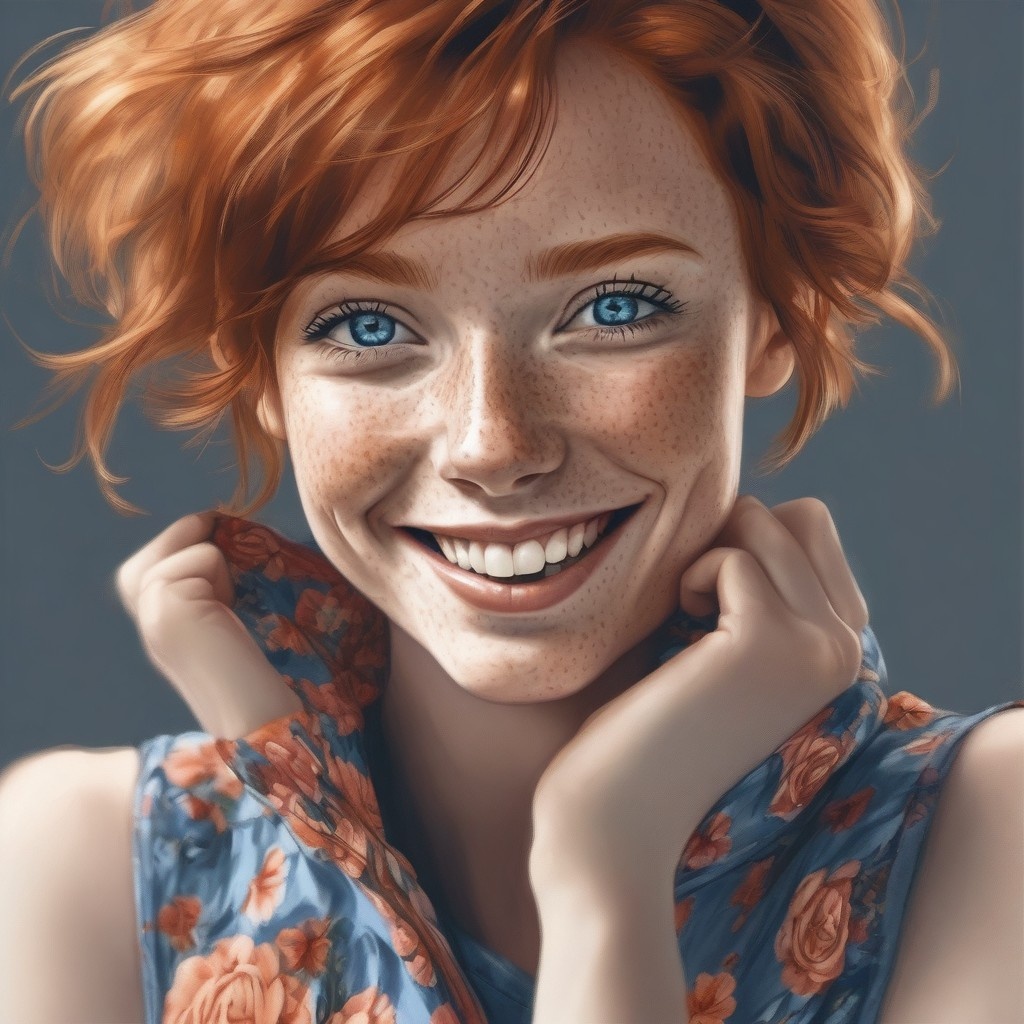} & 
        \includegraphics[width=0.15\linewidth]{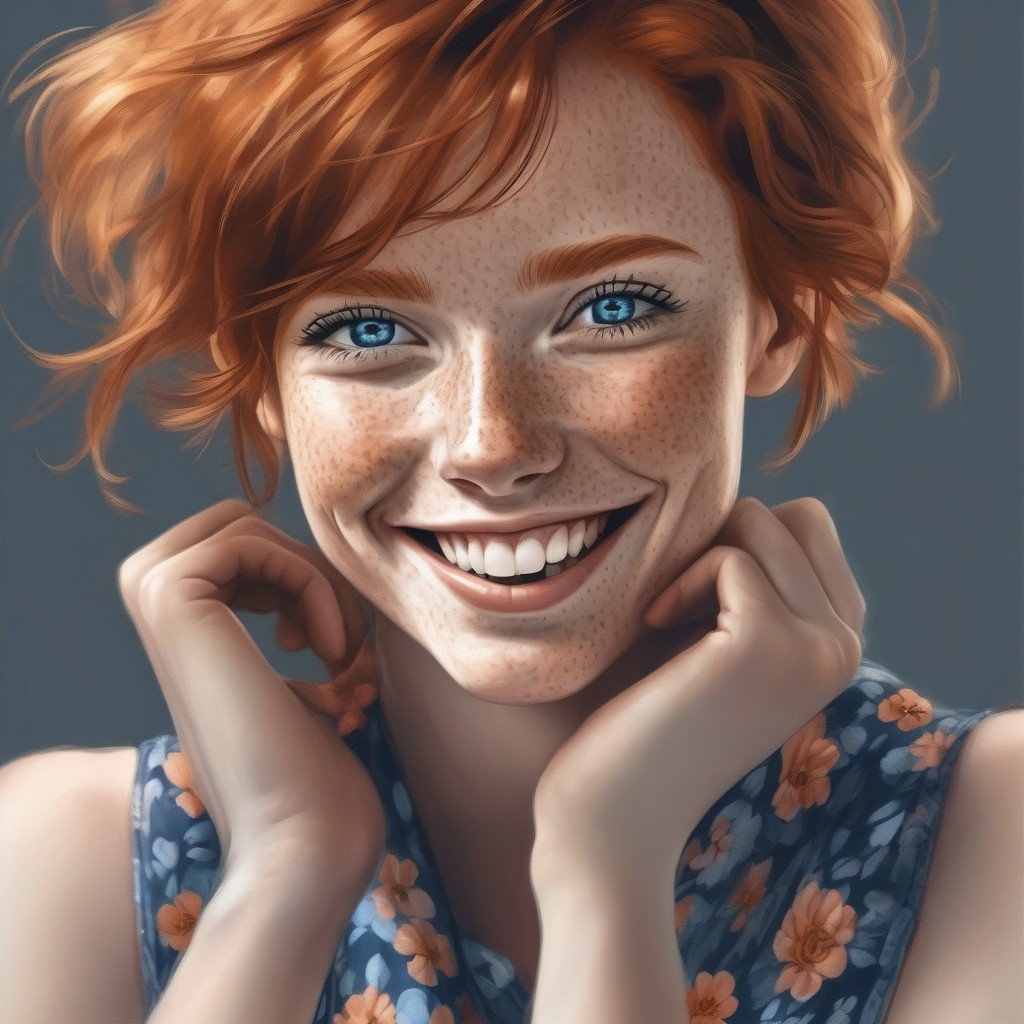} &
        \includegraphics[width=0.15\linewidth]{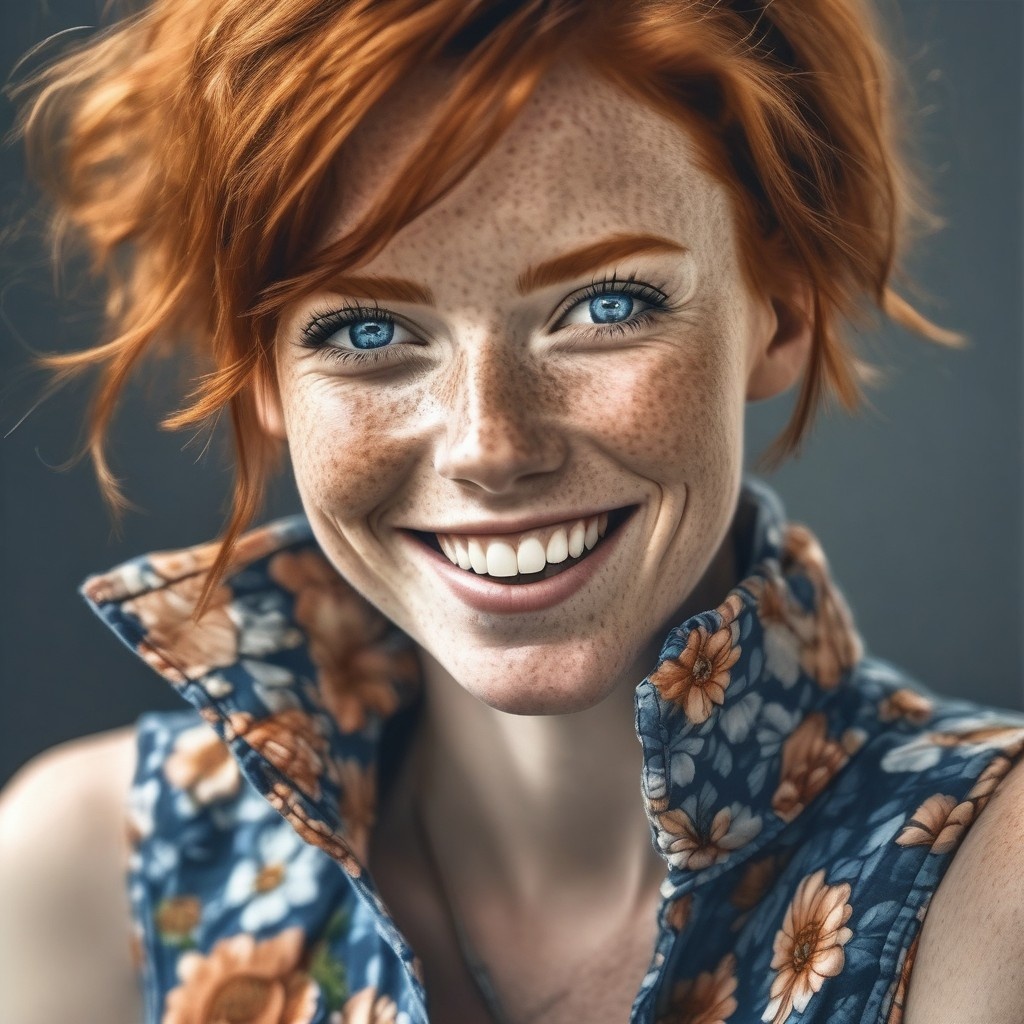} \\
        \includegraphics[width=0.15\linewidth]{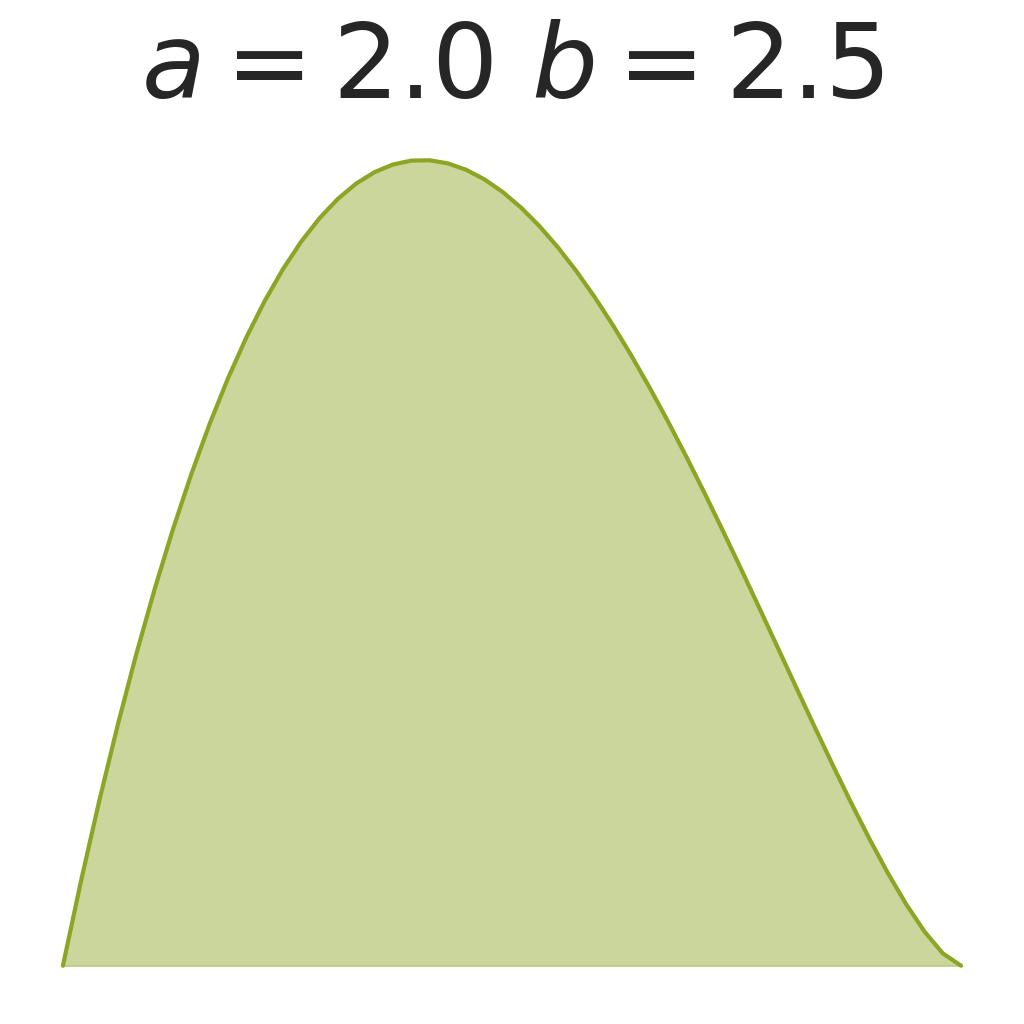} & 
        \includegraphics[width=0.15\linewidth]{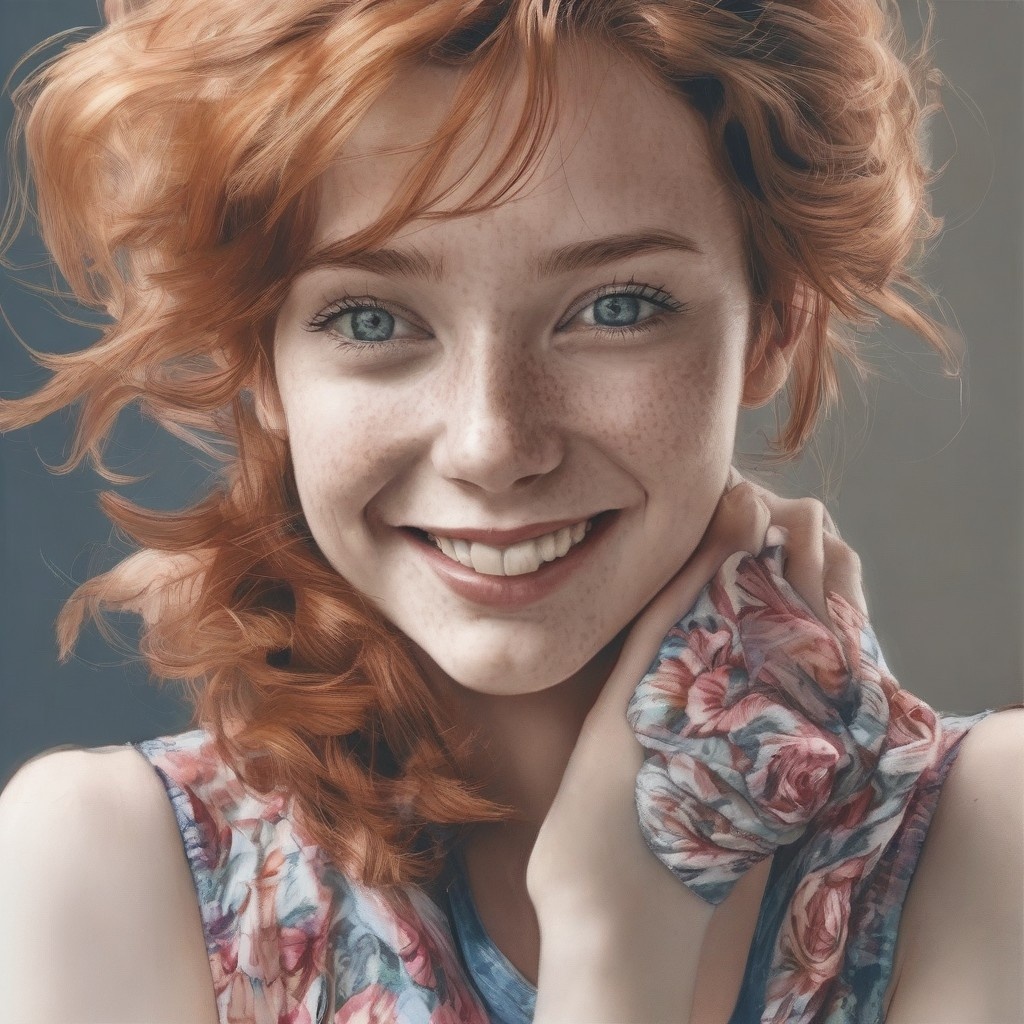} & 
        \includegraphics[width=0.15\linewidth]{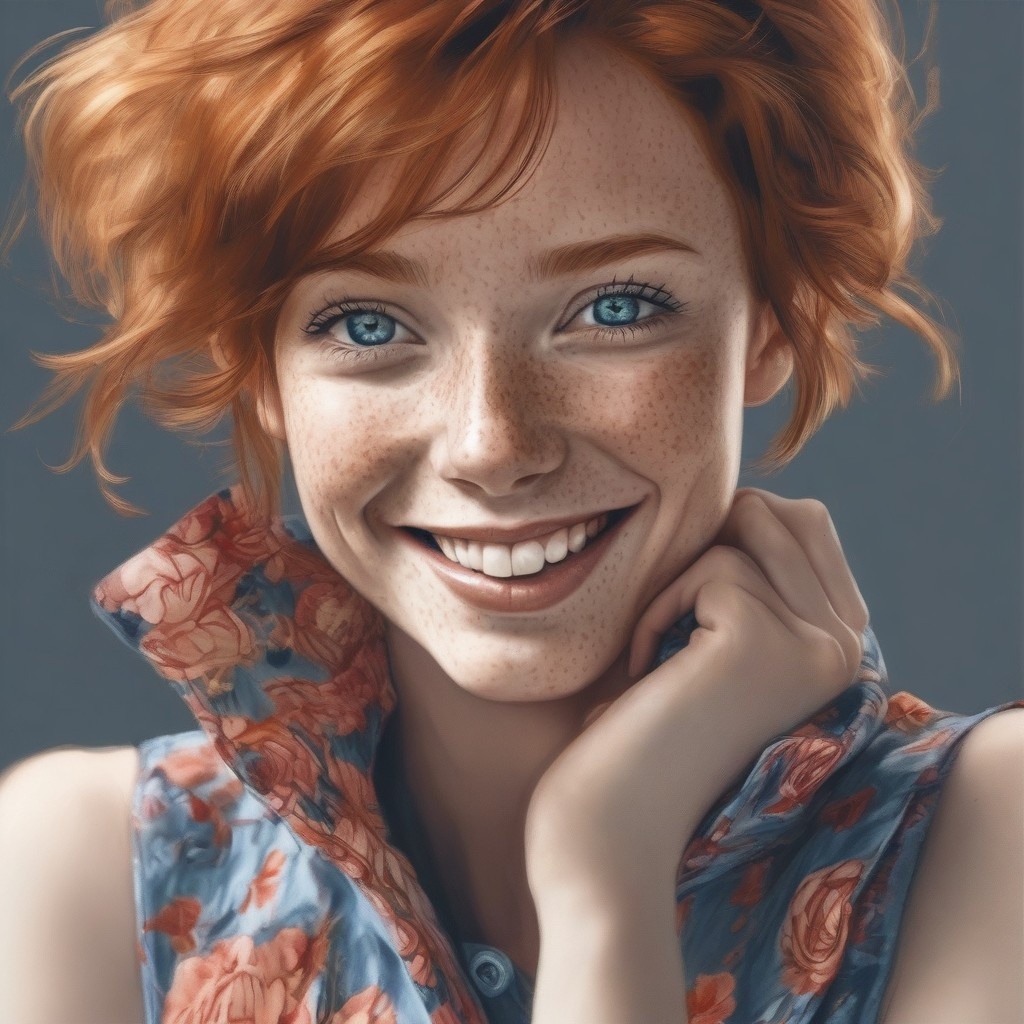} & 
        \includegraphics[width=0.15\linewidth]{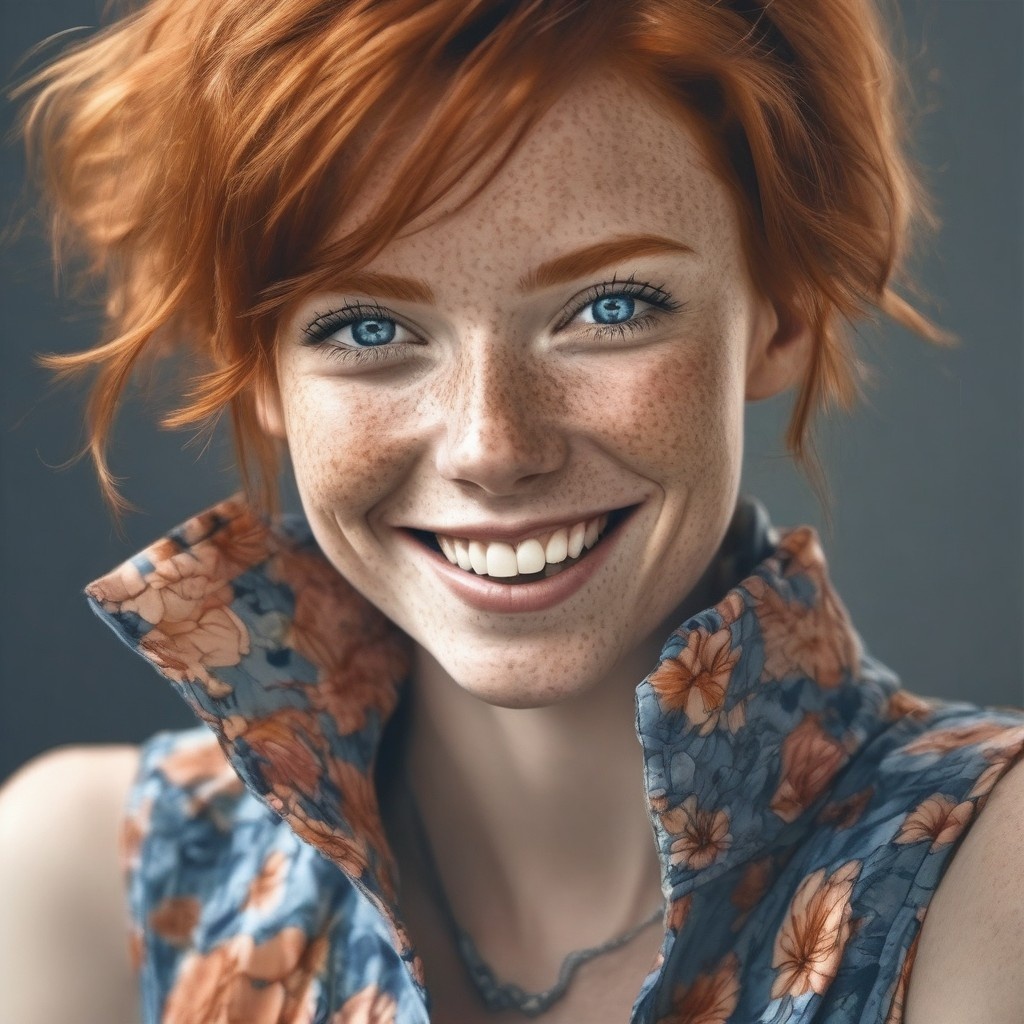} & 
        \includegraphics[width=0.15\linewidth]{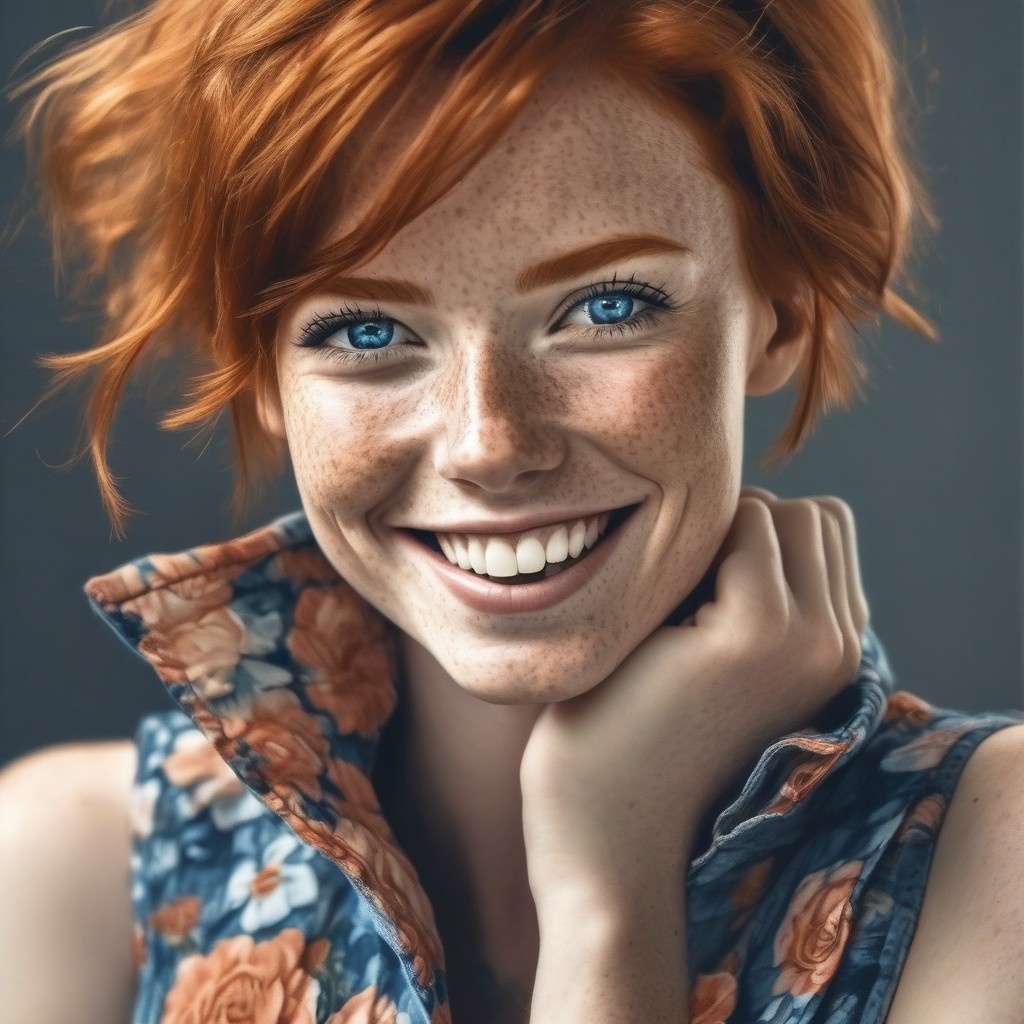} &
        \includegraphics[width=0.15\linewidth]{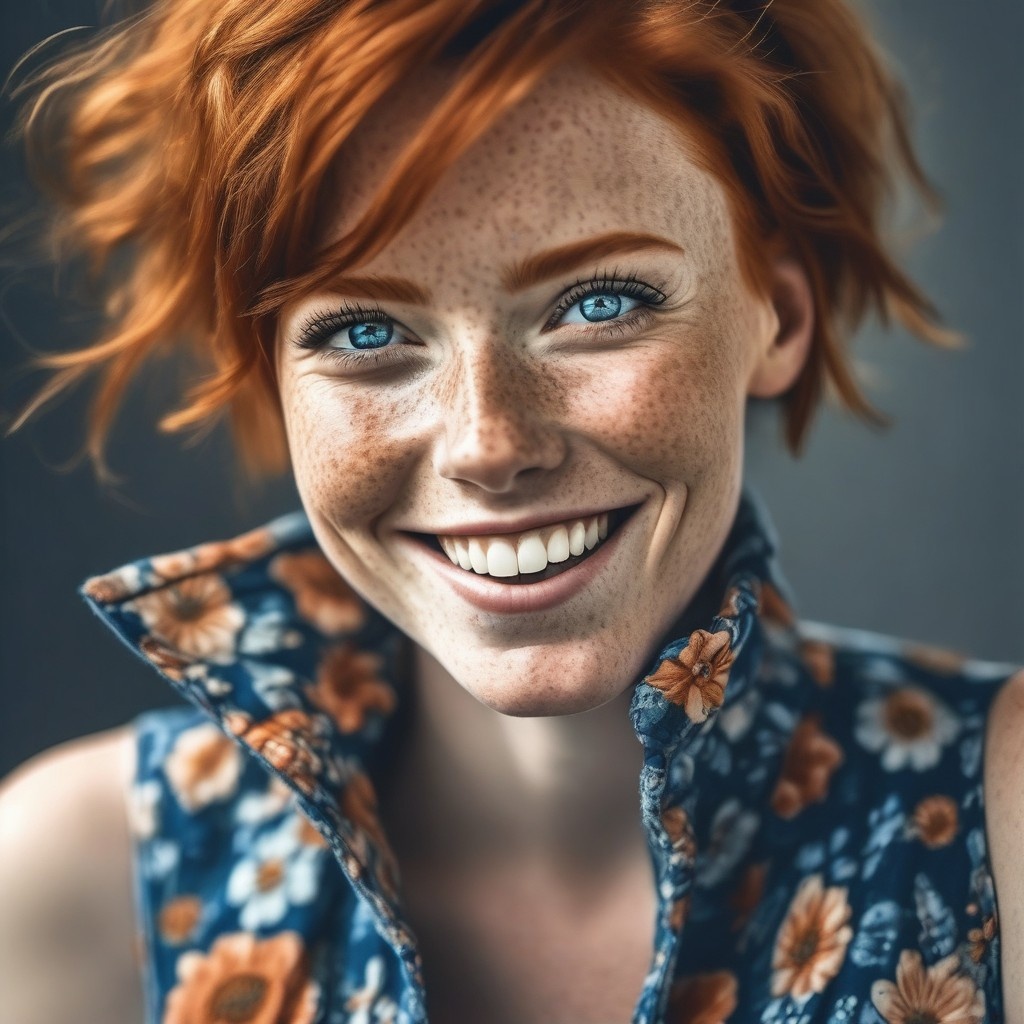} \\
        \includegraphics[width=0.15\linewidth]{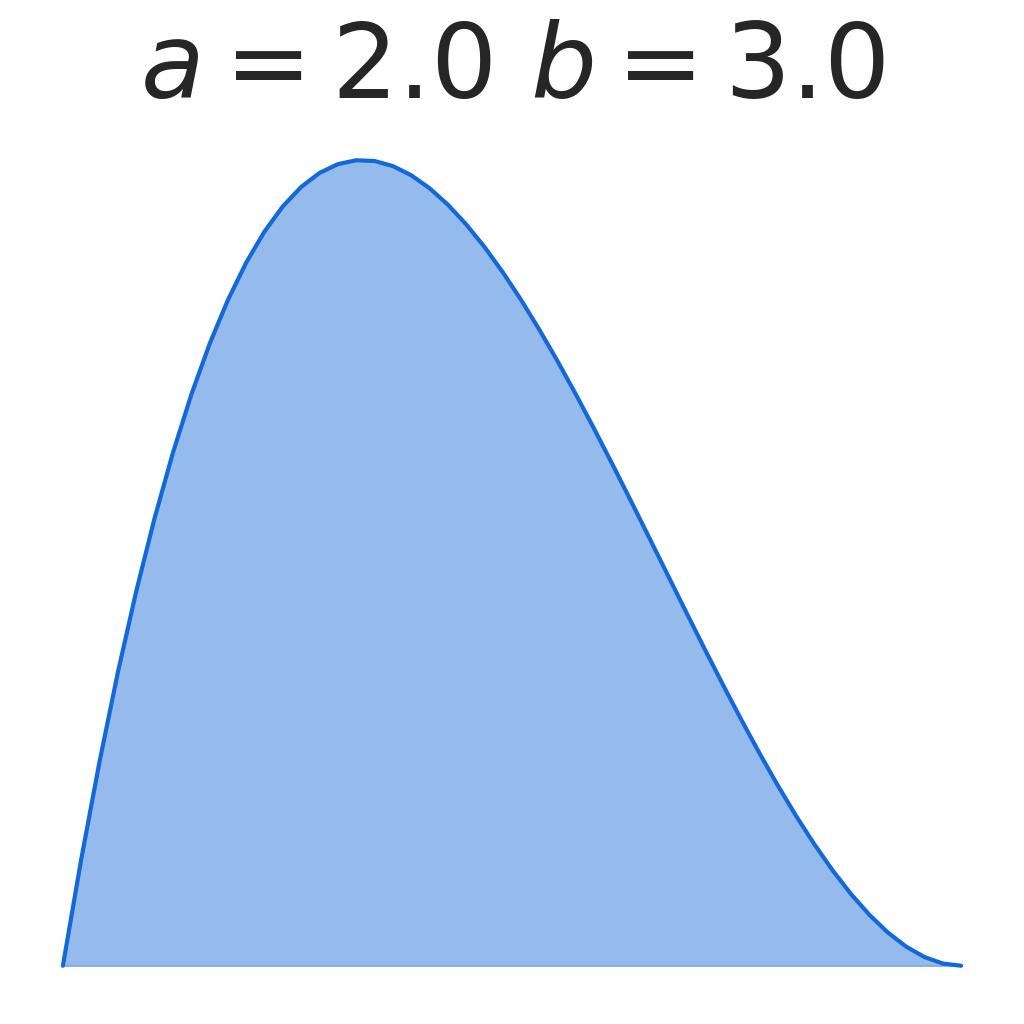} & 
        \includegraphics[width=0.15\linewidth]{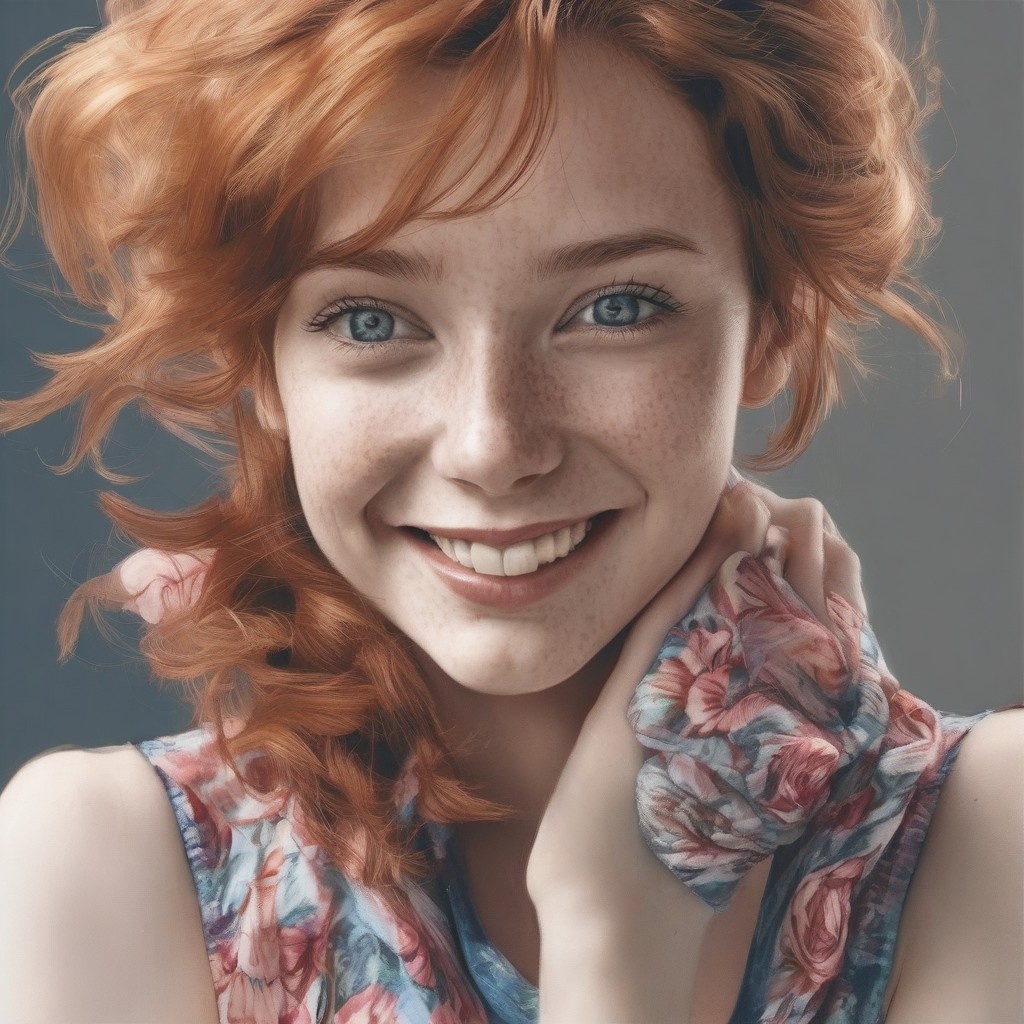} & 
        \includegraphics[width=0.15\linewidth]{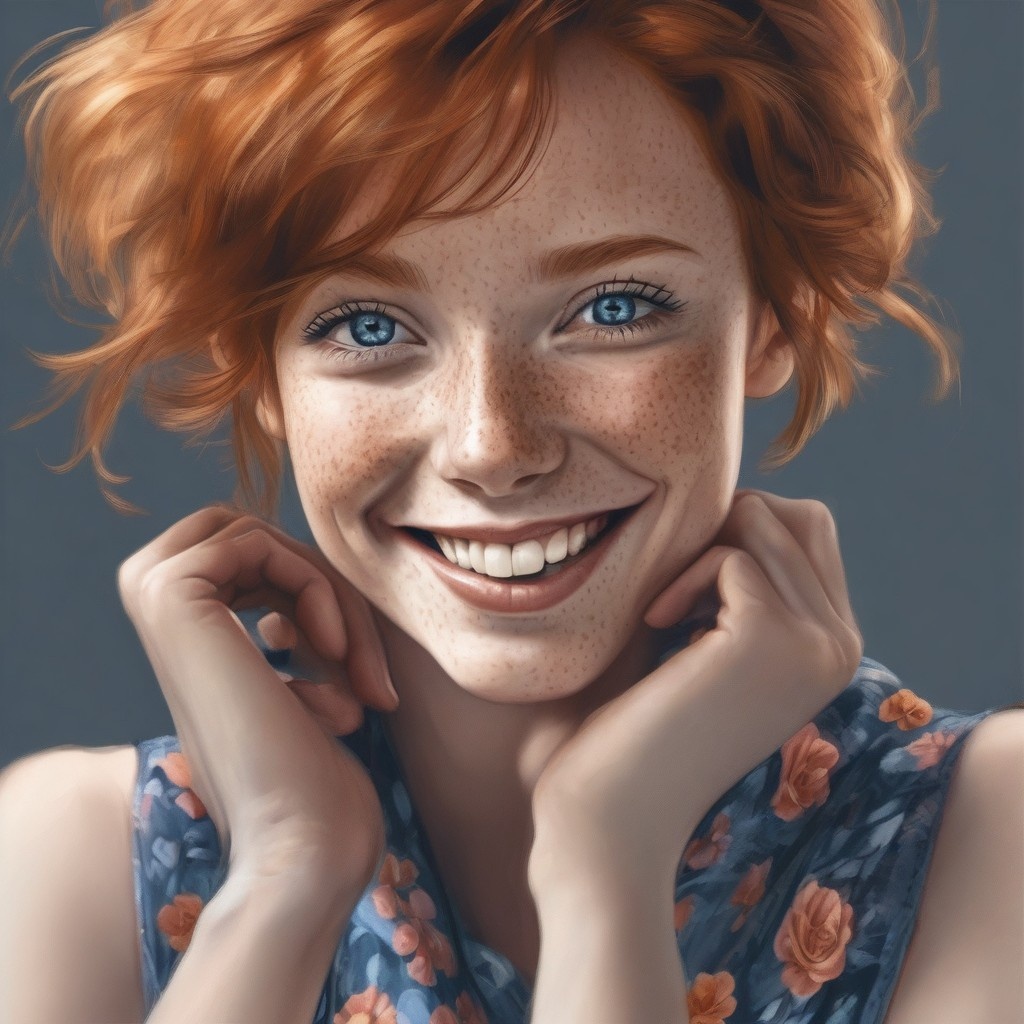} & 
        \includegraphics[width=0.15\linewidth]{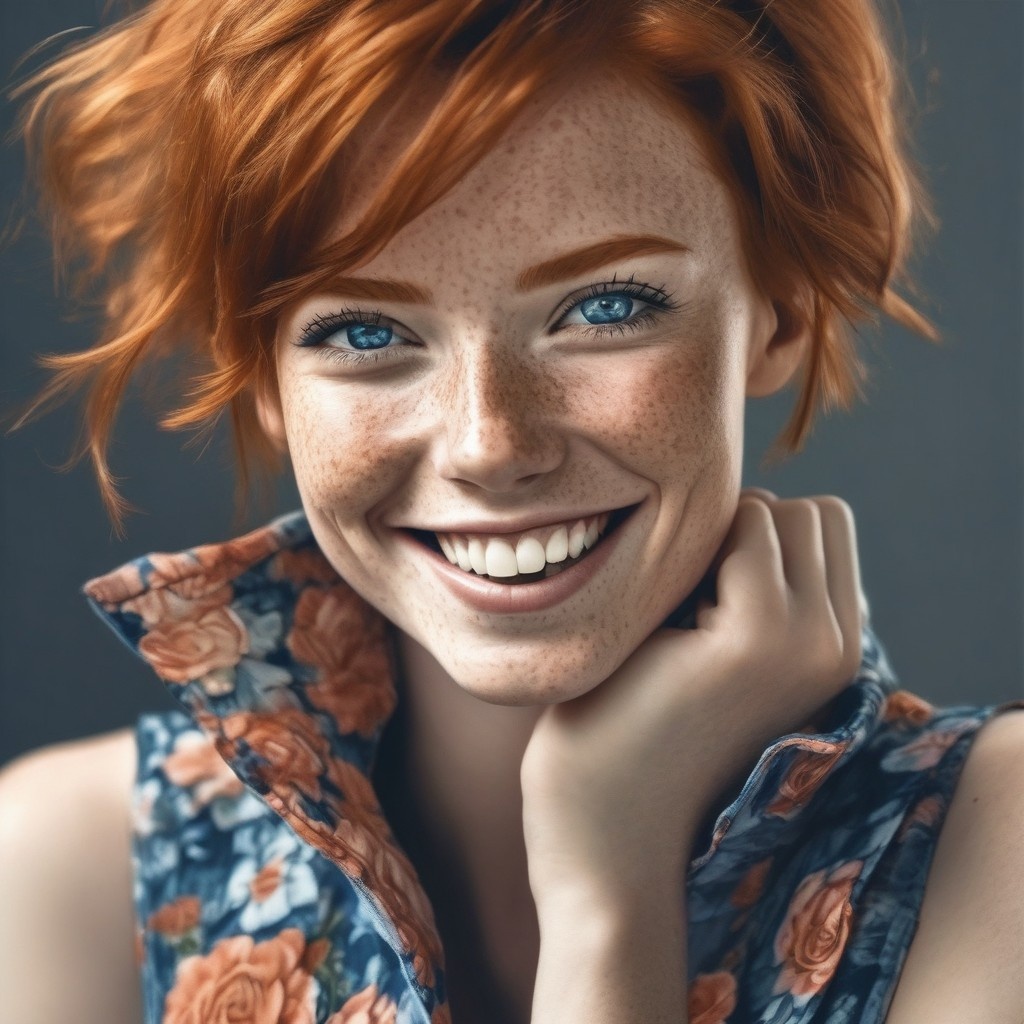} & 
        \includegraphics[width=0.15\linewidth]{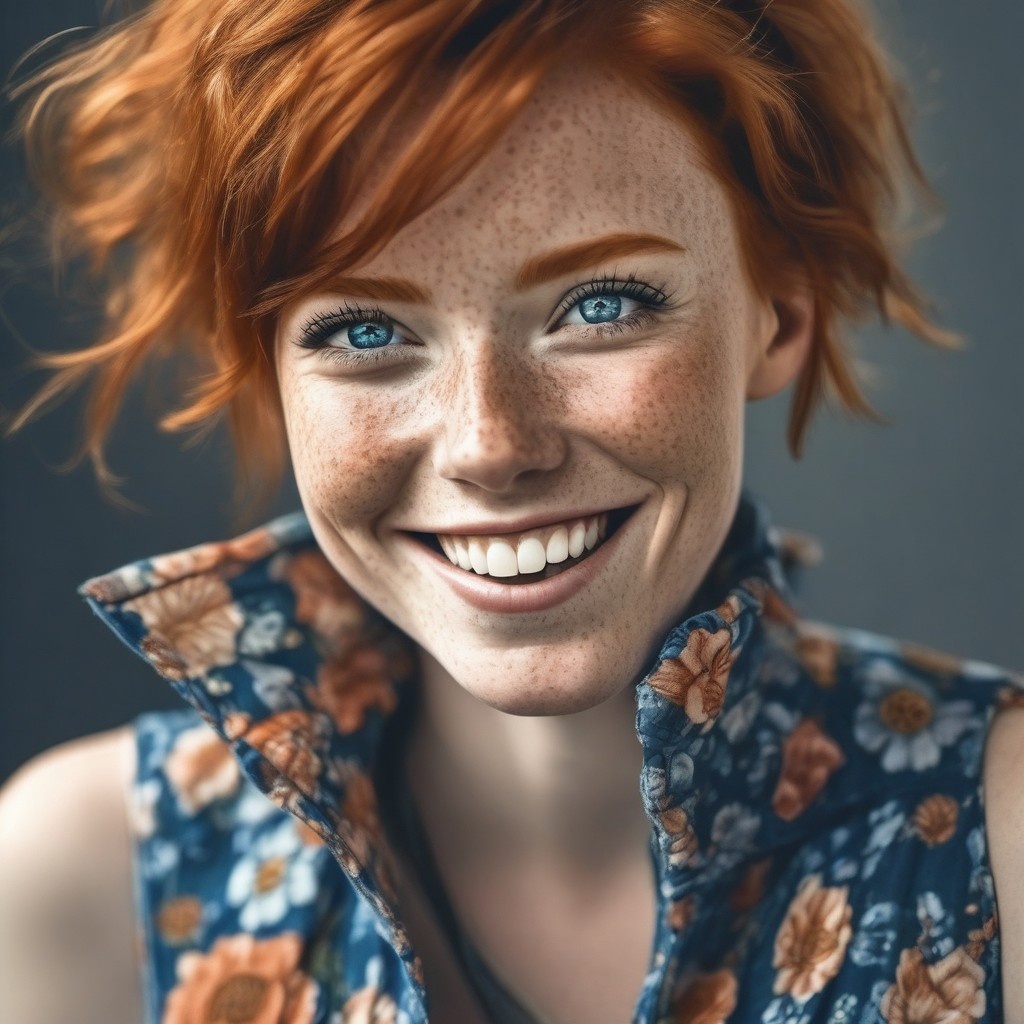} &
        \includegraphics[width=0.15\linewidth]{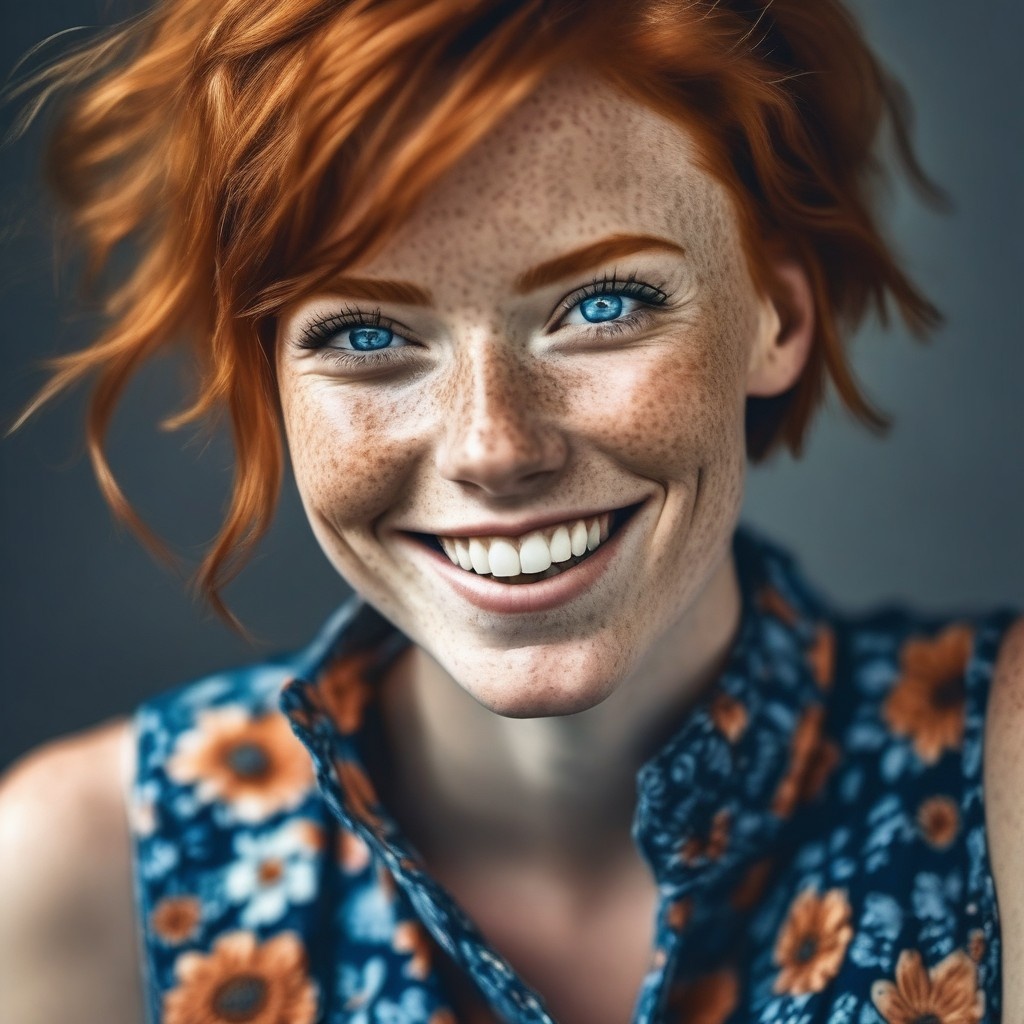} \\

        \includegraphics[width=0.15\linewidth]{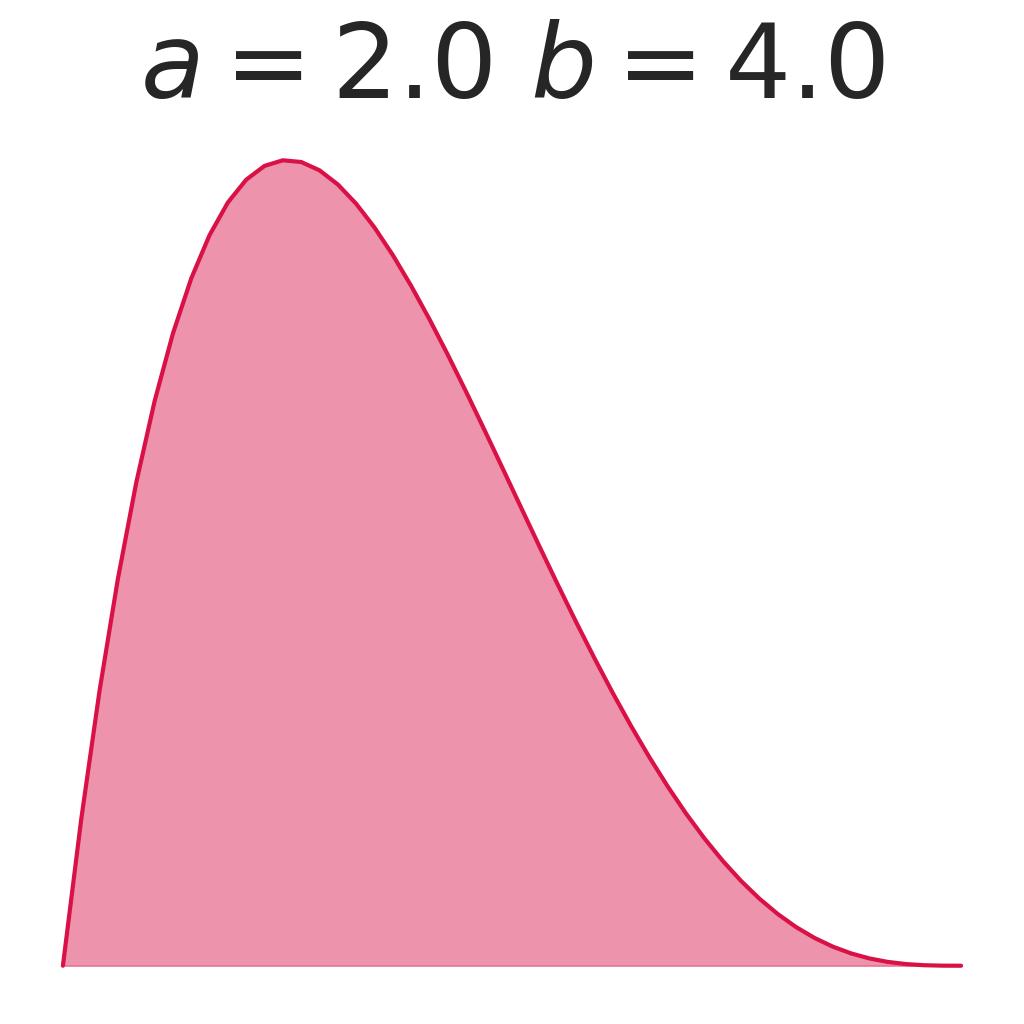} & 
        \includegraphics[width=0.15\linewidth]{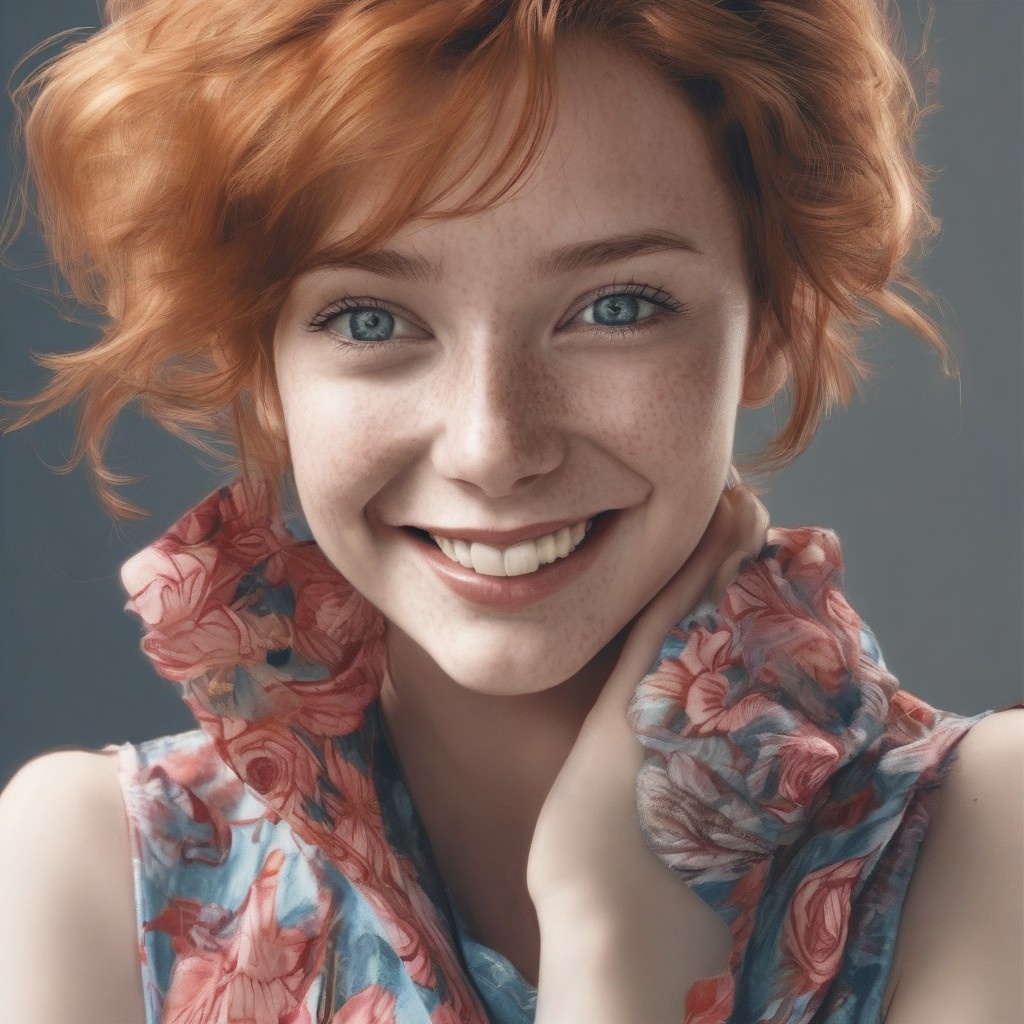} & 
        \includegraphics[width=0.15\linewidth]{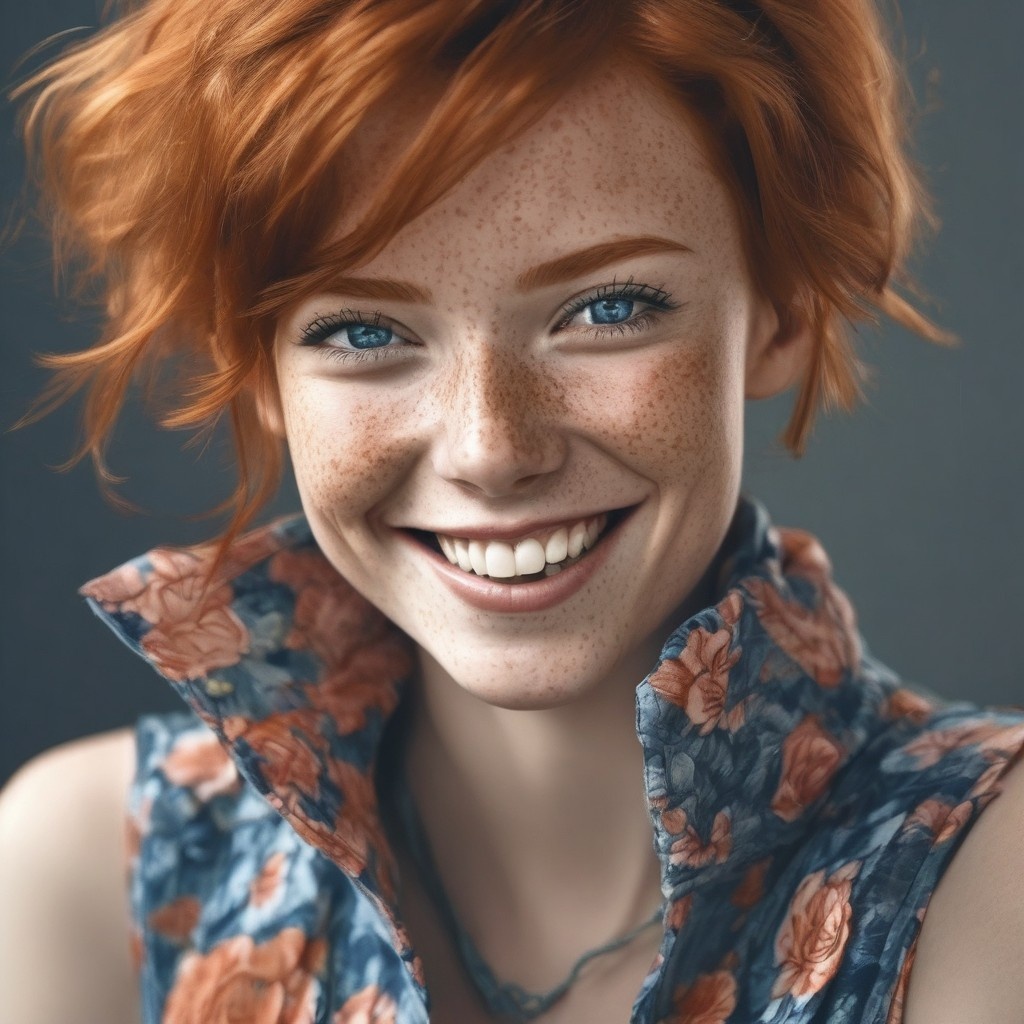} & 
        \includegraphics[width=0.15\linewidth]{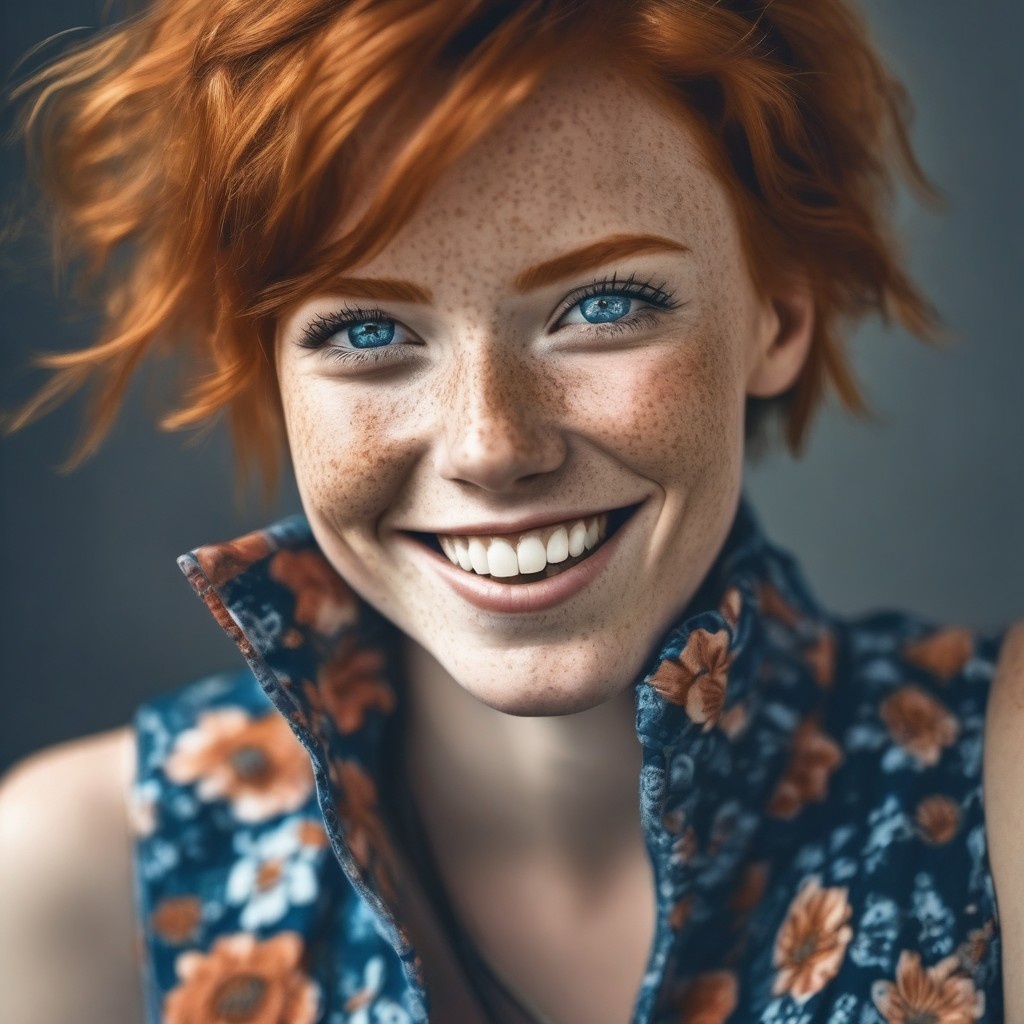} & 
        \includegraphics[width=0.15\linewidth]{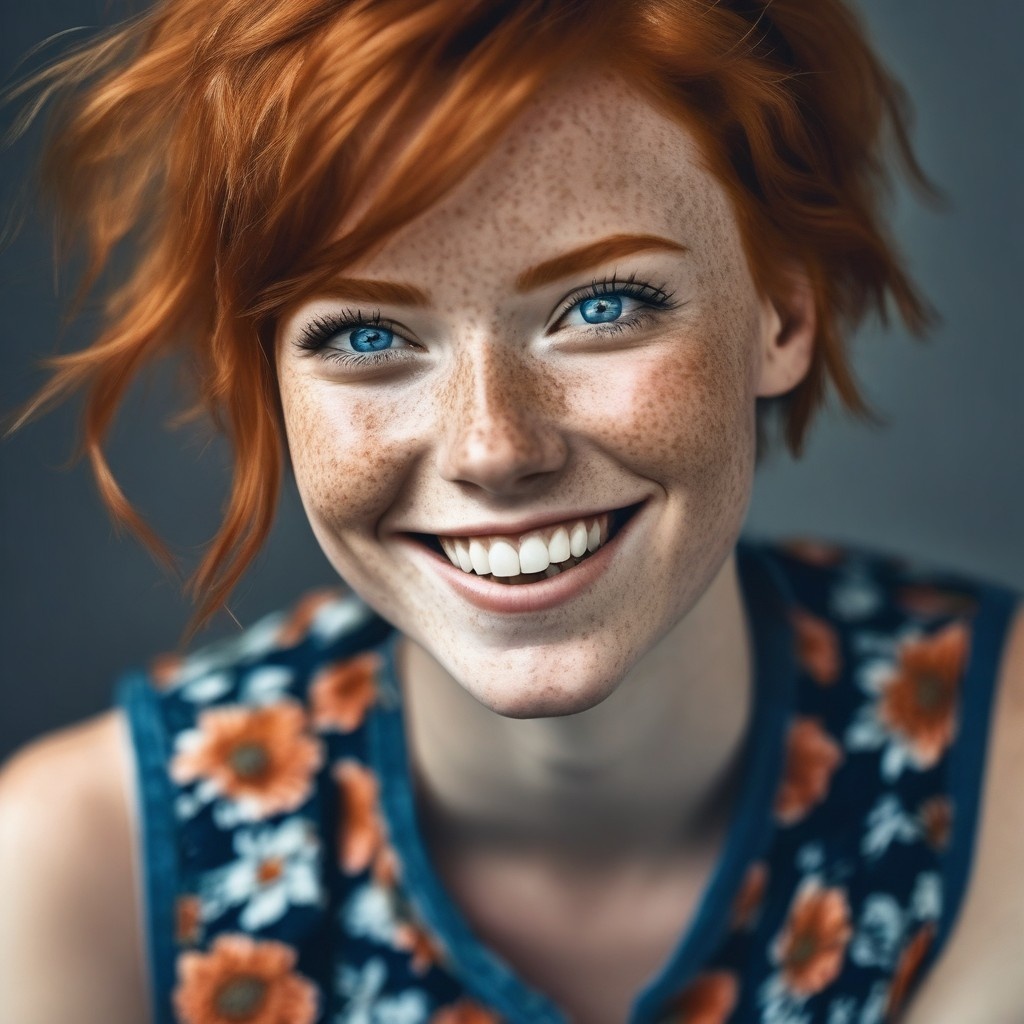} &
        \includegraphics[width=0.15\linewidth]{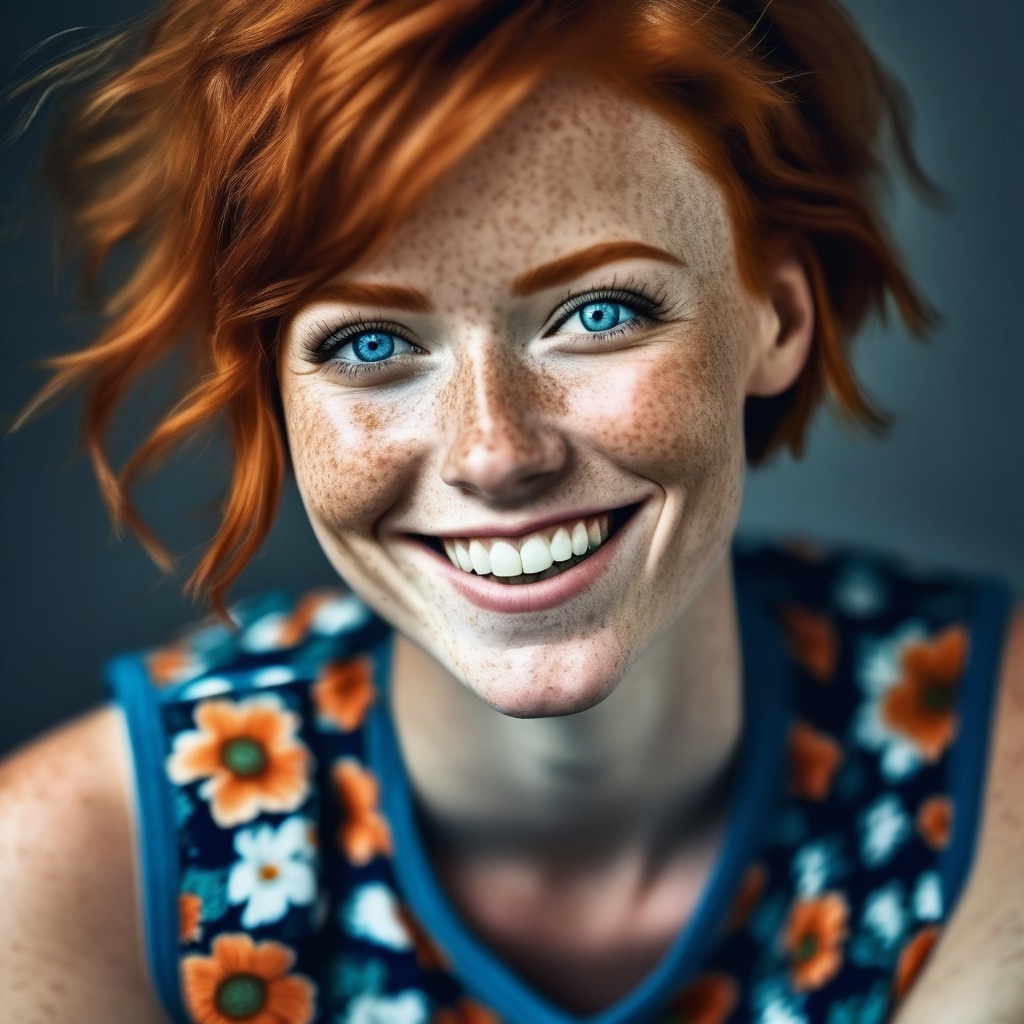} \\[0.2cm]
        \our{}&
        $\omega = 2.0$ & 
        $\omega = 5.0$ &
        $\omega = 7.5$ &
        $\omega = 9.0$ &
        $\omega = 12.5$ \\
        
    \end{tabular}
    \caption{Ablation study of our models on data generated for the prompt: "beautiful lady, freckles, big smile, blue eyes, short ginger hair, wearing a floral blue vest top, soft light, dark gray background." Thanks to the $\beta$-distribution, we can model how the diffusion trajectory behaves near data manifolds. }
    \label{fig:sampling}
\end{figure*}

\paragraph{Classifier GeoGuide.} In \cite{poleski2024geoguide}, the authors propose to modify the classifier guidance with gradient-based normalization to control updates. The process of guiding the diffusion model uses fixed-length updates to force the denoising process to be as close as possible to the data manifold:
\begin{equation}
    \hat{\epsilon}(x_t) = \epsilon(x_t) - w \frac{\sqrt{D}}{T} \frac{\nabla_{x_t} p(y \vert x_t)}{||\nabla_{x_t} p(y \vert x_t)||},
\end{equation}  
where $D$ is data dimensionality and $T$ is the number of diffusion steps.  

\paragraph{Classifier-Free Guidance.}
Classifier-Free Guidance (CFG) \citep{ho2022classifier} is a technique employed in diffusion models to enhance control over the generative process without needing external classifiers. It has shown significant effectiveness in boosting the quality of generated outputs across tasks like image and text generation.

CFG requires access to the additional conditional generative model, for which the conditioning factor $c$ is incorporated as additional input to the denoising component, $\e_c(x_t)$. CFG guides generation by combining conditional and unconditional predictions. For a noisy sample $x_t$, this guidance is implemented by interpolating between these conditional and unconditional predictions as follows:
\begin{align}
    \label{eq:cfg}
    \hat{\epsilon}_c^{w}(x_t) = \epsilon_{\o}(x_t) + w \left( \epsilon_c(x_t) - \epsilon_{ \o}(x_t) \right),    
\end{align}
where $\epsilon_{\o}(x_t)$ represents the model’s prediction of the noise for $x_t$ in unconditional case, while $\epsilon_c(x_t)$ denotes the noise prediction when conditioned on $c$. The parameter $w$ serves as the guidance scale, adjusting the extent to which the conditional information $y$ influences the generated output. The procedure of the reverse diffusion sampling process is given by Algorithm \ref{alg:cfg}. 

Adjusting $w$ allows control over the balance between sample diversity and consistency to the conditioning $y$. Setting $w = 1$ results in standard conditional generation. When $w > 1$, the influence of the conditioning information is amplified, encouraging the model to generate samples that align more closely with $y$, though this may reduce diversity.

\begin{algorithm}[t]
    \caption{Reverse Diffusion with CFG}
	\includegraphics[width=1.0\linewidth]{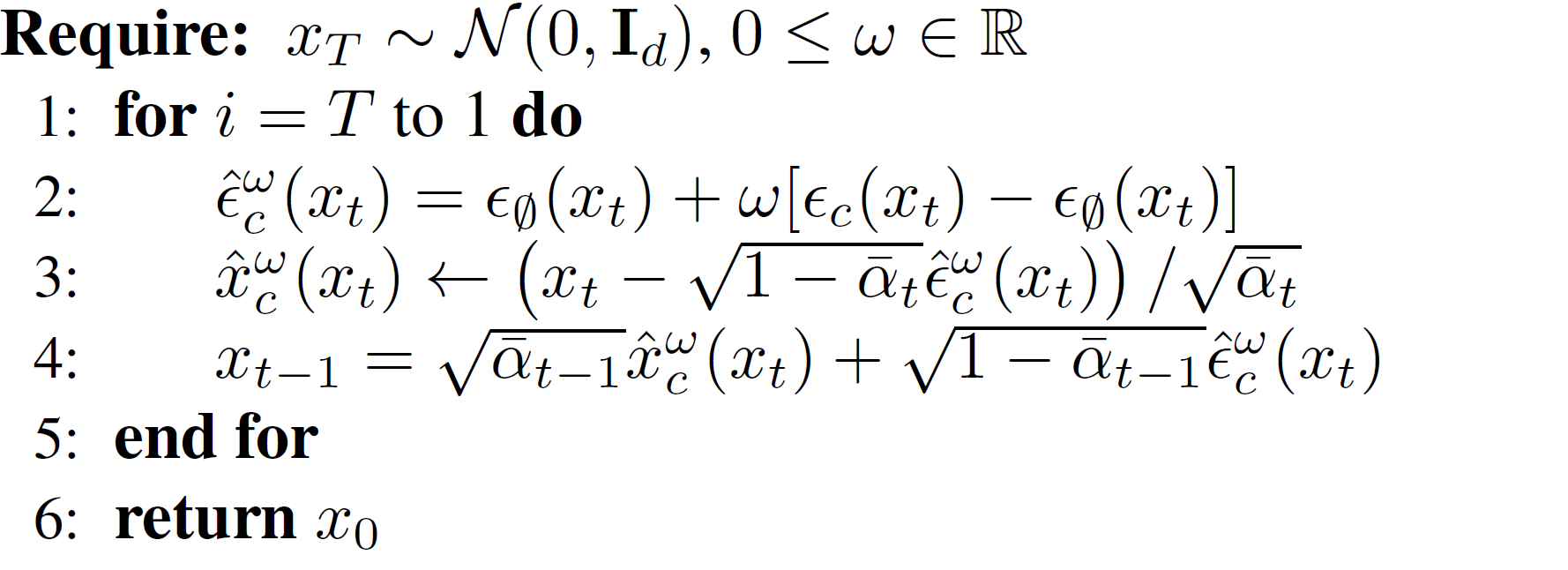}
        \label{alg:cfg}
\end{algorithm}

\begin{algorithm}[t]
    \caption{Reverse Diffusion with CFG++ }
        \label{alg_our}
\includegraphics[width=1.0\linewidth]{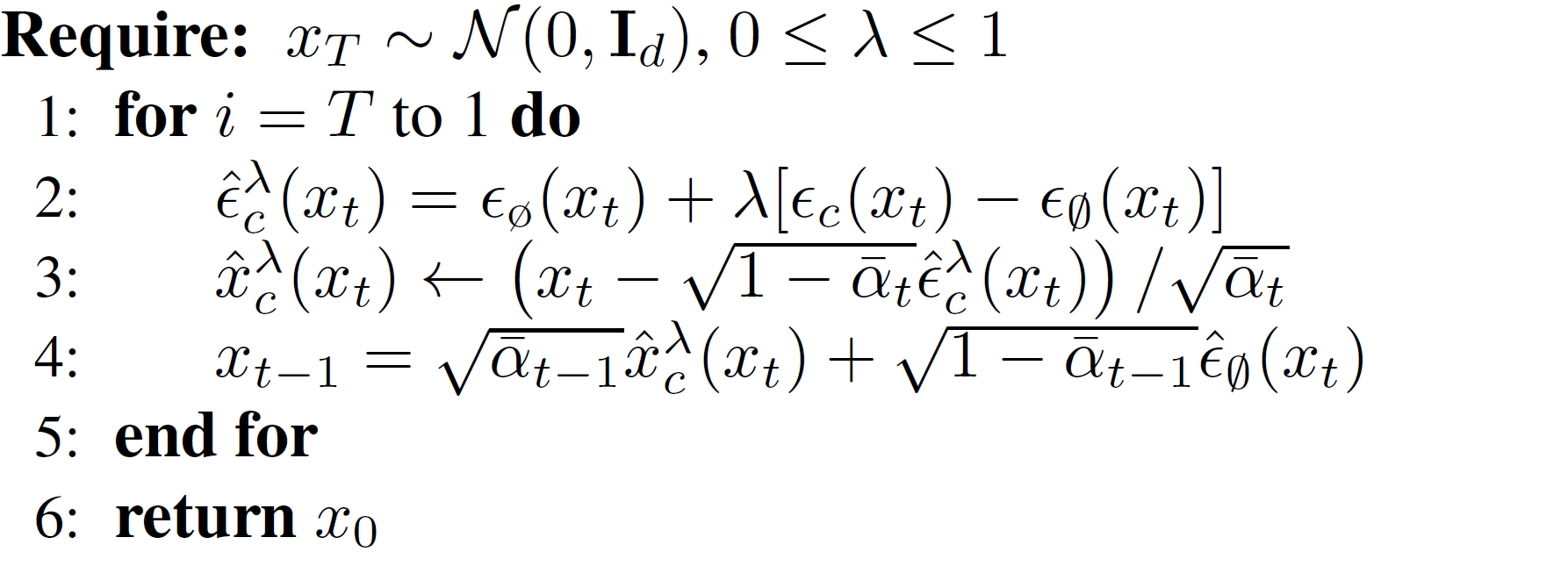}
        \label{alg:cfgplus}
\end{algorithm}
\begin{figure*}[ht]
    \centering
    \renewcommand{\arraystretch}{0}
    \setlength{\tabcolsep}{0pt}
    \begin{tabular}{c@{\;}ccccccc}
         &901 & 801 & 701 & 601 & 501 & 401 & 0\\[5pt]
        \rotatebox{90}{ \qquad CFG} &
        \includegraphics[width=0.13\linewidth]{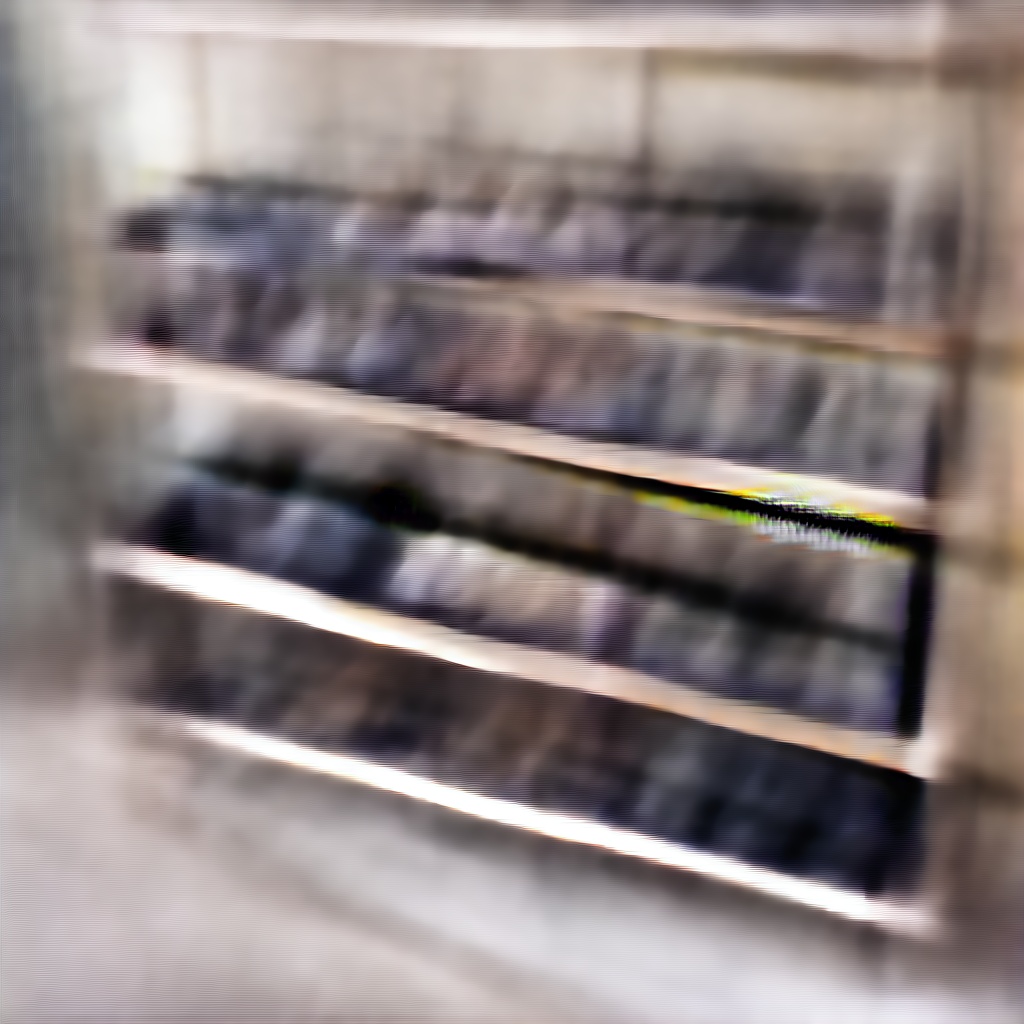} & 
        \includegraphics[width=0.13\linewidth]{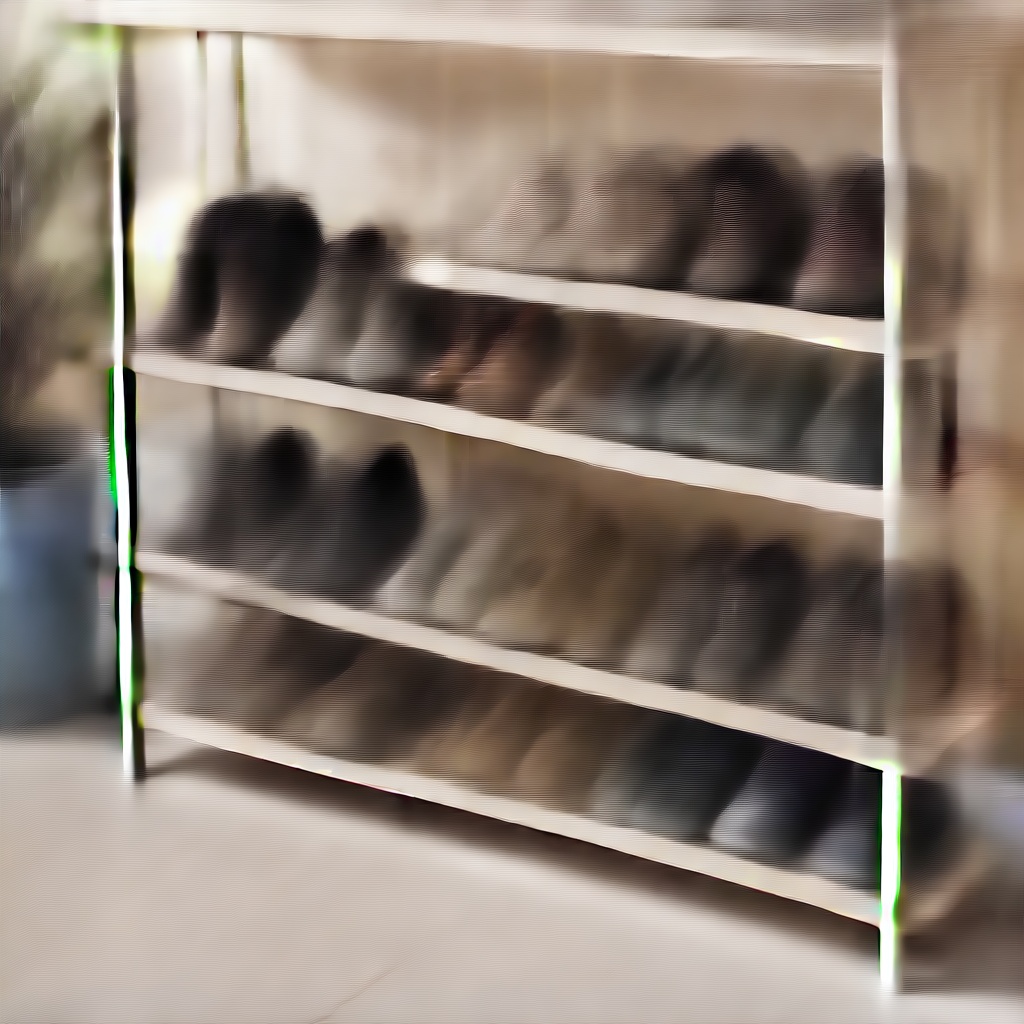} & 
        \includegraphics[width=0.13\linewidth]{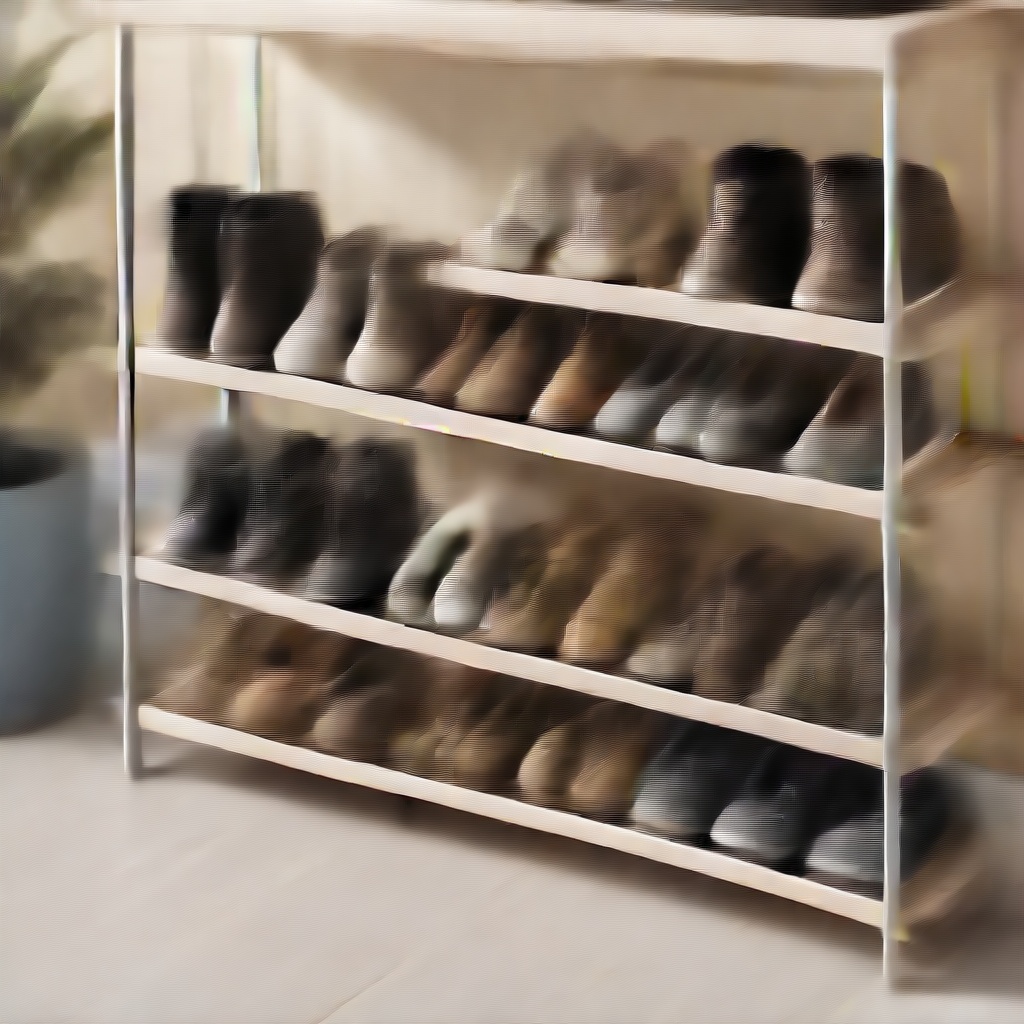} & 
        \includegraphics[width=0.13\linewidth]{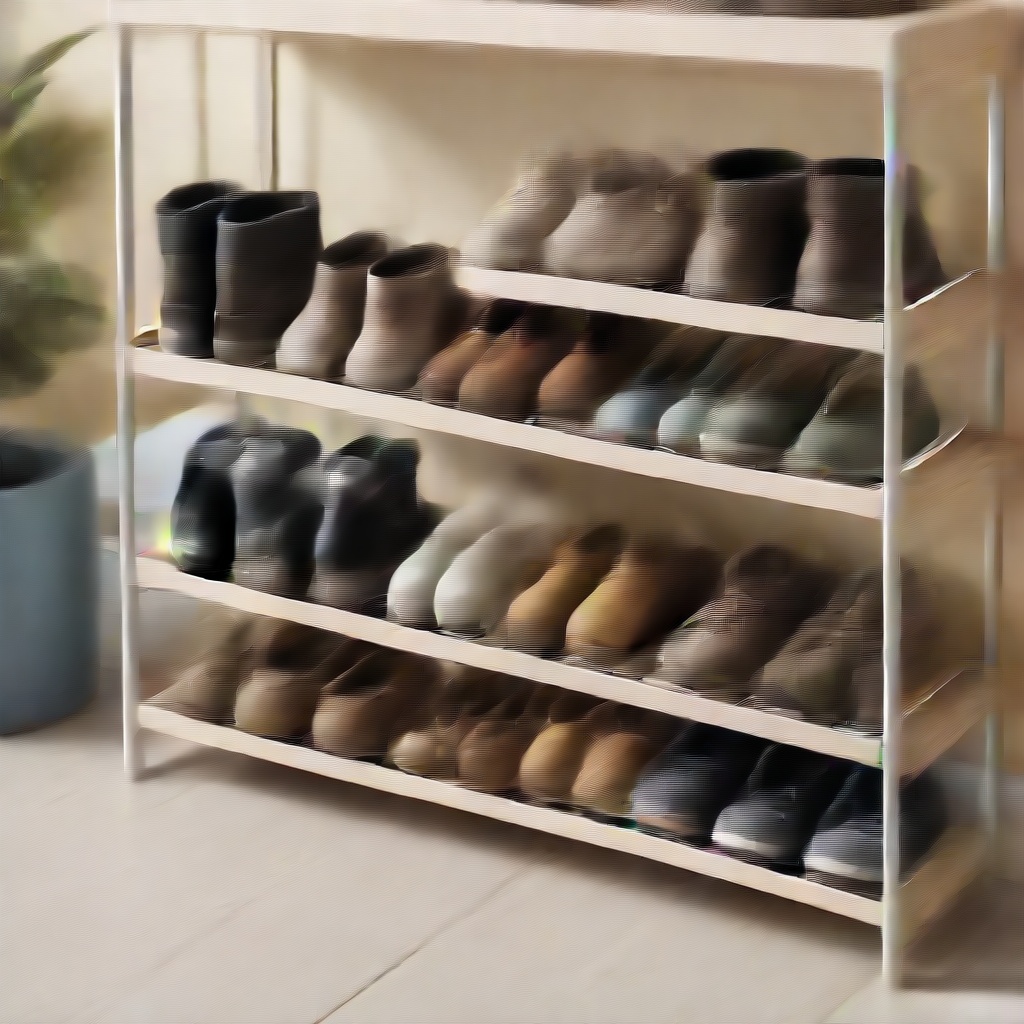} & 
        \includegraphics[width=0.13\linewidth]{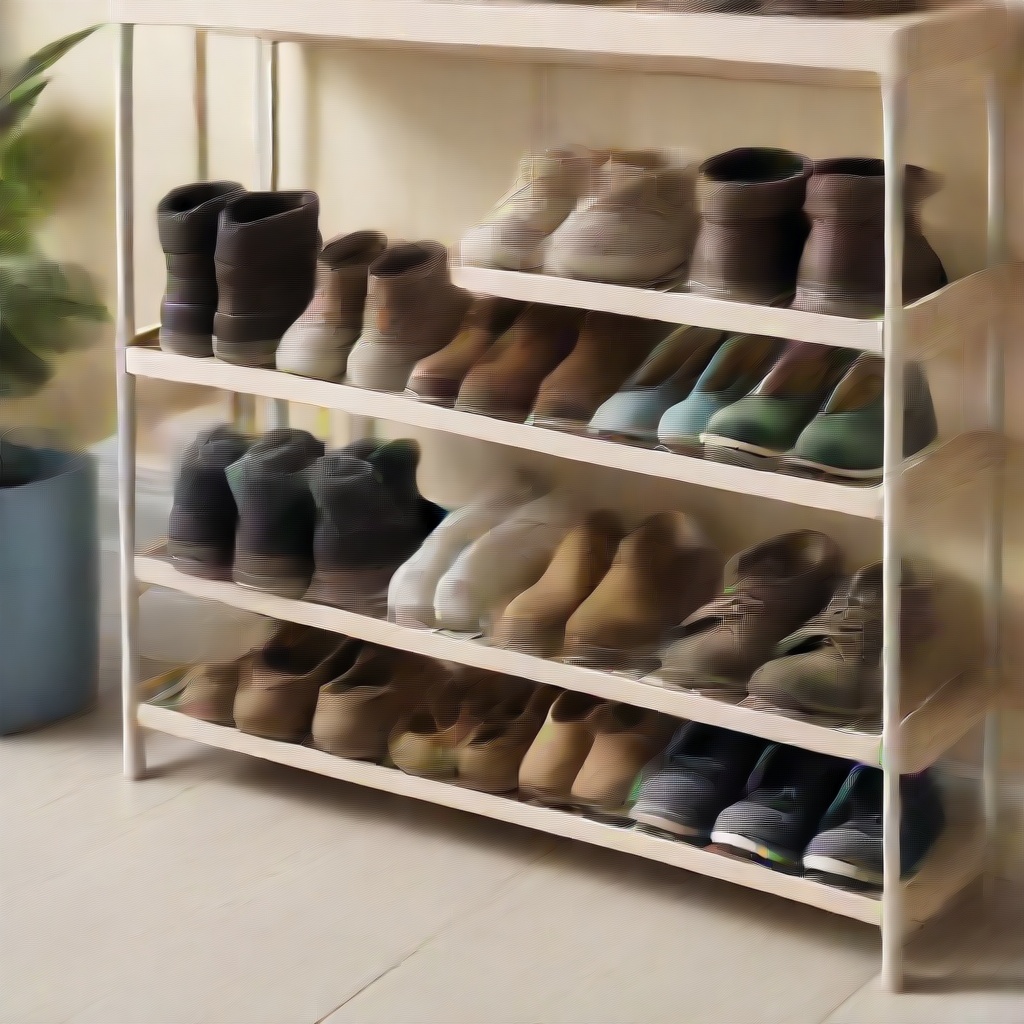} & 
        \includegraphics[width=0.13\linewidth]{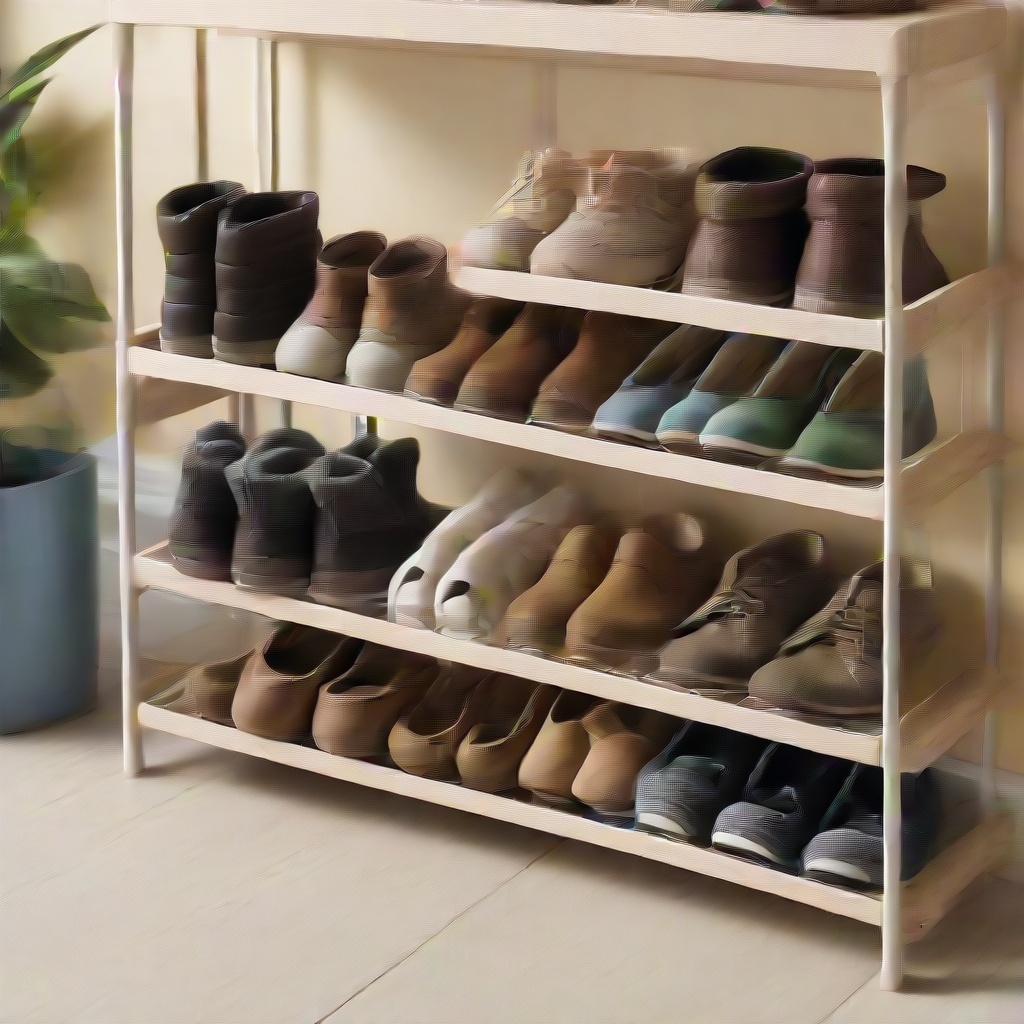} & 
        \includegraphics[width=0.13\linewidth]{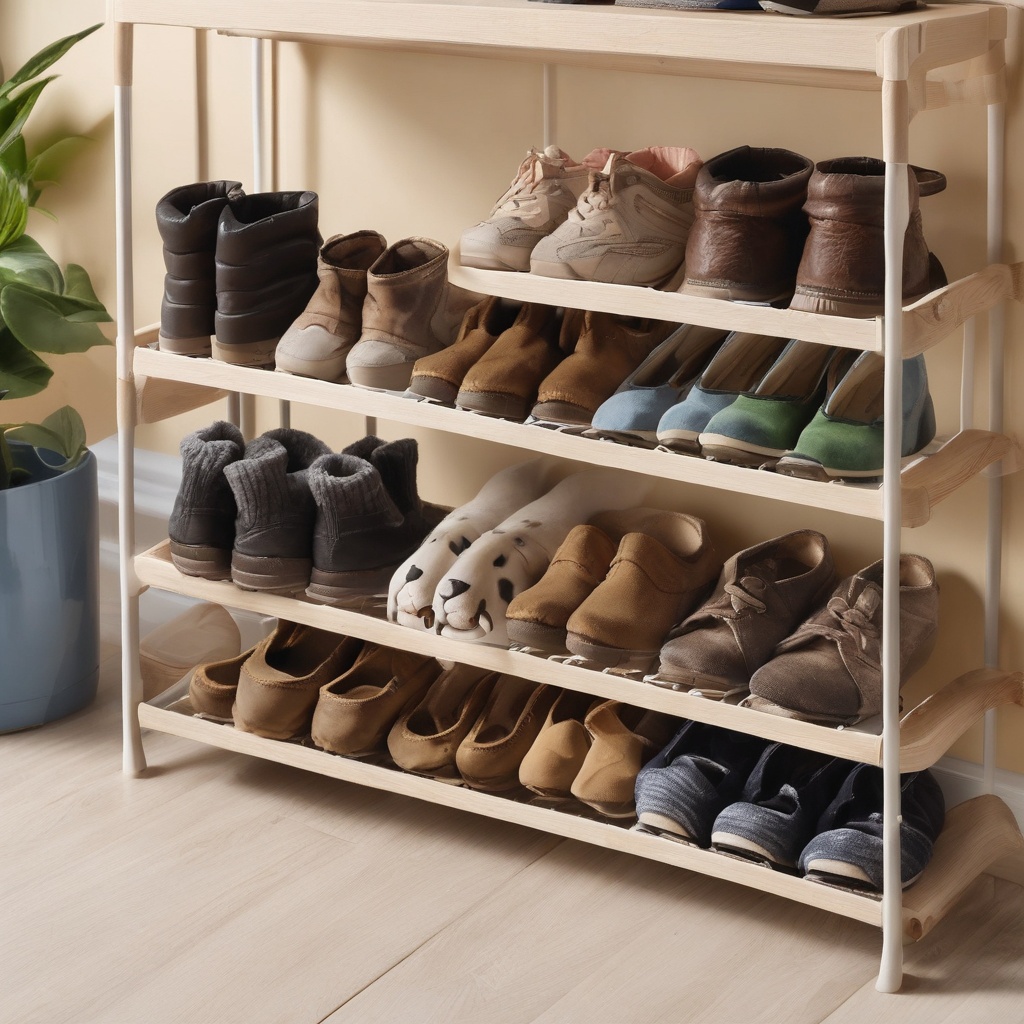} \\
        \rotatebox{90}{ \qquad \our{}} &
        \includegraphics[width=0.13\linewidth]{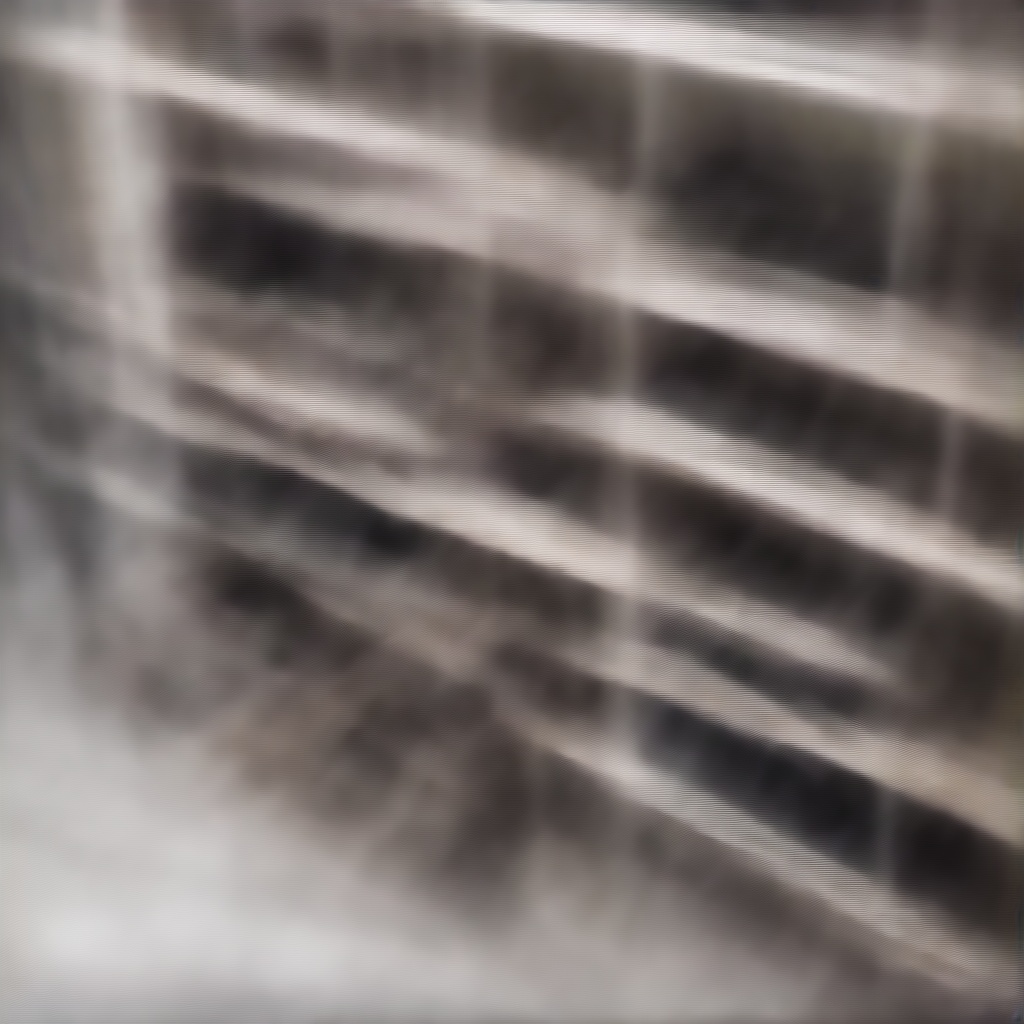} & 
        \includegraphics[width=0.13\linewidth]{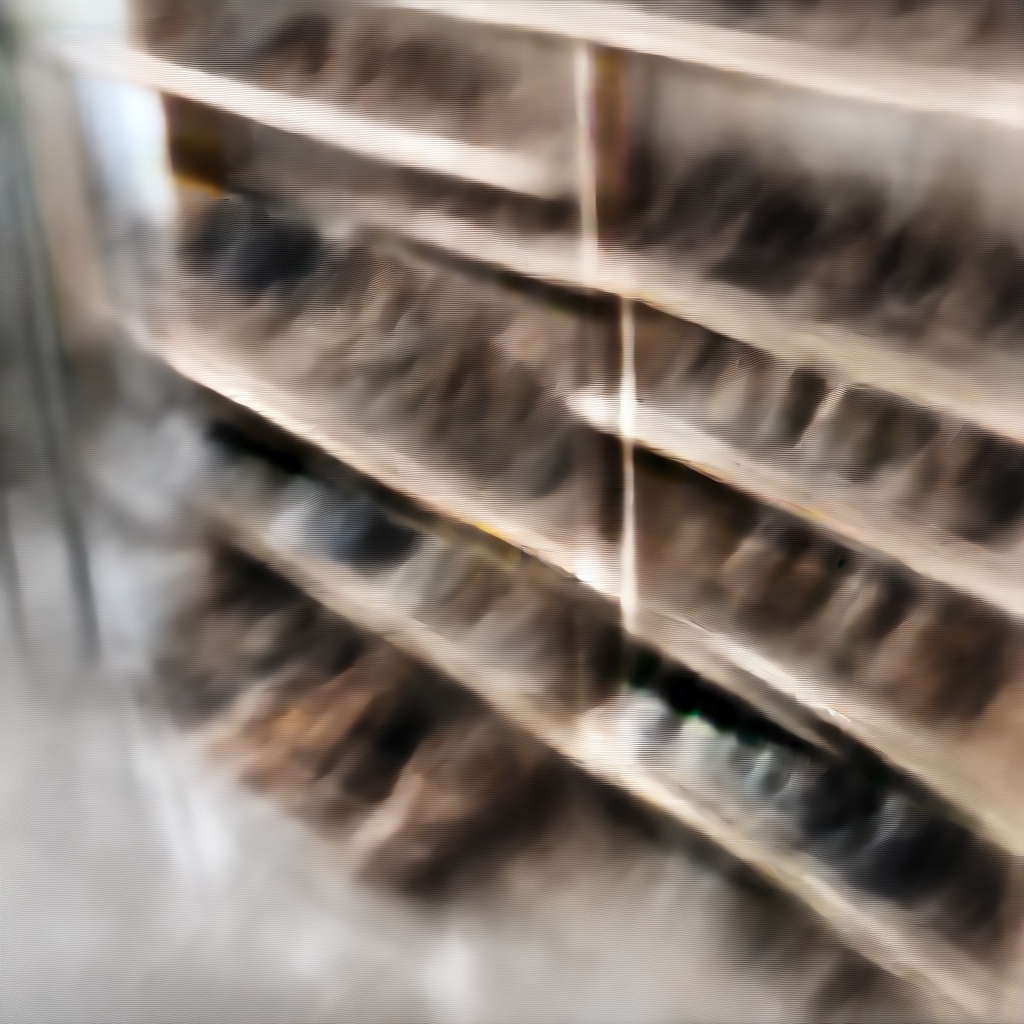} & 
        \includegraphics[width=0.13\linewidth]{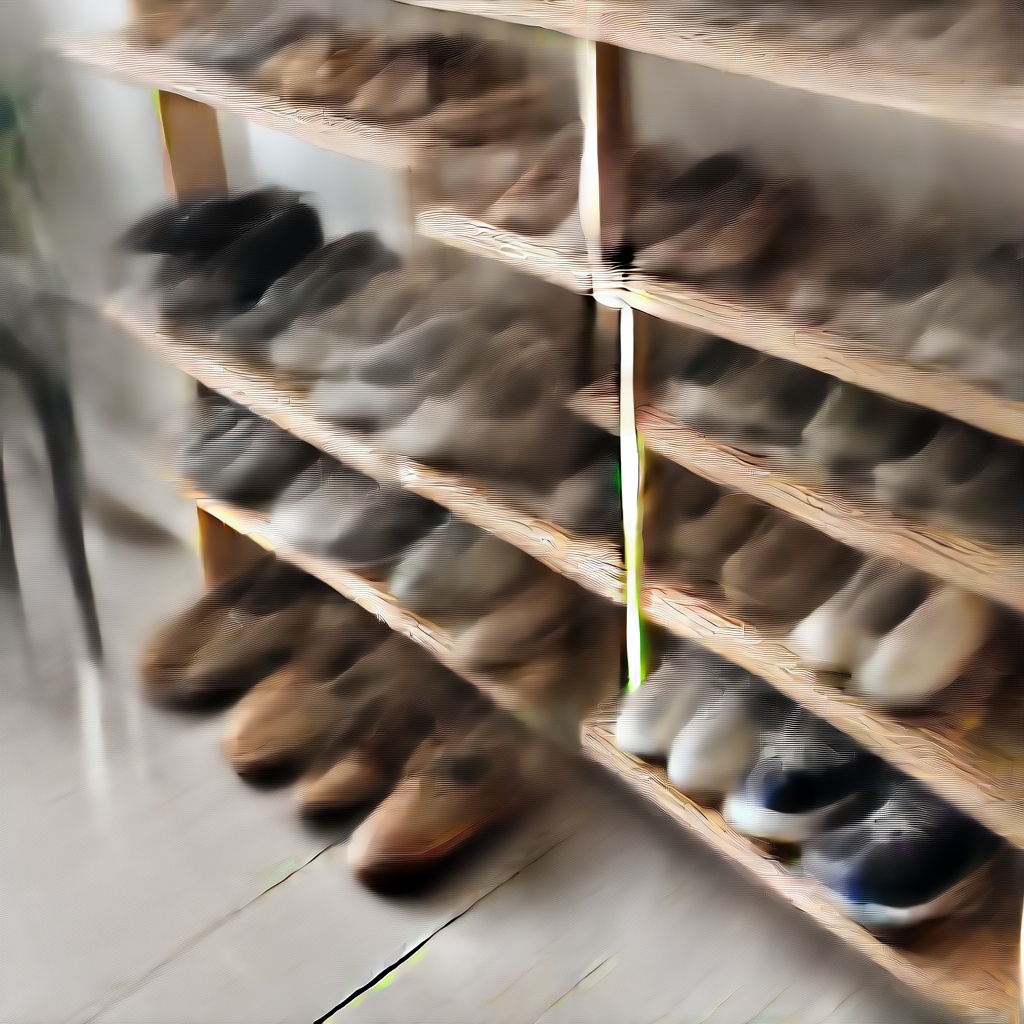} & 
        \includegraphics[width=0.13\linewidth]{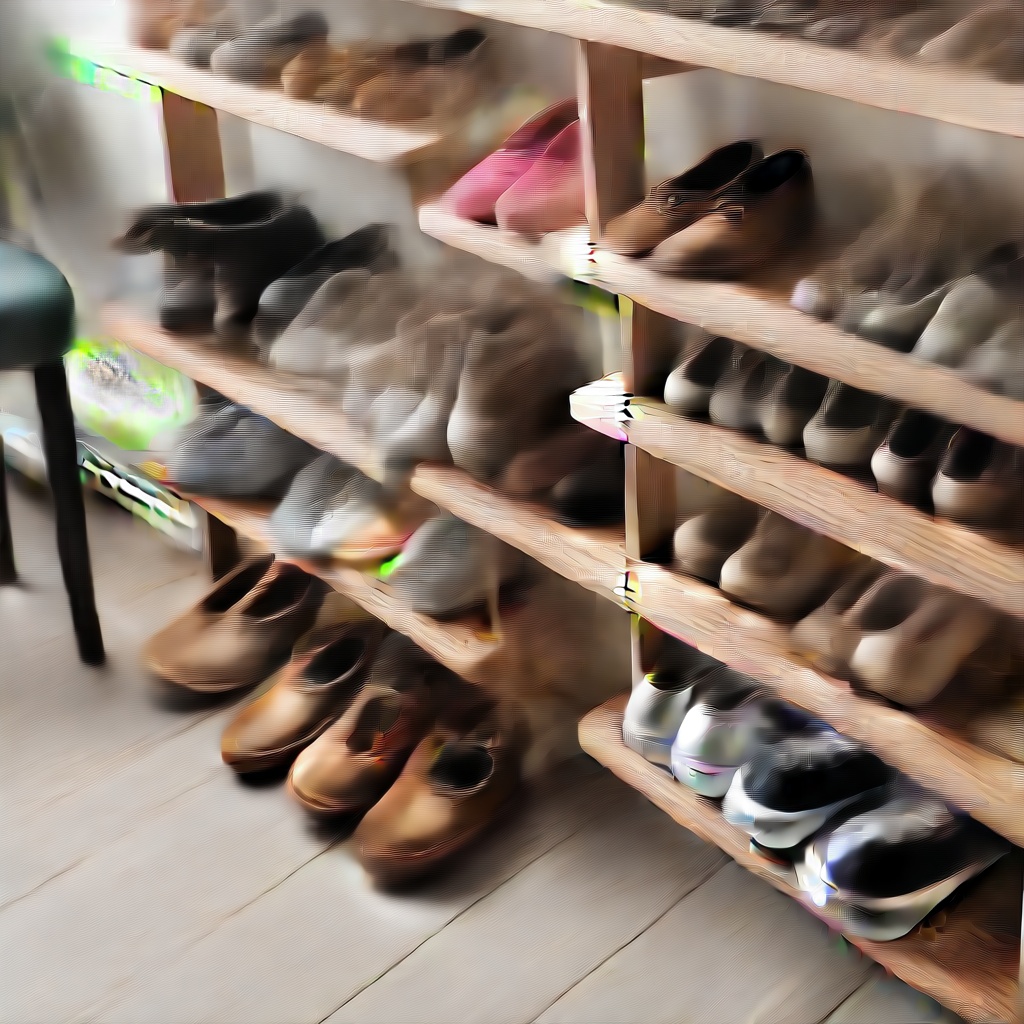} & 
        \includegraphics[width=0.13\linewidth]{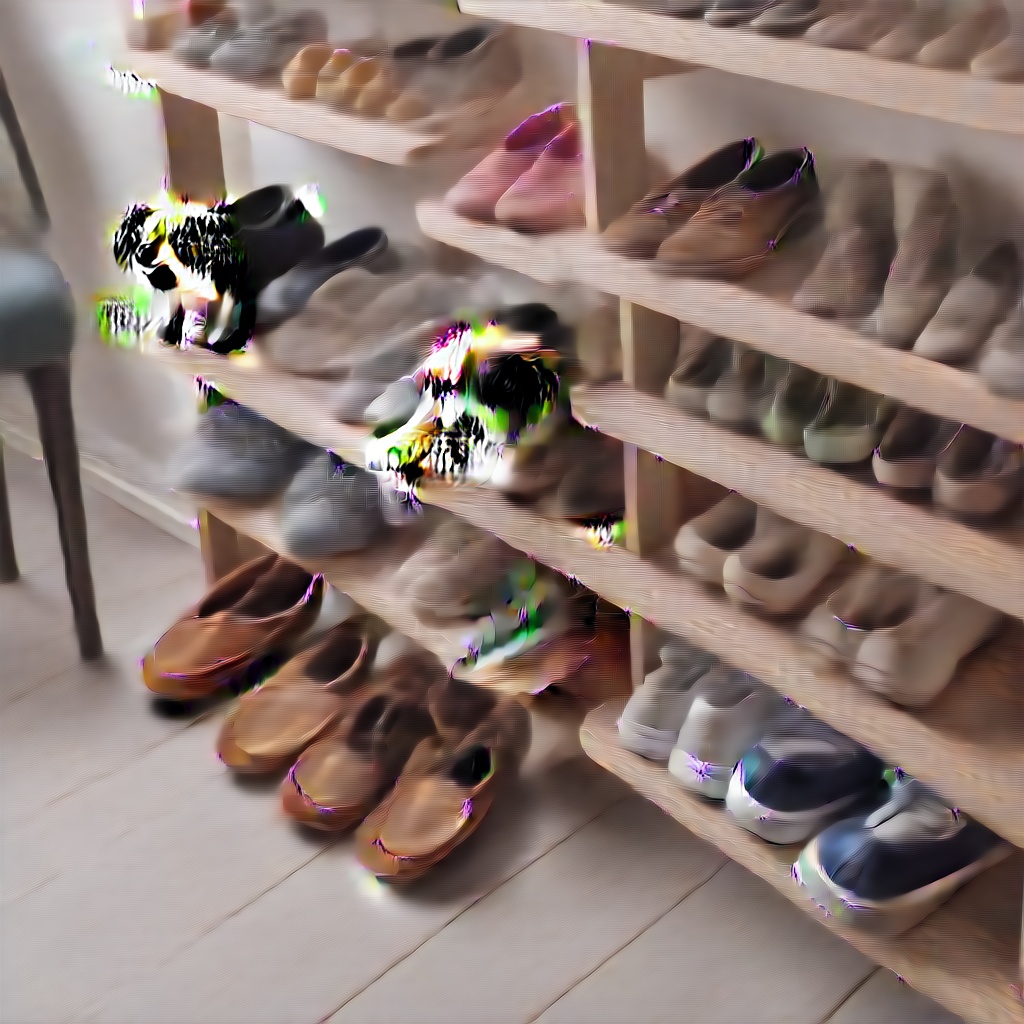} & 
        \includegraphics[width=0.13\linewidth]{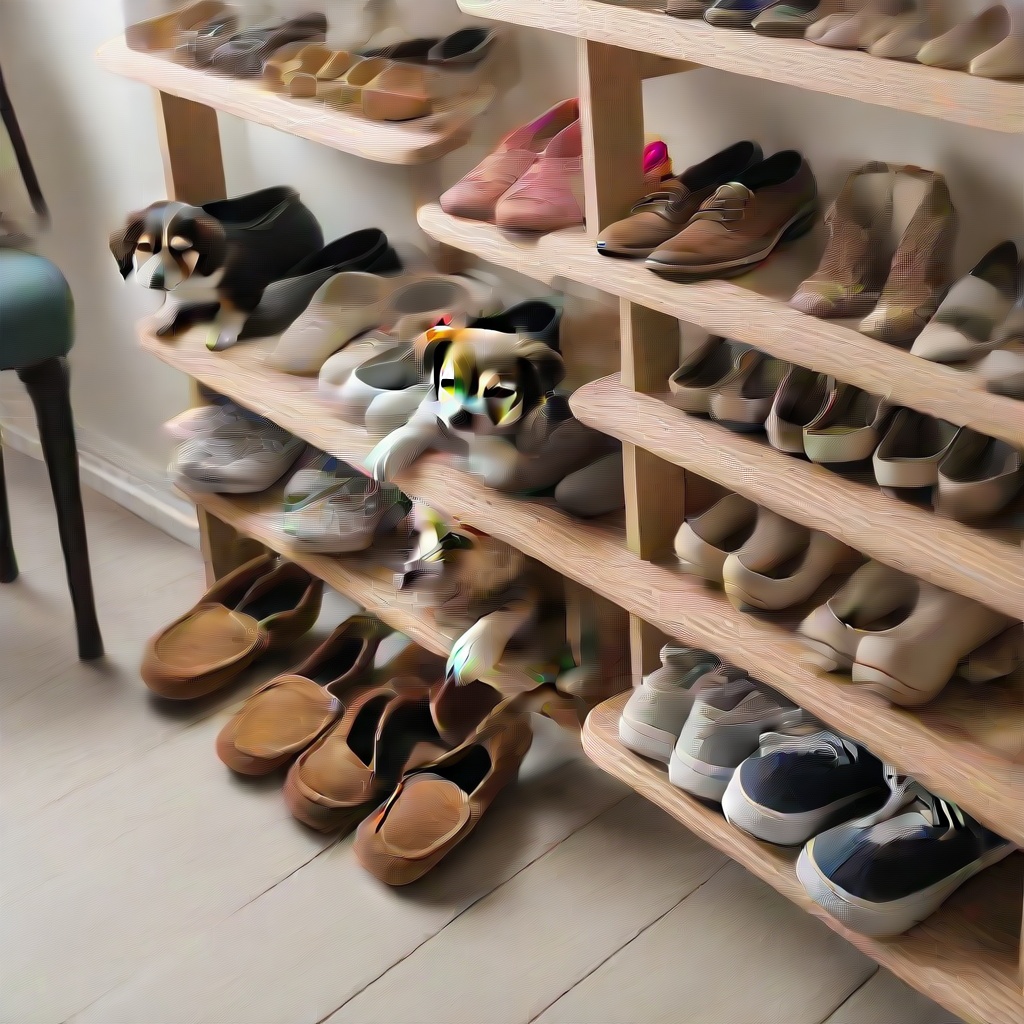} &
        \includegraphics[width=0.13\linewidth]{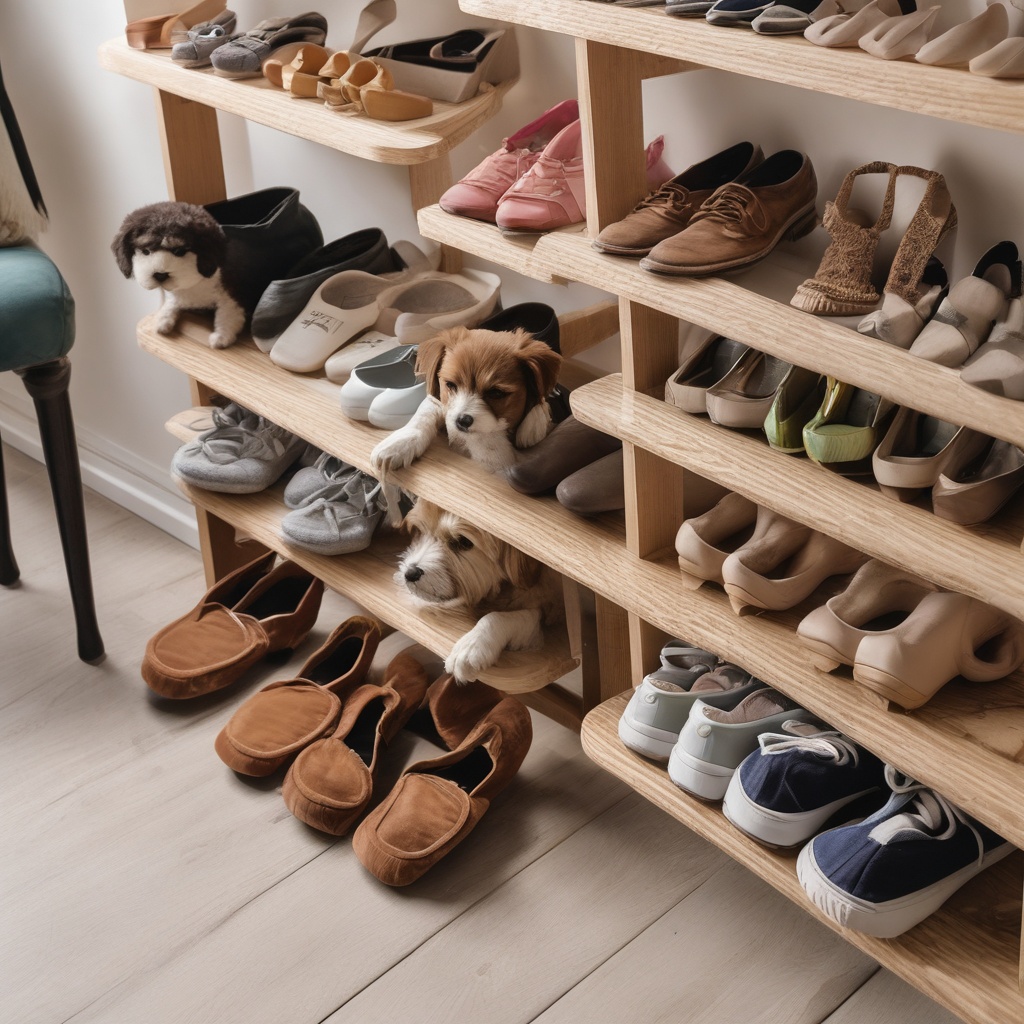}  
   
    \end{tabular}
    \caption{The the evolution of denoised estimates differs between CFG and \our{}. Both methods behave in a similar way at the beginning of the trajectory. However, \our{} converges faster to the data manifold to produce an image that is more consistent with the prompt: "a shoe rack with some shoes and a dog sleeping on them".}
    \label{fig:sampling}
\end{figure*}
\paragraph{CFG++} represents the simple extension of CFG, that utilizes a small guidance scale, typically $0 \leq \lambda \leq 1$, that enables smooth interpolation between unconditional and conditional sampling. The reverse diffusion process that utilizes CFG++ is provided by Algorithm \ref{alg:cfgplus}.

\section{\our{}}

In this section, we introduce \our{}, the novel approach for stabilizing the guidance process with the normalized, dynamic control of the impact of CFG in the denoising process. The section is organized as follows. First, we motivate dynamic scaling by analyzing the impact of CFG on various stages of diffusion sampling. Second, we introduce the general procedure for stabilizing the CFG with the $\beta$ function scaling and gradient normalization.   



\begin{algorithm}[t] 
    \caption{General Reverse Diffusion with CFG}
    \label{al:1}
\includegraphics[width=1.0\linewidth]{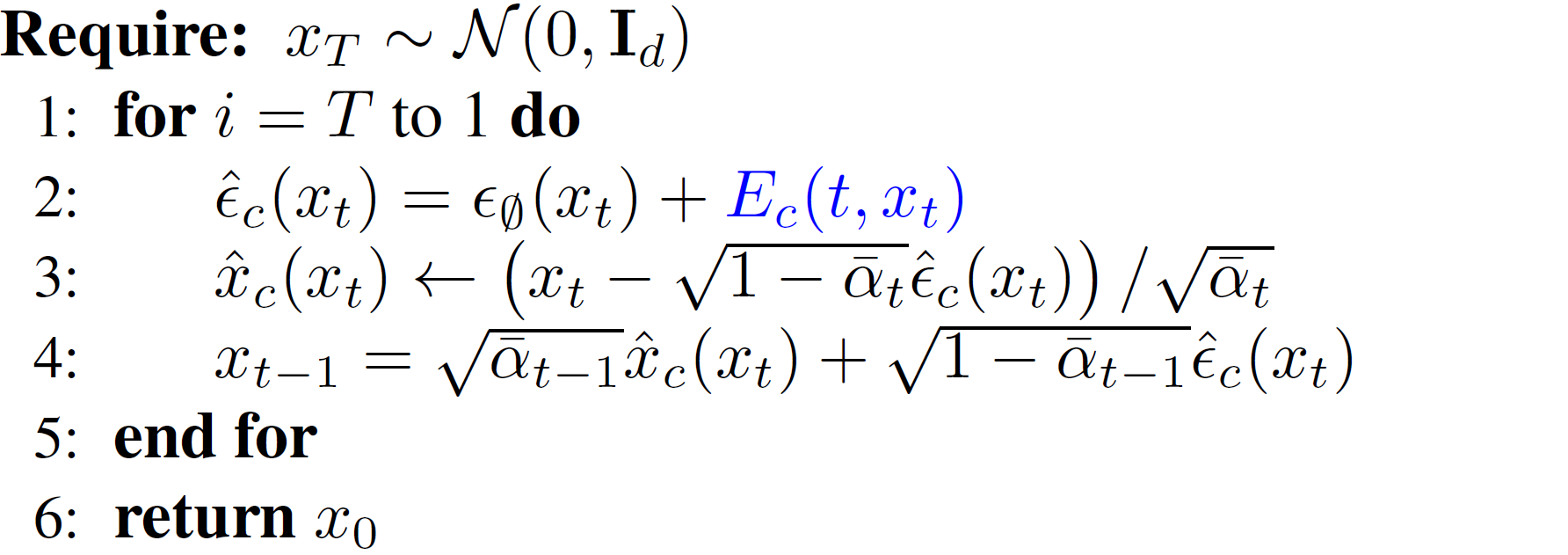}
\end{algorithm}

\subsection{Motivation}

The guided sampling for diffusion models can be generally written as provided in Algorithm \ref{al:1},  where $E_c(t,x_t)$ is the general correction (drift) term, which aims to guide the trajectory towards the region satisfying the desired properties. In the special case where $E_c(t,x_t)=0$, the model serves as the standard DDIM model without any guidance. When the diffusion model is adequately trained, it transforms the example from the data manifold $M$ at $t=0$ to the Gaussian distribution at $t=T$. Empirically, the sample from a Gaussian distribution is located in a small neighborhood of the sphere $S=\{x: \|x\|=\sqrt{d}\}$, where $d$ is the dimensionality of the data. This implies that for values of $t$ close to zero, the trajectories are near the data manifold $M$, whereas for values of $t$ close to $T$, the trajectories are near the sphere $S$.

To investigate the role of the adjustment term (which can be interpreted as a drift) $E_c(t,x_t)$, let us assume that $E_c(t,x_t)=0$ and denote $P(t,x_t) = x_{t-1}$ as the DDIM dynamic process in the following way:
\begin{equation} 
\begin{split}
P(t,x_t)&=\tfrac{\sqrt{{\bar\alpha}_{t-1}}}{\sqrt{\bar{\alpha}_t}}\left( x_t - \sqrt{1 - \bar{\alpha}_{t}}\hat{\epsilon}_{\emptyset}(x_t)\right)  \\
&+\sqrt{1 - {\bar\alpha}_{t-1}}\hat{\epsilon}_{\emptyset}(x_t).
\end{split}
\end{equation}

As a consequence, adding guidance component $E_c(t,\tilde{x}_t)$ modifies the model's dynamics in the following way: 

\begin{equation}
\tilde{x}_{t-1}=P(t,\tilde{x}_t)+e(t) \cdot E_c(t,\tilde{x}_t),
\end{equation}
where $e(t)$ is defined as:
\begin{equation}
e(t)=\sqrt{1-\bar \alpha_{t-1}}-\sqrt{\frac{1}{\alpha_t}-\bar \alpha_{t-1}}.    
\end{equation}

\begin{table*}[h!]
    \centering
    \begin{tabular}{ccccccccccc}
        \hline
        \multirow{2}{*}{Method} & \multicolumn{2}{c}{$\omega = 2.0, \lambda = 0.2$} & \multicolumn{2}{c}{$\omega = 5.0, \lambda = 0.4$} & \multicolumn{2}{c}{$\omega = 7.5, \lambda = 0.6$} & \multicolumn{2}{c}{$\omega = 9.0, \lambda = 0.8$} & \multicolumn{2}{c}{$\omega = 12.5, \lambda = 1.0$} \\
        & FID $\downarrow$ & CLIP $\uparrow$ & FID $\downarrow$ & CLIP $\uparrow$ & FID $\downarrow$ & CLIP $\uparrow$ & FID $\downarrow$ & CLIP $\uparrow$ & FID $\downarrow$ & CLIP $\uparrow$ \\
        \hline
        CFG & 13.94 & 0.306 & 16.16 & \bf 0.318 & 18.98 & \bf 0.319 & 20.16 & \bf 0.320 & 22.32 & \bf 0.320 \\
        CFG++ & \bf 13.36 & \bf 0.311 & 16.02 & \bf 0.318 & 18.57 & \bf 0.319 & 20.48 & \bf 0.320 & 21.97 & \bf 0.320 \\
        \our{} & 15.02 & 0.306
        & \bf 15.76 & 0.317
        & \bf 17.99 & \bf 0.319
        & \bf 18.94 & 0.319
        & \bf 20.97 & \bf 0.320
        \\
        \hline

    \end{tabular}
    \caption{Quantitative evaluation (FID, CLIP-similarity) of 50NFE DDIM T2I with SD v1.5 on COCO 10k.
    }
    \label{tab:all}
\end{table*}

Since $\alpha_t=1-\beta_t$, and $\beta_t \neq 0$, we conclude that the function $e(t)$ is generally nonzero.

The correct DDIM training ensures that manifolds $M$ and $S$ act as attractors of the model at times $t=0$ and $t=T$. That is, every trajectory starting on the manifold $S$ is drawn towards $M$ and arrives at $M$ at $t=0$. Moreover, adding a sufficiently small noise at some point $x_t$ (for some $t>0$) will be compensated for by the model, ensuring convergence to $M$. Conversely, when reversing time, a trajectory starting in a small neighborhood of $M$ at $t=0$ will arrive in a neighborhood of $S$ at $t=T$.

However, this property is no longer assured if we incorporate a correction term $e(t) \cdot  E_c(t,x_t)$. Specifically, the trajectory at $t=0$ will usually diverge from the desired data manifold $M$. To illustrate this, consider the final step at~$t=1$:
\begin{equation}
\begin{array}{c}
\tilde{x}_0=P(1,\tilde{x}_1)+e(1) \cdot E_c(1,\tilde{x}_1).
\end{array}
\end{equation}
Since the correct iterative procedure ensures $m=x_0=P(1,\tilde{x}_1) \in M$, adding the correction $e(1) \cdot E_c(1,\tilde{x}_1)$ results in:
\begin{equation}
\tilde{x}_0=m+e(1) \cdot E_c(1,\tilde{x}_1), \quad \text{for some } m \in M.
\end{equation}

This implies that, in general, $\tilde{x}_0$ will not lie in the data manifold $M$.
This could be a significant drawback since we strongly prefer the images generated by the diffusion model to remain in the data manifold. This property is preserved if the function $E_c(1,\tilde{x}_1)$ satisfies the following boundary conditions:
\begin{equation}
\lim_{t \to 0}E_c(t,x)=0 \text{ and } \lim_{t \to T}E_c(t,x)=0.
\label{eq:conditions}
\end{equation}

To enforce these conditions, we propose multiplying the correction term by a continuous function that vanishes at the time limits. In our paper, we implement this by applying the beta distribution:
\begin{equation} 
 \beta(t)=\frac{t^{a - 1} (1 - t)^{b - 1}}{B(a, b)}, 
 \label{eq:beta}
\end{equation}
where $B(a, b)$ is \emph{Beta} function, and $a$ and $b$ are the hyperparameters that control the curvature of the density function. The function is defined for $ t \in [0, 1]$, so the integer indexing should be rescaled to this interval. We propose this kind of function due to the flexibility of modeling and shifting function with one mode, assuming $a>1$ and $b>1$, which guarantee that $\beta(0)=\beta(1)=0$. 

Thus the general model for an arbitrary is given by
Algorithm \ref{al:1}, with function $E$ multiplied by 
$\beta_{a,b}(t/T)$. In the next subsection we present the algorithm devised for CFG.

\begin{algorithm}[t]
    \caption{Reverse Diffusion with \our{}$_{\gamma}$ }
        \label{alg_our}
\includegraphics[width=1.0\linewidth]{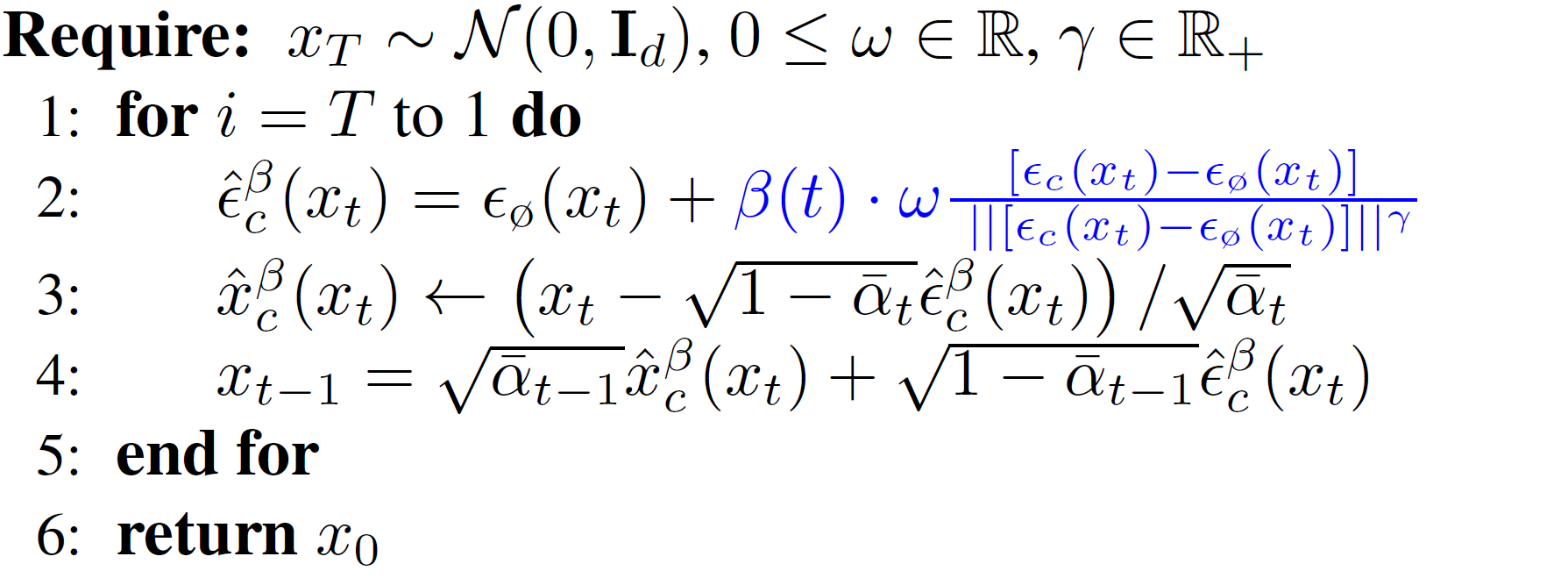}
        \label{alg:our}
\end{algorithm}




\begin{figure*}[t]
    \centering
    \renewcommand{\arraystretch}{0}
    \setlength{\tabcolsep}{0.4pt}
    \begin{tabular}{c@{}c@{}c@{}c@{}c@{}c@{}}
        
        \includegraphics[width=0.15\linewidth]{img/images/beta/beta_distribution_single_2.0_2.0.jpg} &
        \includegraphics[width=0.15\linewidth]{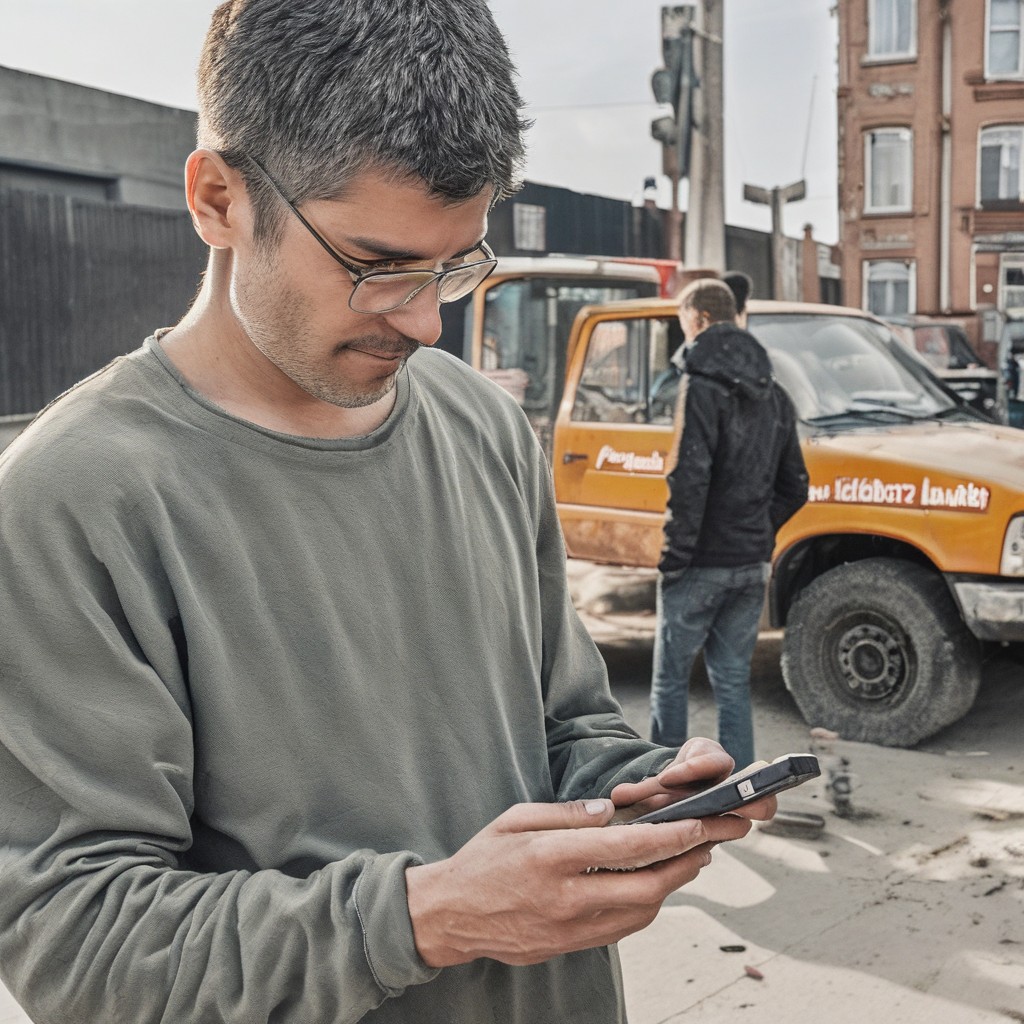} &
        \includegraphics[width=0.15\linewidth]{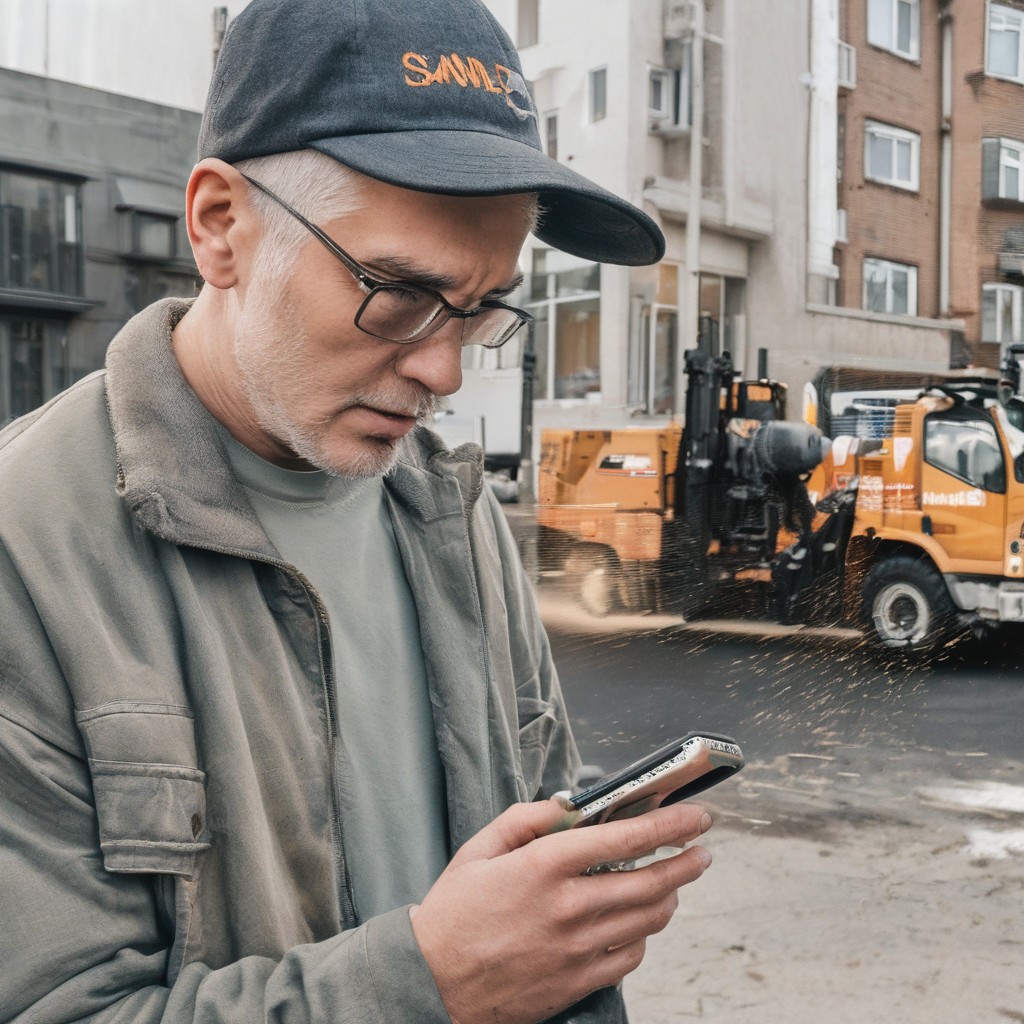} &
        \includegraphics[width=0.15\linewidth]{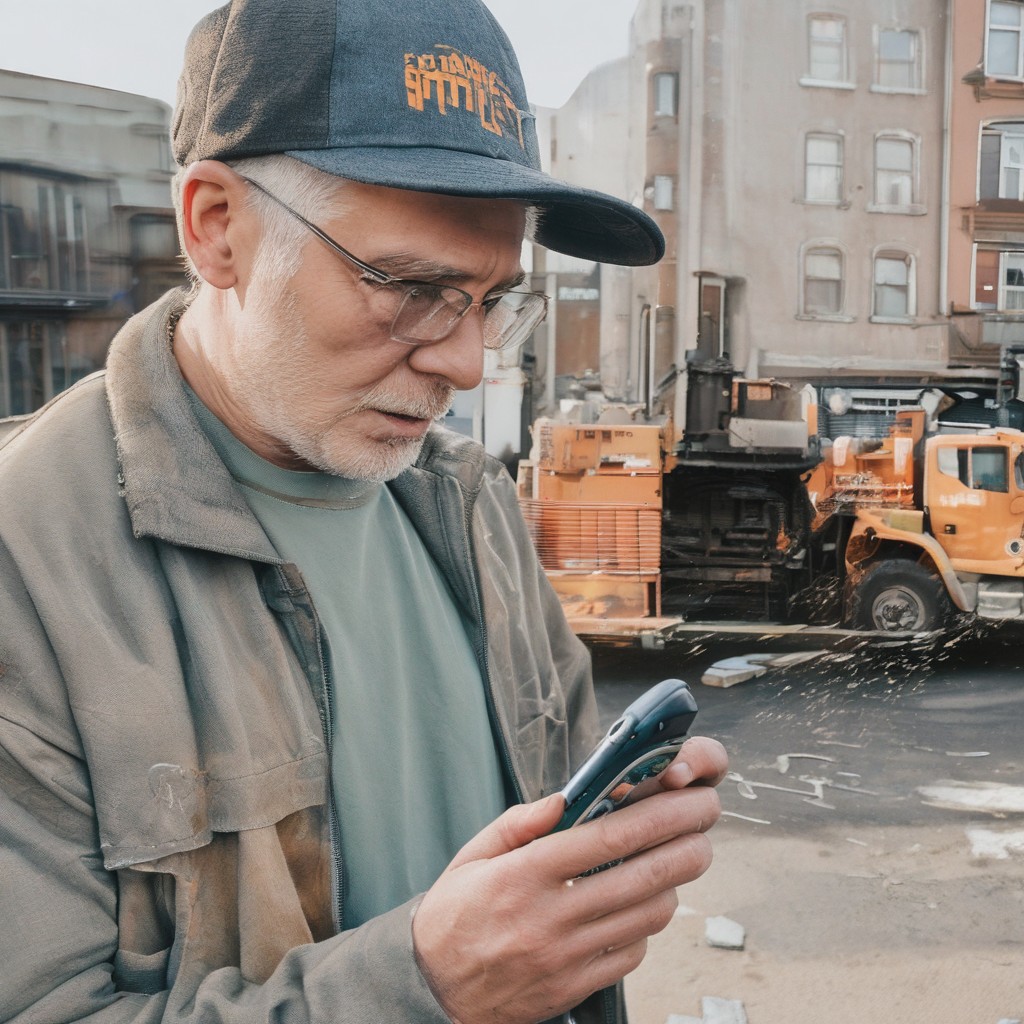} &
        \includegraphics[width=0.15\linewidth]{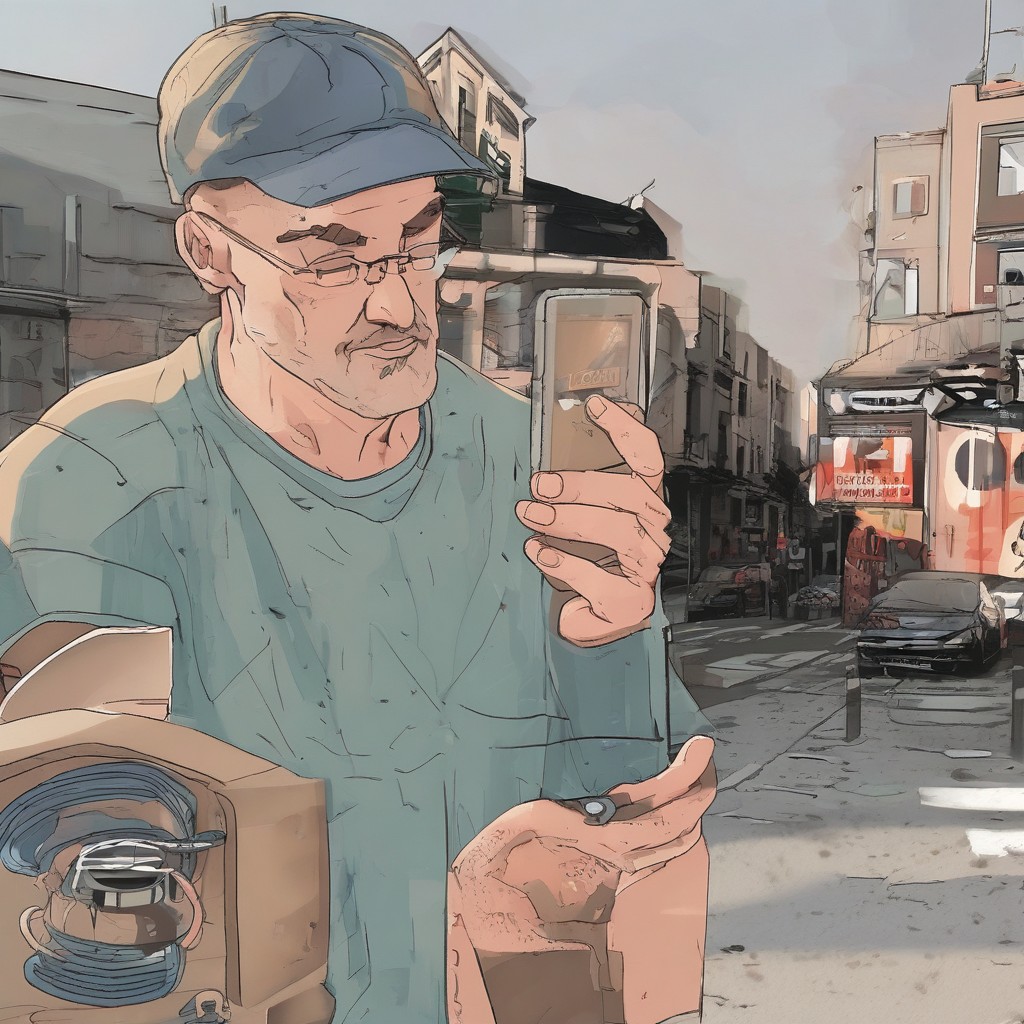} &
        \includegraphics[width=0.15\linewidth]{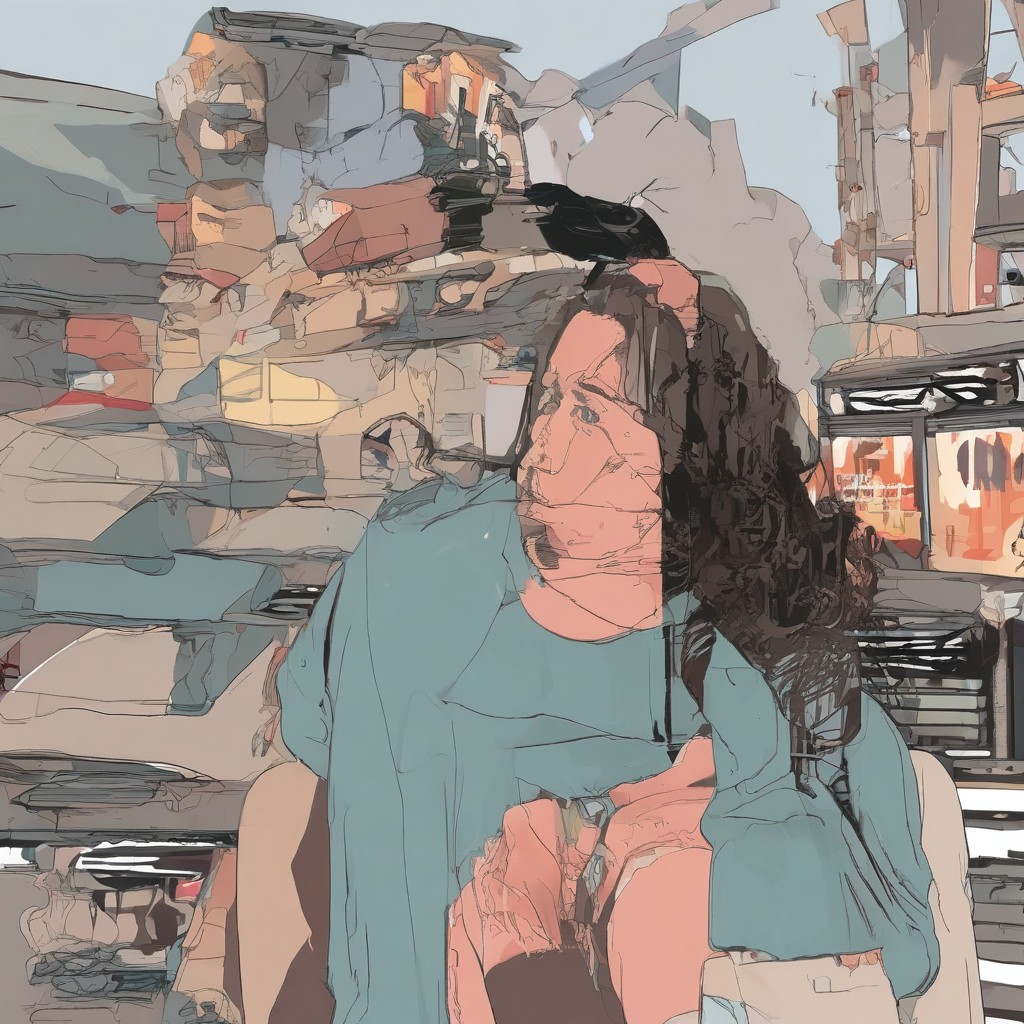} \\[0.2cm]
        \our{} & $\gamma = 0.0 $ & $\gamma = 0.25 $ & $\gamma = 0.5 $  & $\gamma = 1.0 $ & $\gamma = 2.0 $ \\
    \end{tabular}
    \caption{Example of sampled element according to $\gamma$ parameters. Prompt: "A man holding a phone while sanding next to a street.".}
    \label{fig:sampling_example2}
\end{figure*}

\subsection{Sampling with \our{}} 

As shown in the previous subsection, the CFG may track the generated sample in the regions outside the data manifold. As a consequence, the impact of CFG should be different for some particular stages of sample generation. The initial sampling stage should focus on general templates of images, so the impact of the conditional model should be minor. During the intermediate stage, the model should follow the path determined by the conditioning factor $c$, increasing the importance of the CFG component. During the final stage of generating, the impact of CFG should be minor in order to locate the generated sample in the data manifold. To incorporate this, we propose to modify the CFG by simply scaling this term with the dynamic function:
\begin{equation} 
\hat{\epsilon}^{\beta}_{c}(x_t) = \epsilon_{\o}(x_t) 
+ \beta(t) \cdot \omega  
\frac{[\epsilon_c(x_t) - \epsilon_{\o}(x_t)]}{\ ||[\epsilon_c(x_t) - \epsilon_{\o}(x_t)] ||^{\gamma} },
\end{equation}
where $\beta(t)$ is the function that controls the impact of normalized CFG during the various training stages, and $\omega$ hyperparameter controls the magnitude of the scaling function. Moreover, drawing inspiration from the GeGuide approach, we propose normalizing the guidance term using the norm to the power of $\gamma \in \mathbb{R}_{+}$. Consequently, we ensure that the scaled updates remain independent of the dimensionality of the data.

In general, any $\beta(t) \geq 0$ that enforces conditions given by \eqref{eq:conditions} can be used to model the dynamics for scaling CFG. In this work, we postulate to utilize the density function for $\beta$ distribution given by the equation \eqref{eq:beta} that has desired properties and preserves the volume. The modified procedure of sampling with our approach is provided by Algorithm \ref{alg:our}.

\our{} can be also easily adopted to the CFG++ process, where the guidance step 2 from Algorithm \ref{alg:cfgplus} is replaced by the following update:
\begin{equation} 
\hat{\epsilon}^{\beta++}_{c}(x_t) = \epsilon_{\o}(x_t) 
+ \beta(t) \cdot \lambda  
\frac{[\epsilon_c(x_t) - \epsilon_{\o}(x_t)]}{\ ||[\epsilon_c(x_t) - \epsilon_{\o}(x_t)] ||^{\gamma} }.
\end{equation}

\section{Experiments}

This section presents a series of experiments designed to evaluate the performance of our method in comparison to reference CFG and CFG++. We start with a simple 2D example to visually demonstrate our model's behavior. Then, we conduct both quantitative and qualitative comparisons against for both SD v1.5 and SDXL models, using 50 NFE DDIM sampling. Finally, we report the results of our ablation studies.

\begin{figure}
    \centering
        \includegraphics[width=\linewidth]{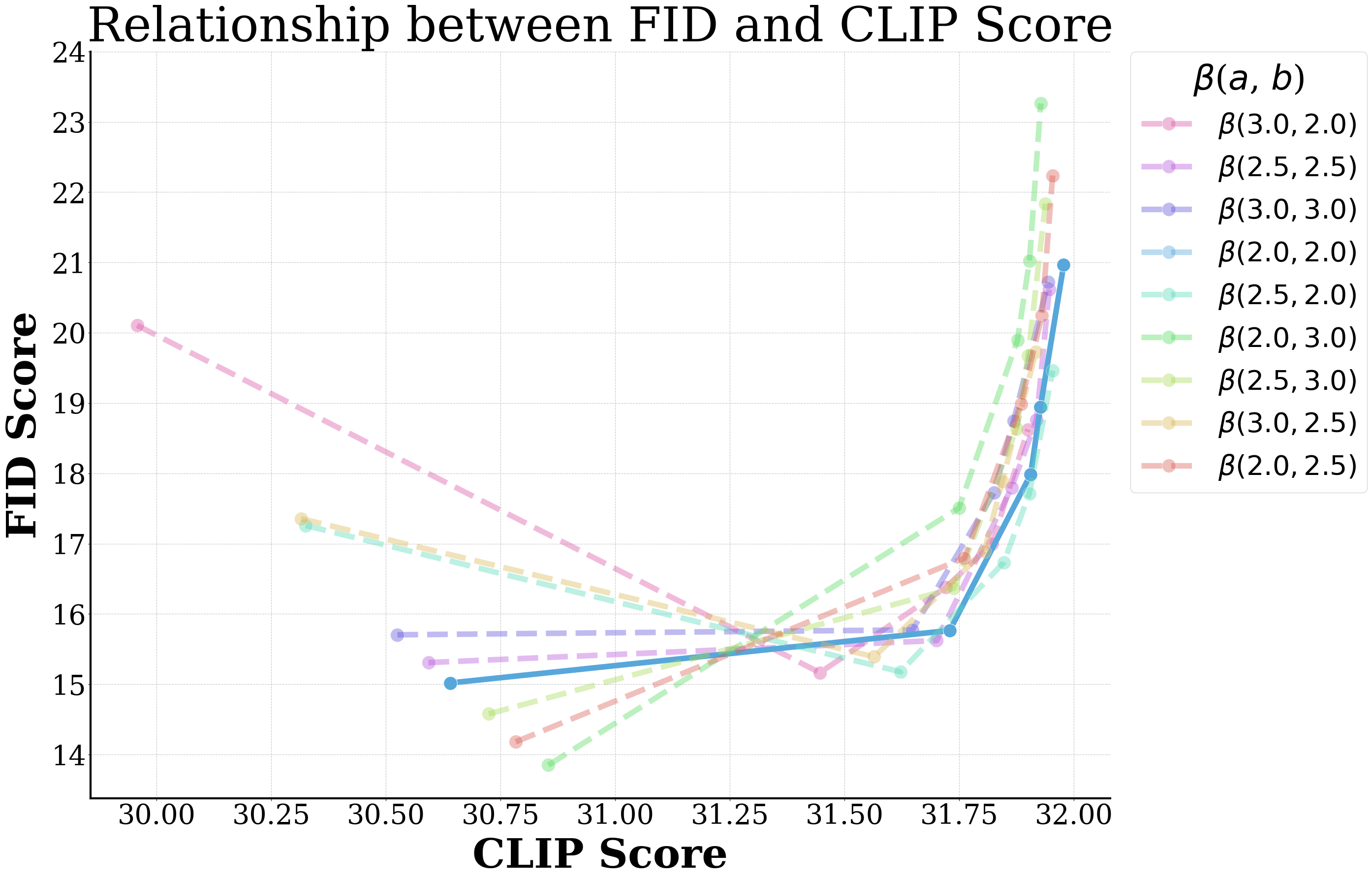}
        \vspace{-0.3cm}
    \caption{Ablation study of \our{} models according to $\beta$-distribution parameters. We present the FID and CLIP score relation when the $\omega$ parameter is changed.  }
    \label{fig:ablation_cfg_beta}
    \vspace{-0.3cm}
\end{figure}

\paragraph{Toy example 2D}
To illustrate why unguided diffusion models often produce poor images and how CFG mitigates this, as discussed in \cite{karras2024guiding}, the authors present a 2D toy example. A simple denoiser is trained for conditional diffusion on a synthetic dataset (Fig.~\ref{fig:toy-example}), designed with low local dimensionality and anisotropic structure. As noise decreases, local details emerge, mimicking real-world image manifolds \cite{brown2022verifying}.

In contrast to direct sampling from the original distribution (as depicted in Fig.~\ref{fig:toy-example} (a)), the unguided diffusion approach illustrated in Fig. 1b yields a significant quantity of highly improbable samples that lie beyond the main part of the distribution. In the context of generating images, these would equate to distorted or inadequate images Fig.~\ref{fig:toy-example} (a) and (b) display the learned score field and implied density in our illustrative example for two models with different capacities at a mid-level of noise. The classical CFG model encapsulates the data more closely, whereas the weaker model without guidance exhibits a more dispersed density.
\our{} model fits the target distribution more precisely than CDF. Additionally, it generates fewer outlier elements, as depicted in Fig.~\ref{fig:toy-example} (c).

\paragraph{Text to image generation}
In this experiment, we evaluate the quality of generated images (FID score) and match the prompt (CLIP score).
Utilizing specific scales for $\omega$ and $\lambda$, we directly compare the T2I task performance of SD v1.5. Tab.~\ref{tab:all} provides quantitative metrics based on 10,000 images created with COCO captions \cite{lin2014microsoft}. In practical application, \our{} achieves an improved FID score, as shown in Tab.~\ref{tab:all}, with a similar CLIP score.  Fig.~\ref{fig:sampling}) present samples generated from SDXL model.

\paragraph{Ablation study}
In \our{}, two significant parameters are employed. The first is the $\beta$-distribution utilized in the experiment. Fig.~\ref{fig:sampling} show relation between FID and CLIP score. The model with $\beta(2,2)$ parameters achieves the highest score. 
In Fig. \ref{fig:sampling_example2}, we illustrate the impact of the $\beta$ parameters on the sampling process. When $\gamma$ equals 1, the trajectory aligns precisely with the $\beta$-distribution. When $\gamma$ lies between 0 and 1, it modifies the intermediate phase of the diffusion process. For $\gamma$ values exceeding 1, it becomes evident that the guidance is overly strong.


\section{Conclusions}

In this paper, we explored the impact of classifier-free guidance (CFG) on text-driven diffusion models, highlighting its trade-off between image quality and prompt adherence. We analyzed how CFG behaves across different noise levels, influencing the sampling process at various stages. To address the inherent limitations of CFG, we introduced \our{} ($\beta$-adaptive scaling in Classifier-Free Guidance), which dynamically adjusts guidance strength throughout the generation process. By employing time-dependent $\beta$-distribution scaling, \our{} effectively balances prompt alignment and image fidelity. Experimental results demonstrated that our approach achieves improved FID scores while maintaining text-to-image CLIP similarity comparable to standard CFG. 
\paragraph{Limitations} The primary drawback is the introduction of three extra meta-parameters. For future work, we intend to develop a mechanism for automatic parameter matching.

\section*{Impact Statement}

This paper presents work whose goal is to advance the field of Machine Learning. There are many potential societal consequences of our work, none which we feel must be specifically highlighted here.

\bibliographystyle{icml2025}

\newpage
\appendix
\onecolumn
\section{Ablation study.}

\begin{figure*}[!h]
    \centering
    \renewcommand{\arraystretch}{0}
    \setlength{\tabcolsep}{0.4pt}
    \begin{tabular}{c@{}c@{}c@{}c@{}c@{}c@{}}
        
        \includegraphics[width=0.15\linewidth]{img/images/beta/beta_distribution_single_2.0_2.0.jpg} &
        \includegraphics[width=0.15\linewidth]{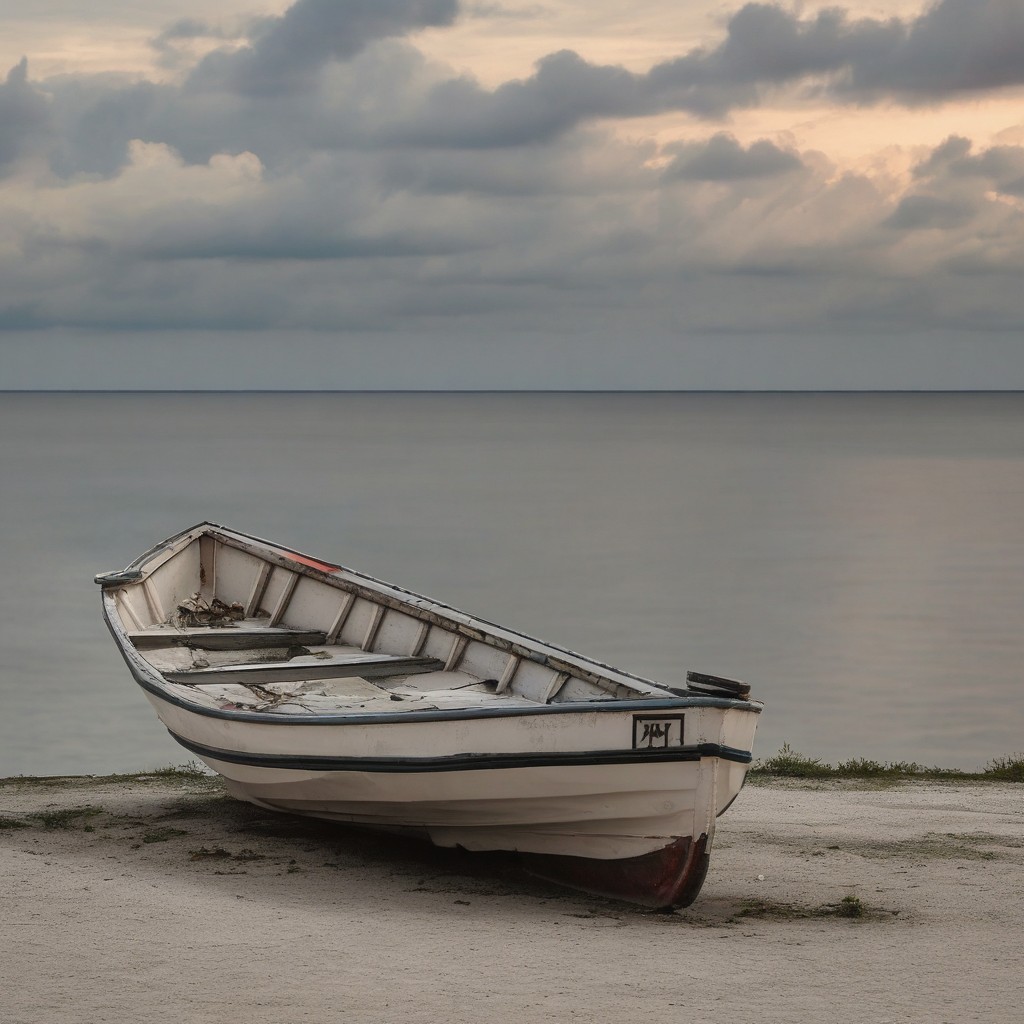} &
        \includegraphics[width=0.15\linewidth]{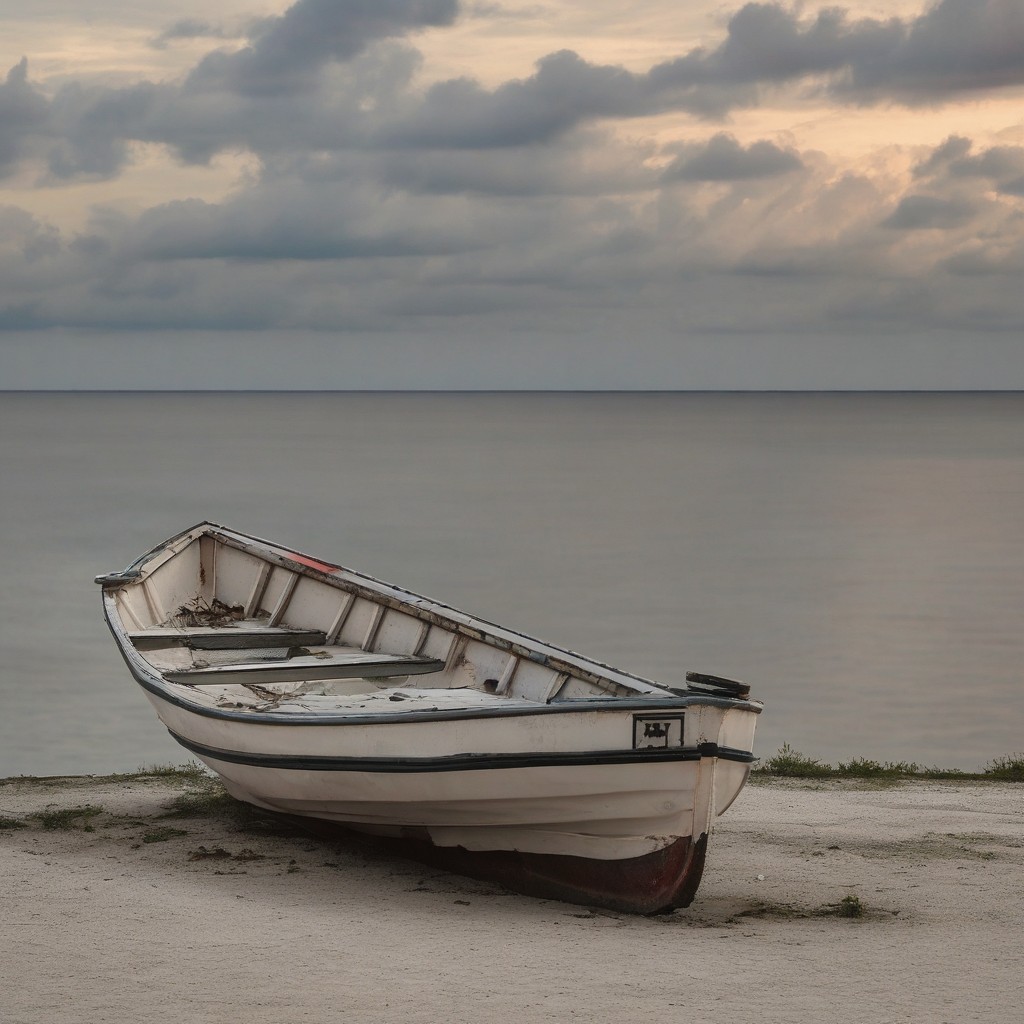} &
        \includegraphics[width=0.15\linewidth]{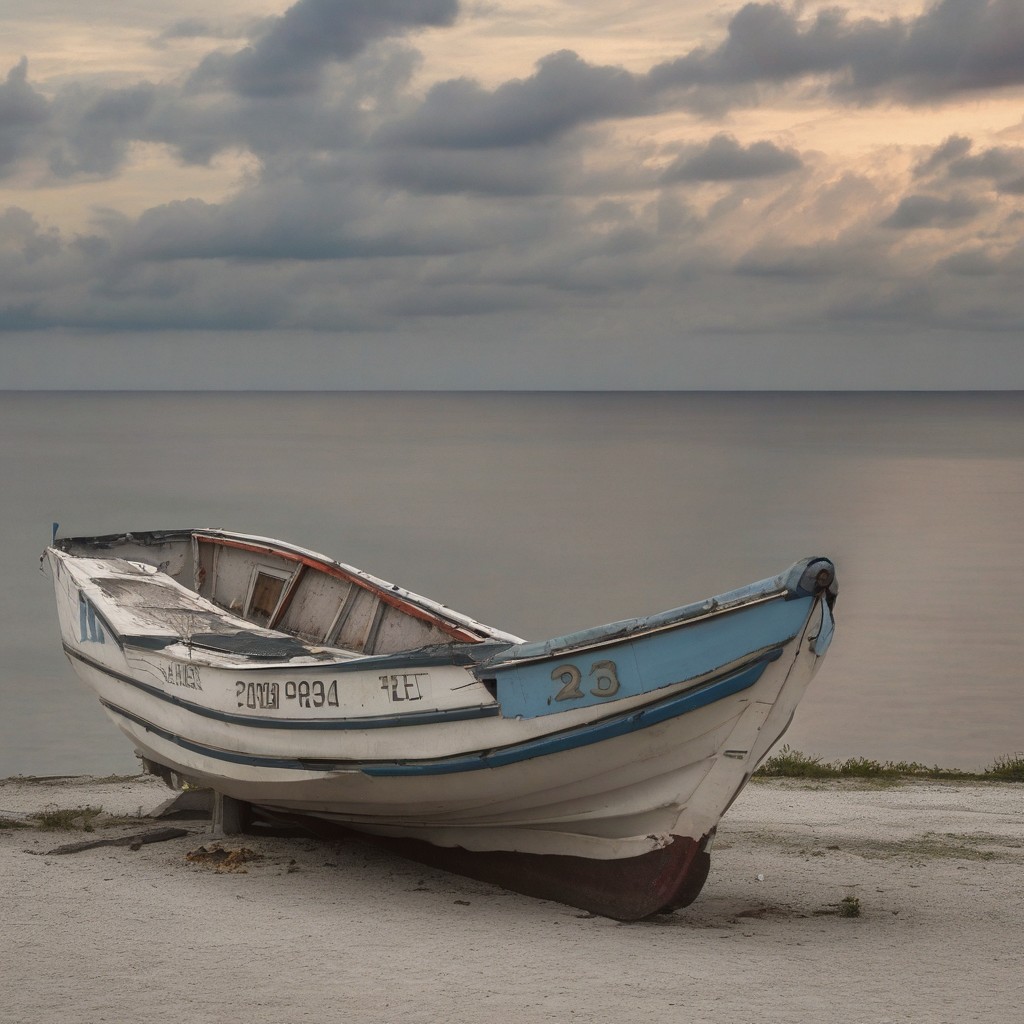} &
        \includegraphics[width=0.15\linewidth]{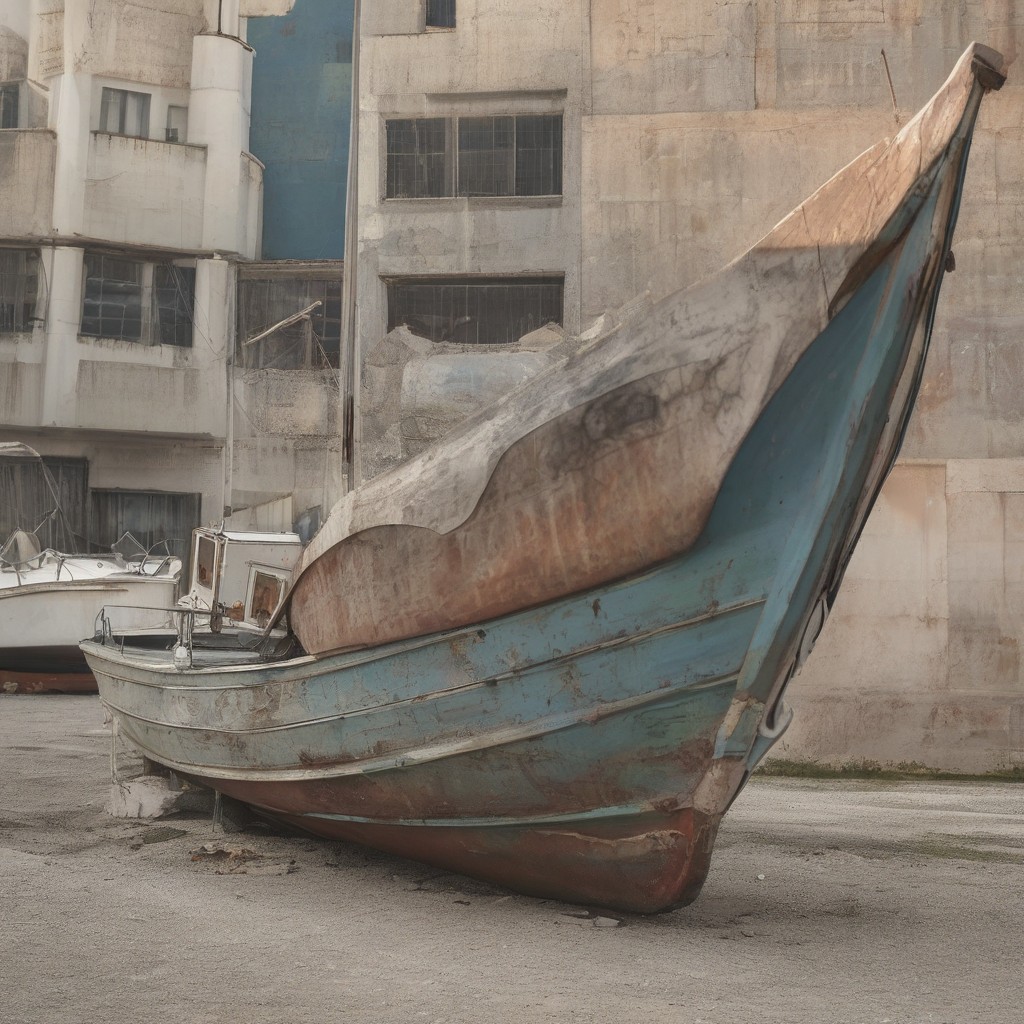} &
        \includegraphics[width=0.15\linewidth]{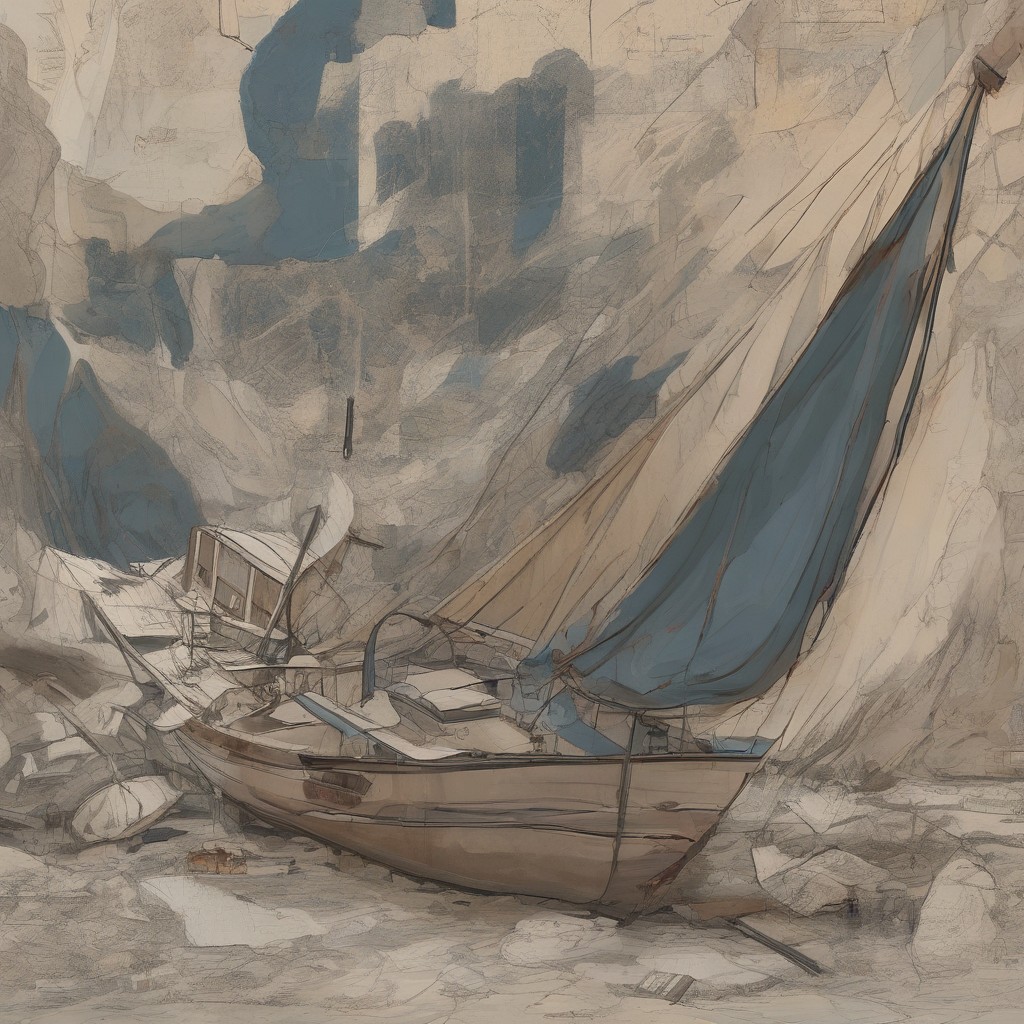} \\
        \includegraphics[width=0.15\linewidth]{img/images/beta/beta_distribution_single_2.0_2.0.jpg} &
        \includegraphics[width=0.15\linewidth]{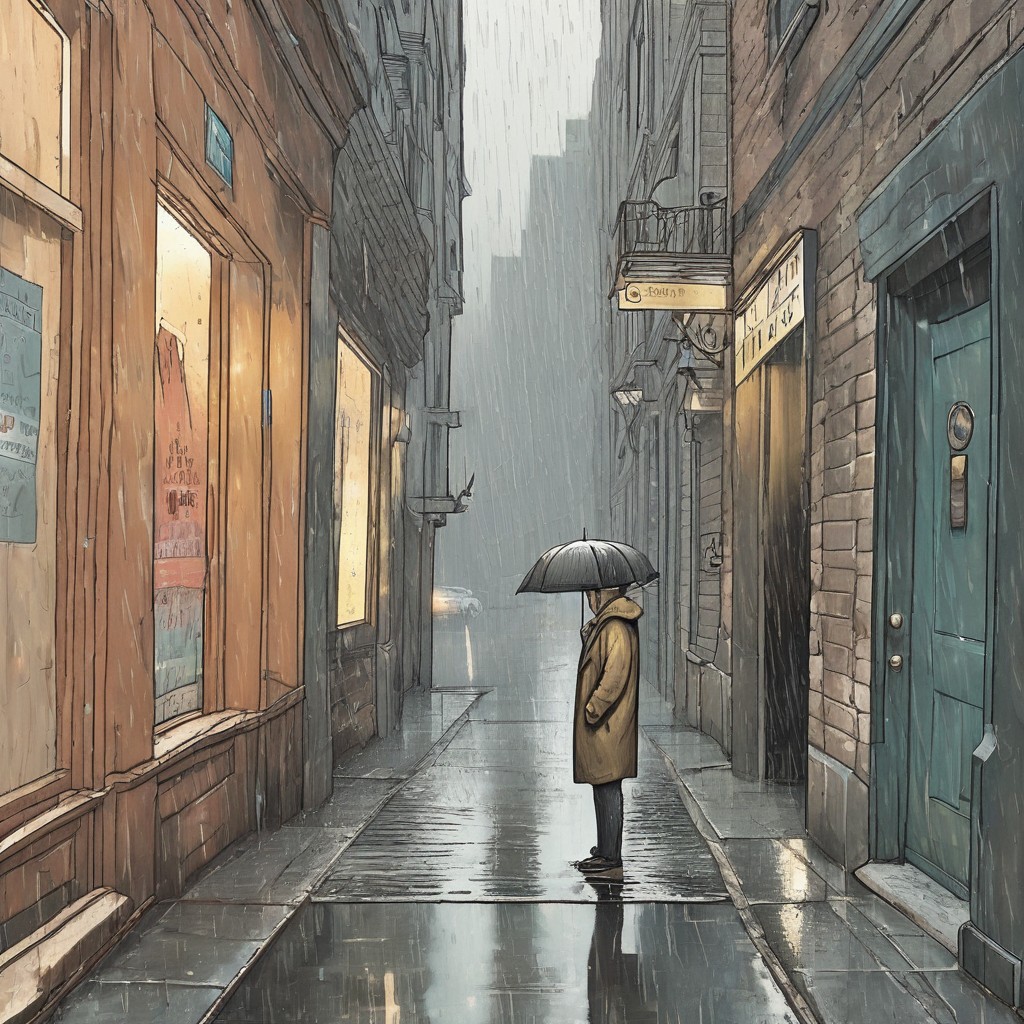} &
        \includegraphics[width=0.15\linewidth]{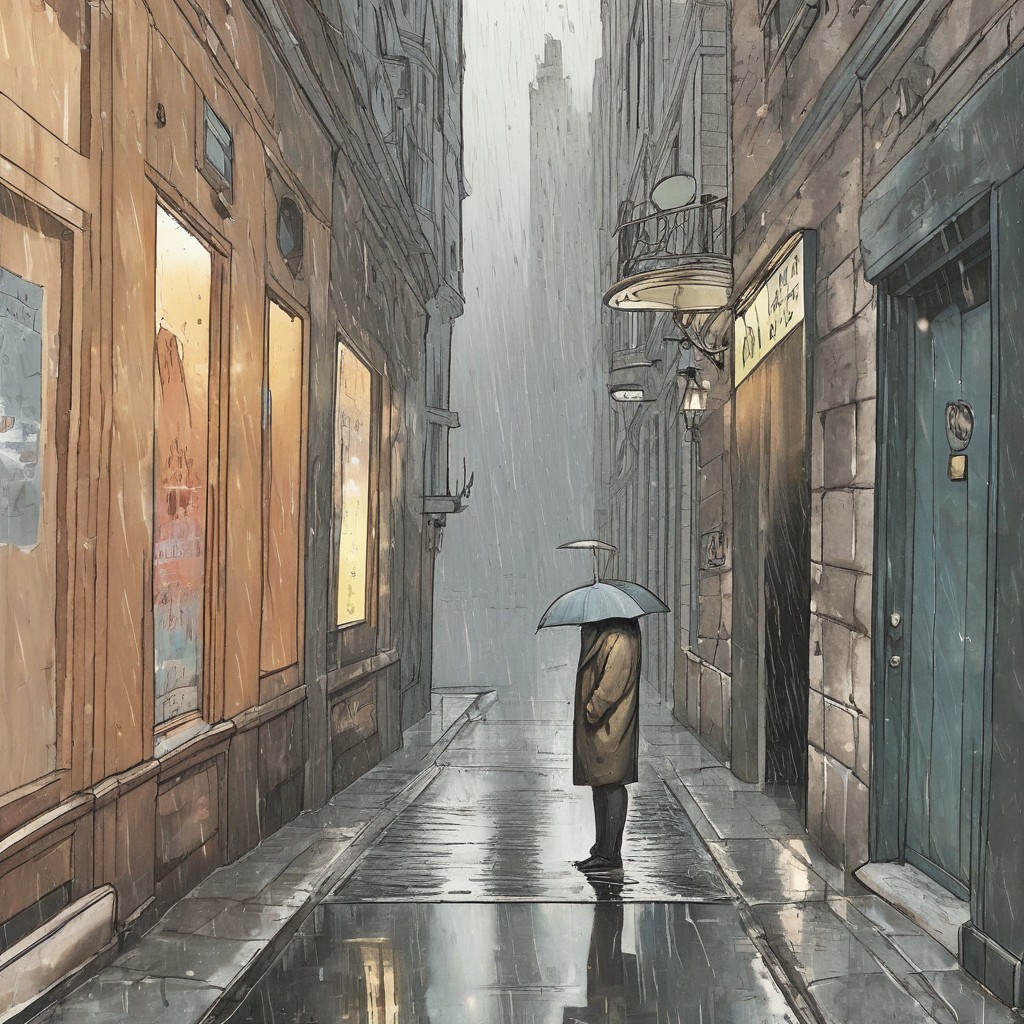} &
        \includegraphics[width=0.15\linewidth]{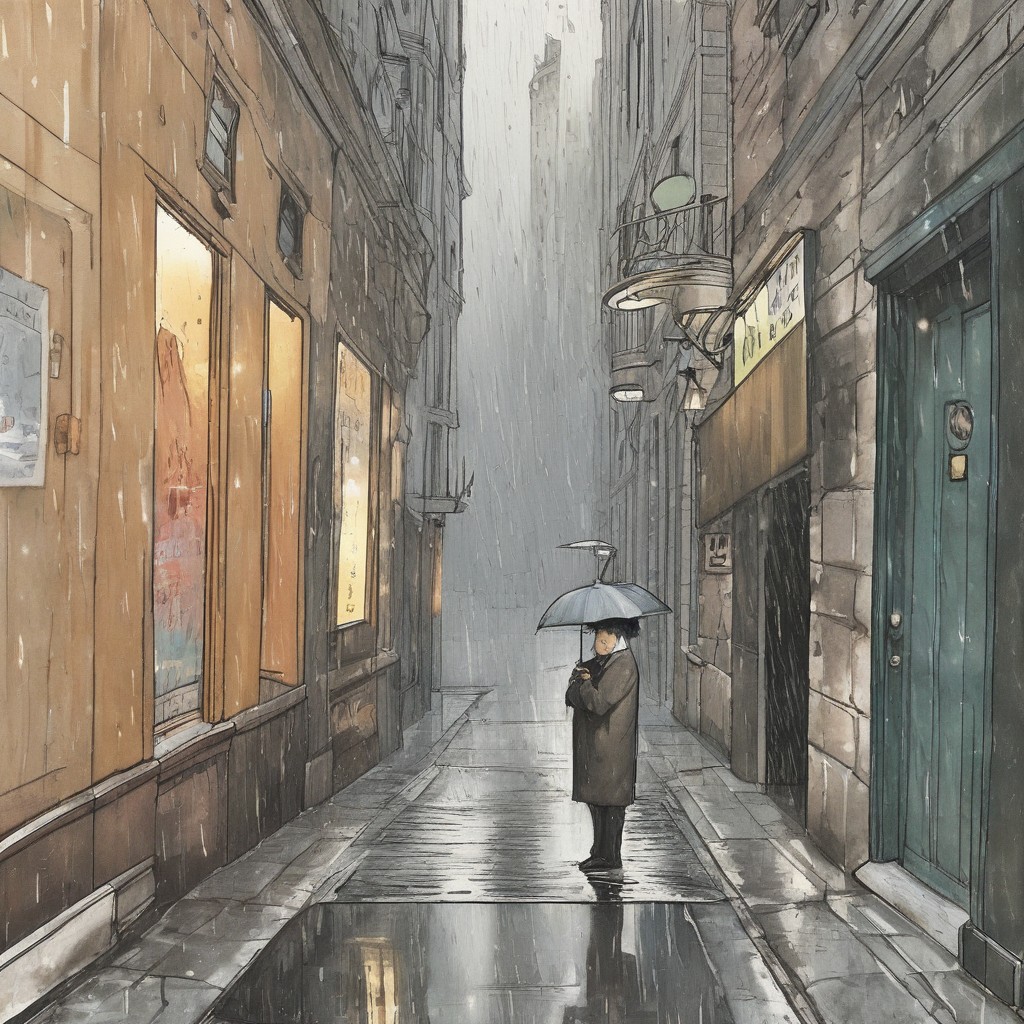} &
        \includegraphics[width=0.15\linewidth]{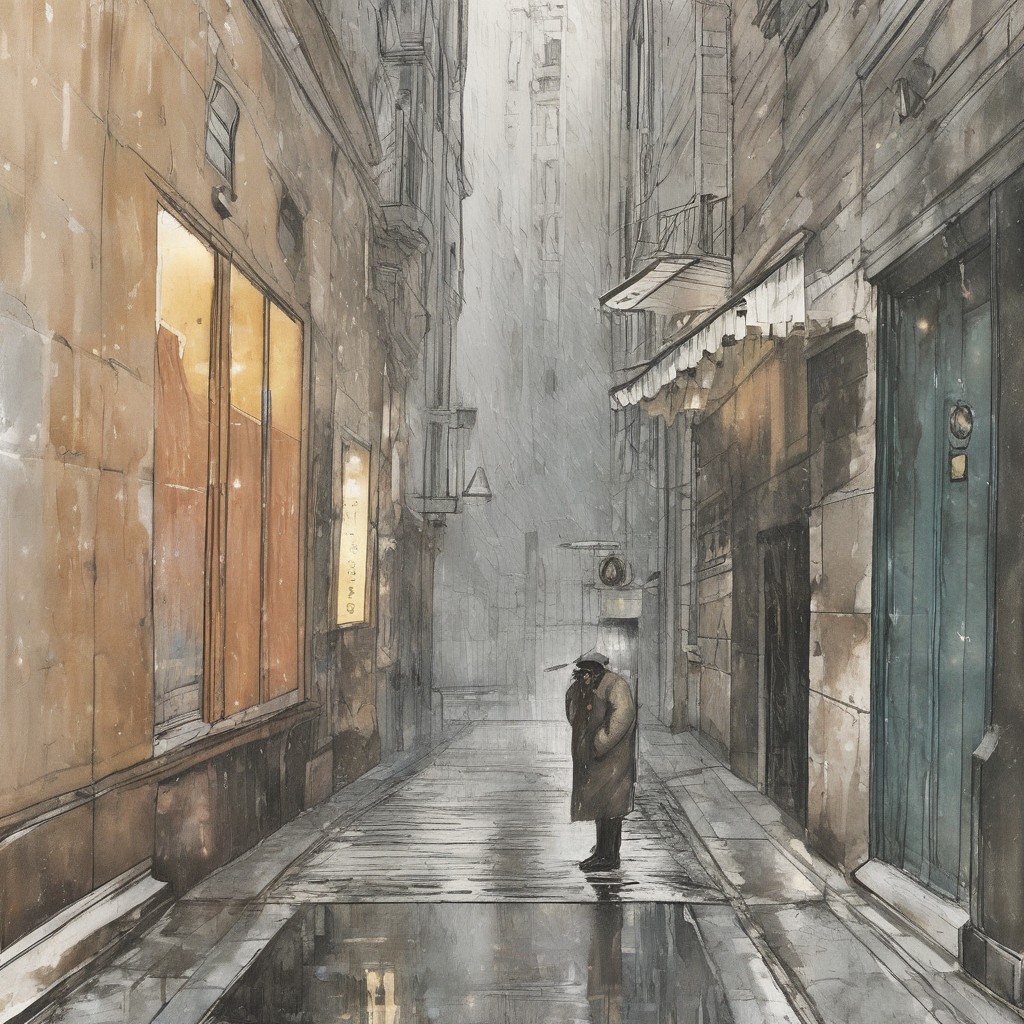} &
        \includegraphics[width=0.15\linewidth]{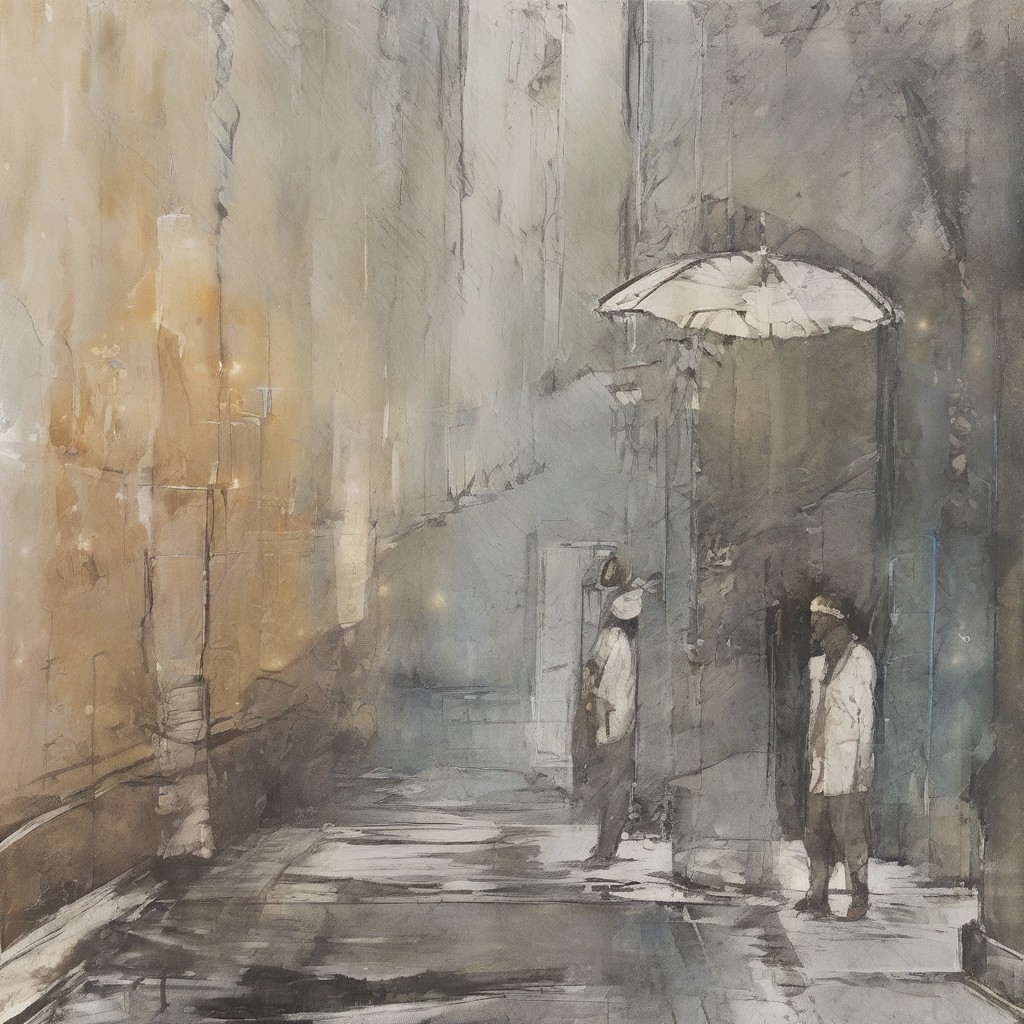} \\
        \includegraphics[width=0.15\linewidth]{img/images/beta/beta_distribution_single_2.0_2.0.jpg} &
        \includegraphics[width=0.15\linewidth]{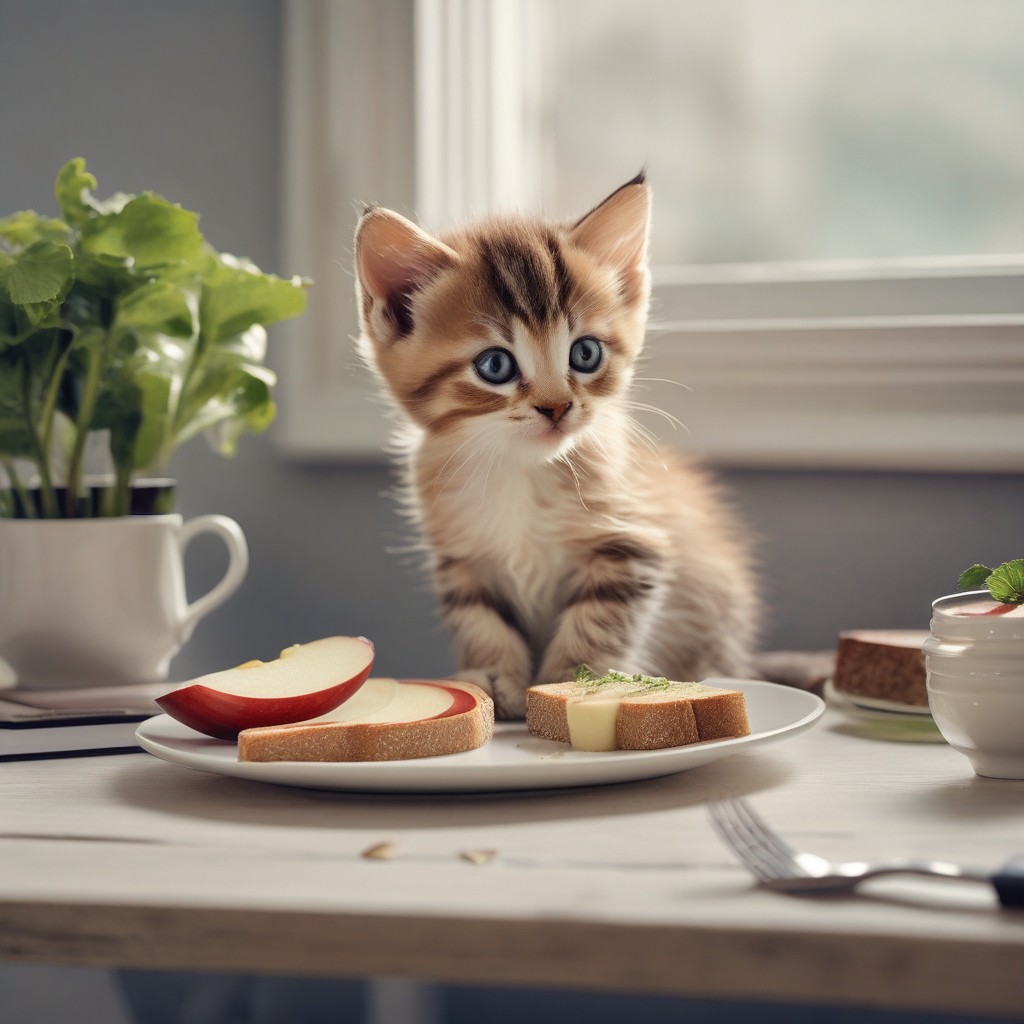} &
        \includegraphics[width=0.15\linewidth]{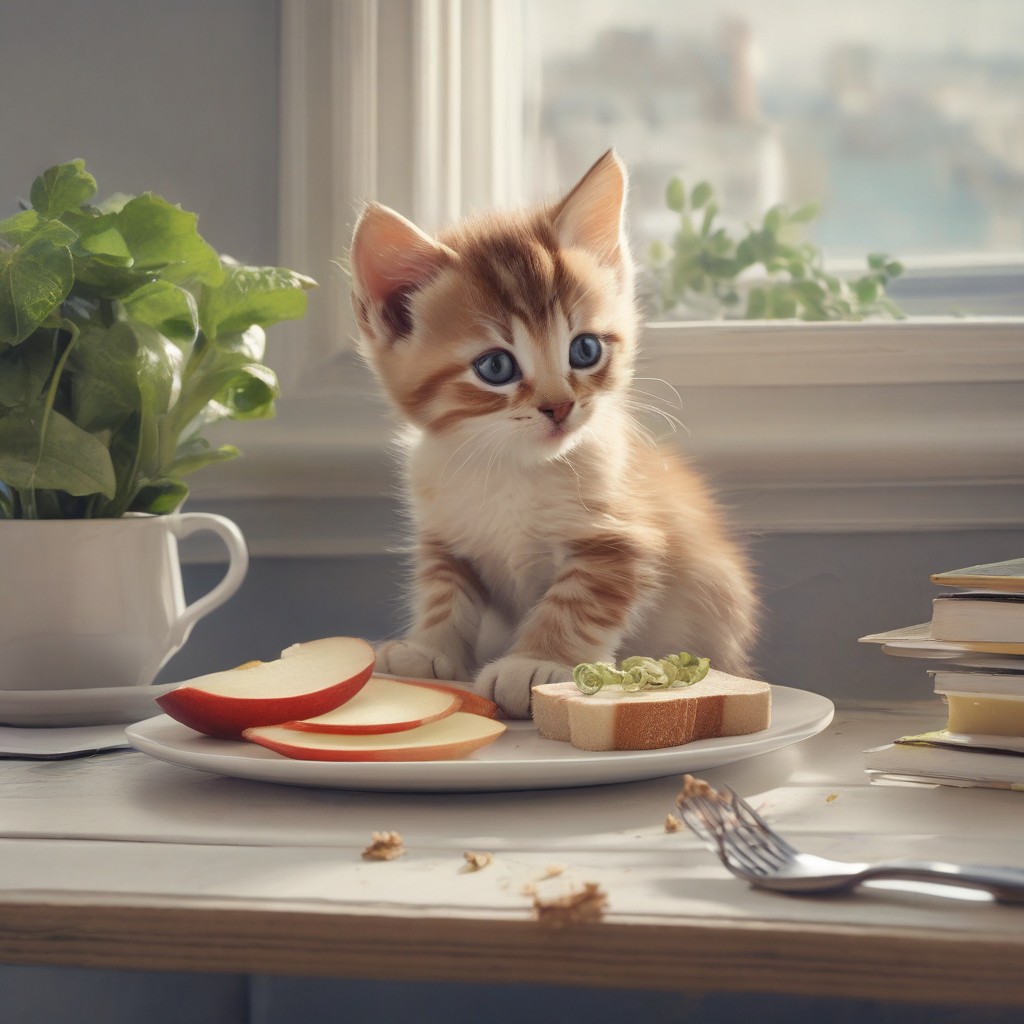} &
        \includegraphics[width=0.15\linewidth]{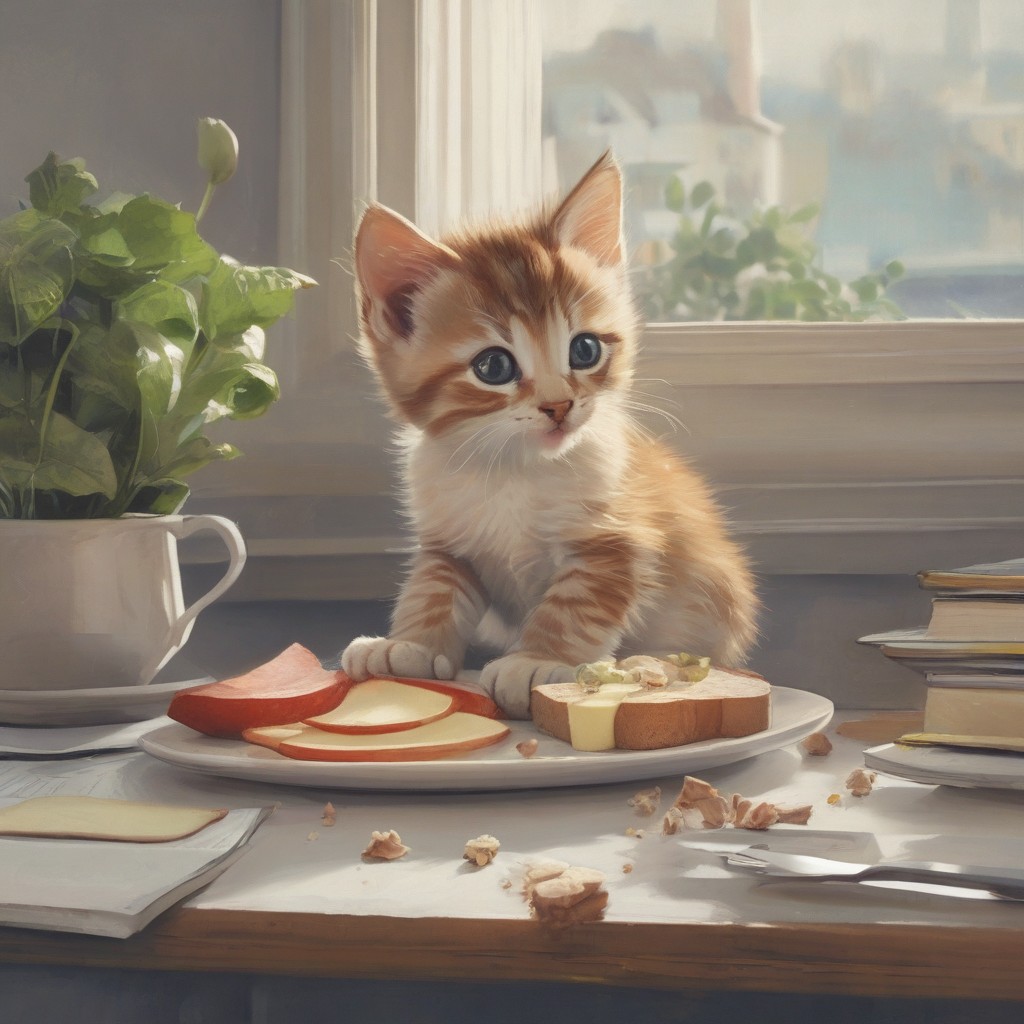} &
        \includegraphics[width=0.15\linewidth]{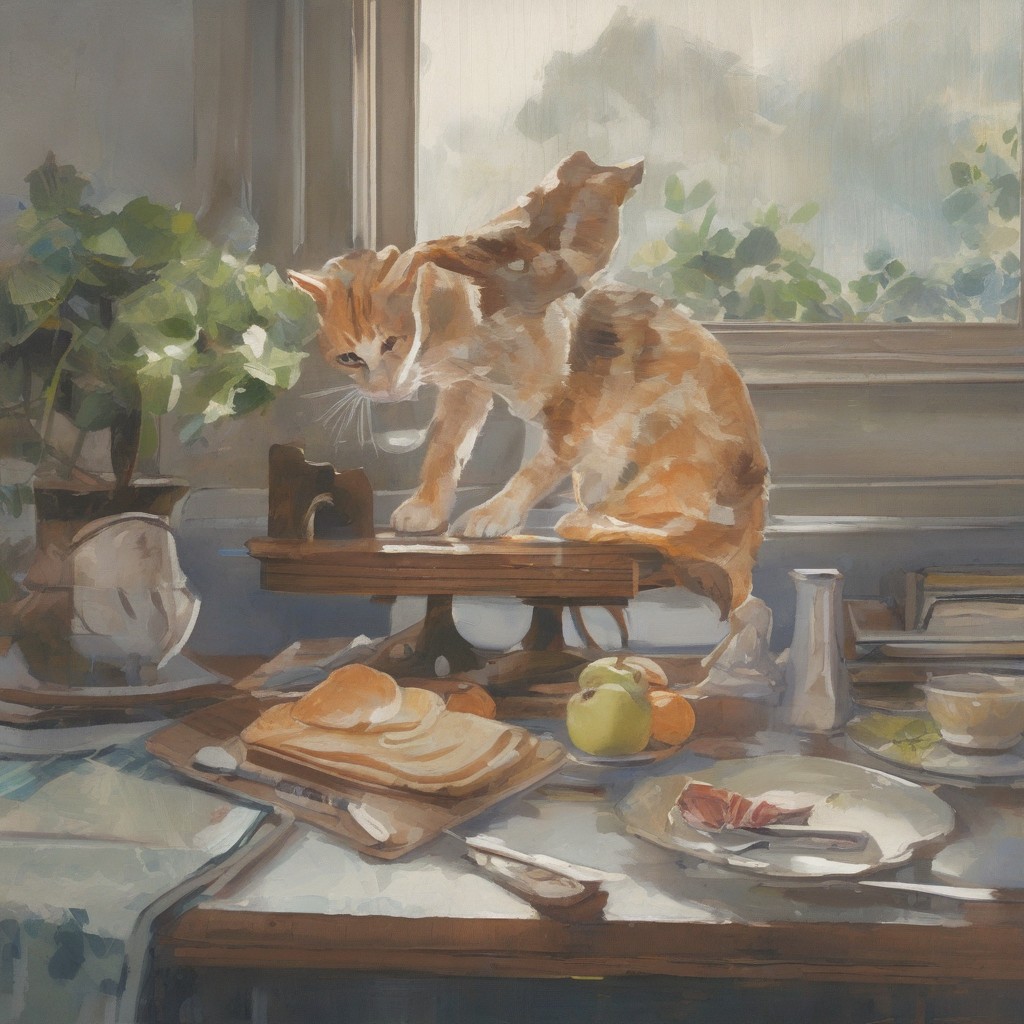} &
        \includegraphics[width=0.15\linewidth]{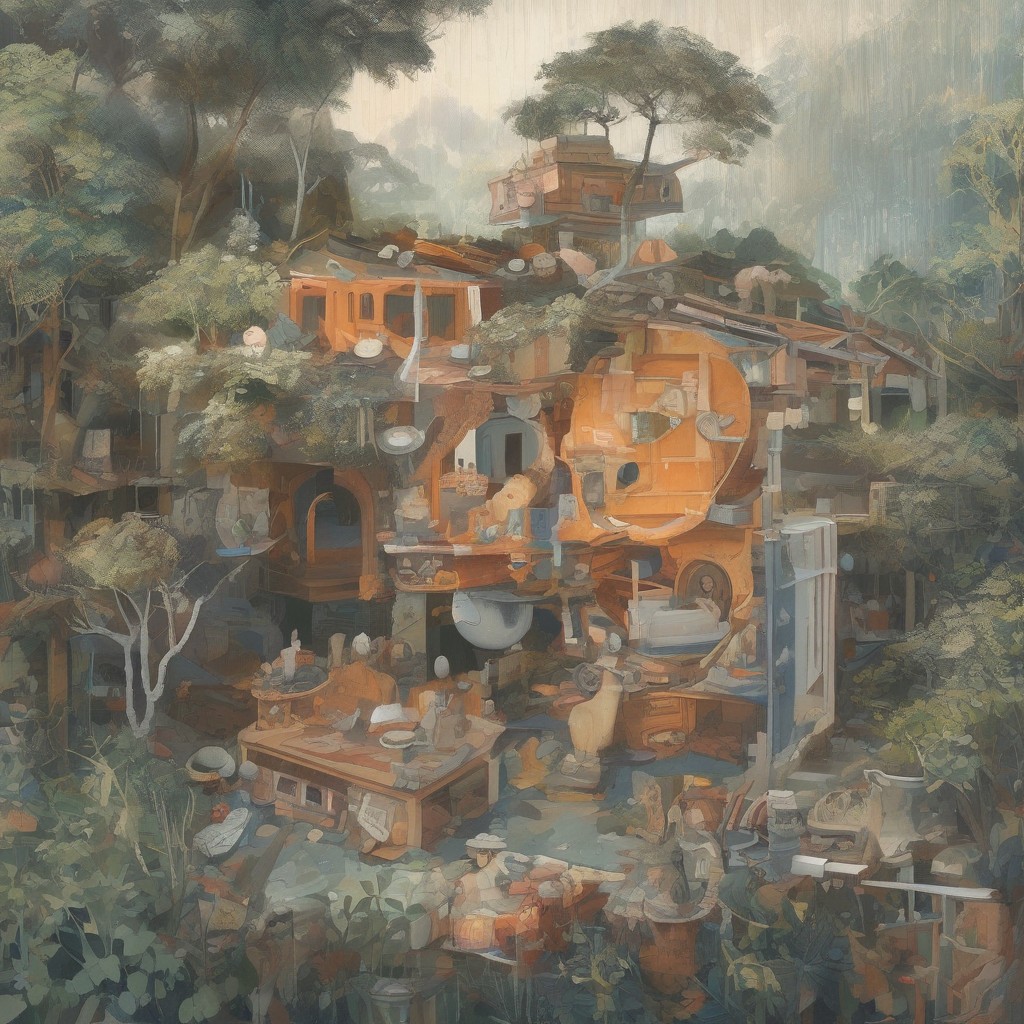} \\[0.2cm]
        \our{} & $\gamma = 0.0 $ & $\gamma = 0.25 $ & $\gamma = 0.5 $  & $\gamma = 1.0 $ & $\gamma = 2.0 $ \\
    \end{tabular}
    \caption{Example of sampled element according to $\gamma$ parameters. Prompts: "A boat is parked ashore without a passenger.", "A man sticking his head out of a doorway into a rainy city street.", "A kitten on a desk with an open sandwich and apple.".}
    \label{fig:sampling_gamma}
\end{figure*}

\begin{figure*}[!h]
    \centering
    \renewcommand{\arraystretch}{0}
    \setlength{\tabcolsep}{0pt}
\includegraphics[width=0.9\linewidth]{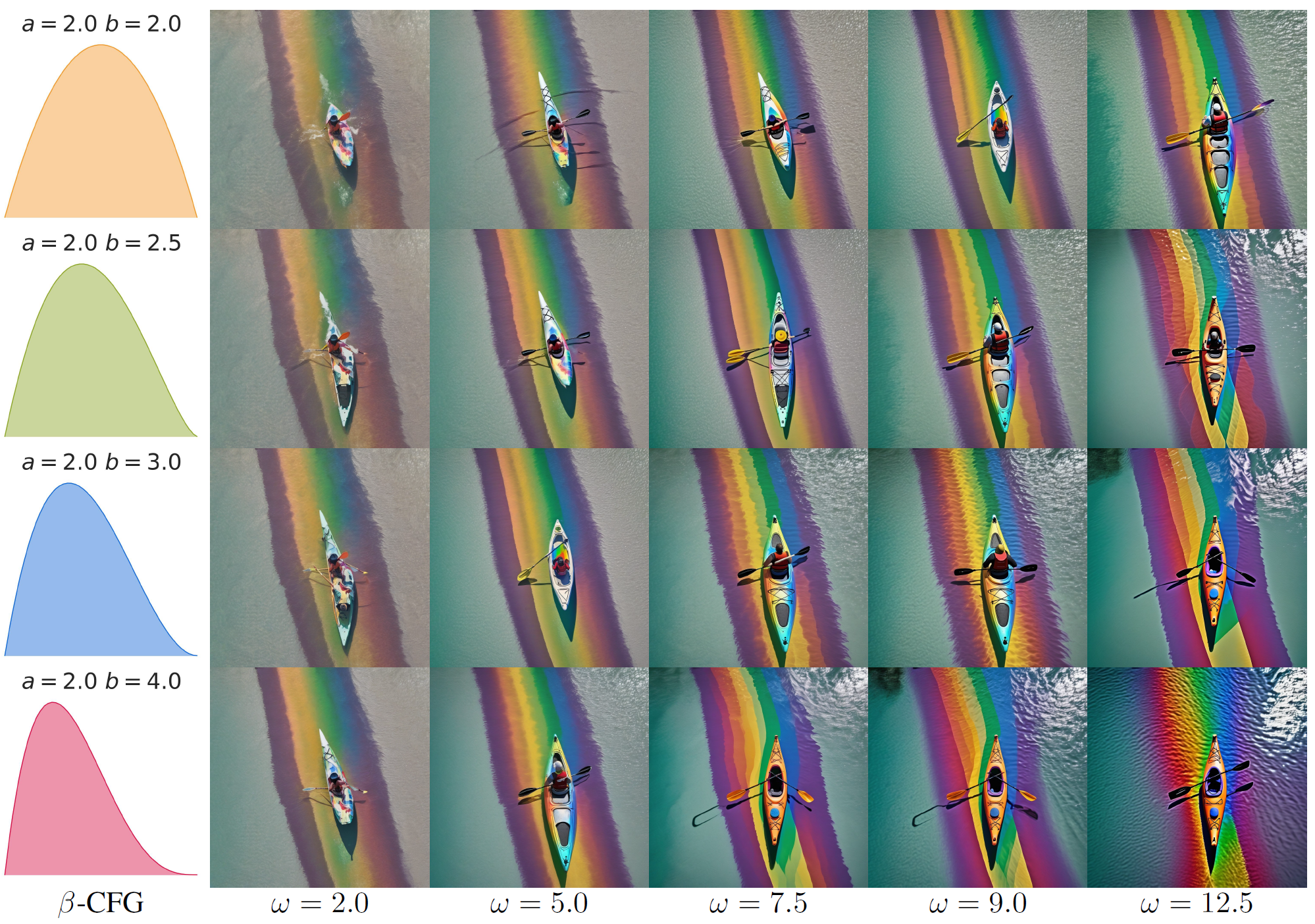}
    \vspace{-0.2cm}
    \caption{Prompt: "kayak in the water, optical color, aerial view, rainbow"}
    \label{fig:ap1}
\end{figure*}

\begin{figure*}[!h]
    \centering
    \renewcommand{\arraystretch}{0}
    \setlength{\tabcolsep}{0pt}
\includegraphics[width=0.9\linewidth]{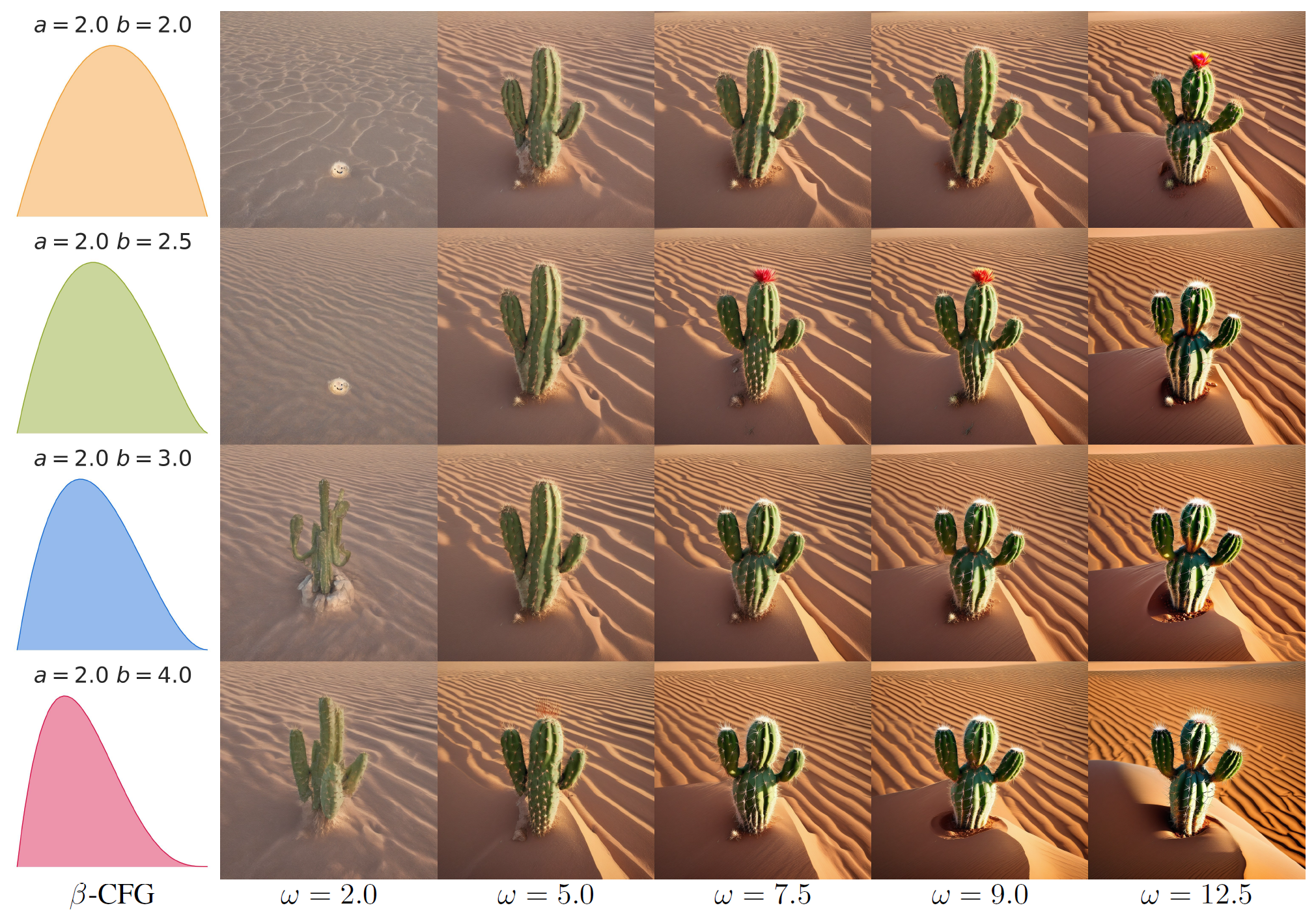}
    \vspace{-0.2cm}
    \caption{Prompt: "A small cactus with a happy face in the sahara desert"}
    \label{fig:ap2}
\end{figure*}

\begin{figure*}[!h]
    \centering
    \renewcommand{\arraystretch}{0}
    \setlength{\tabcolsep}{0pt}
\includegraphics[width=0.9\linewidth]{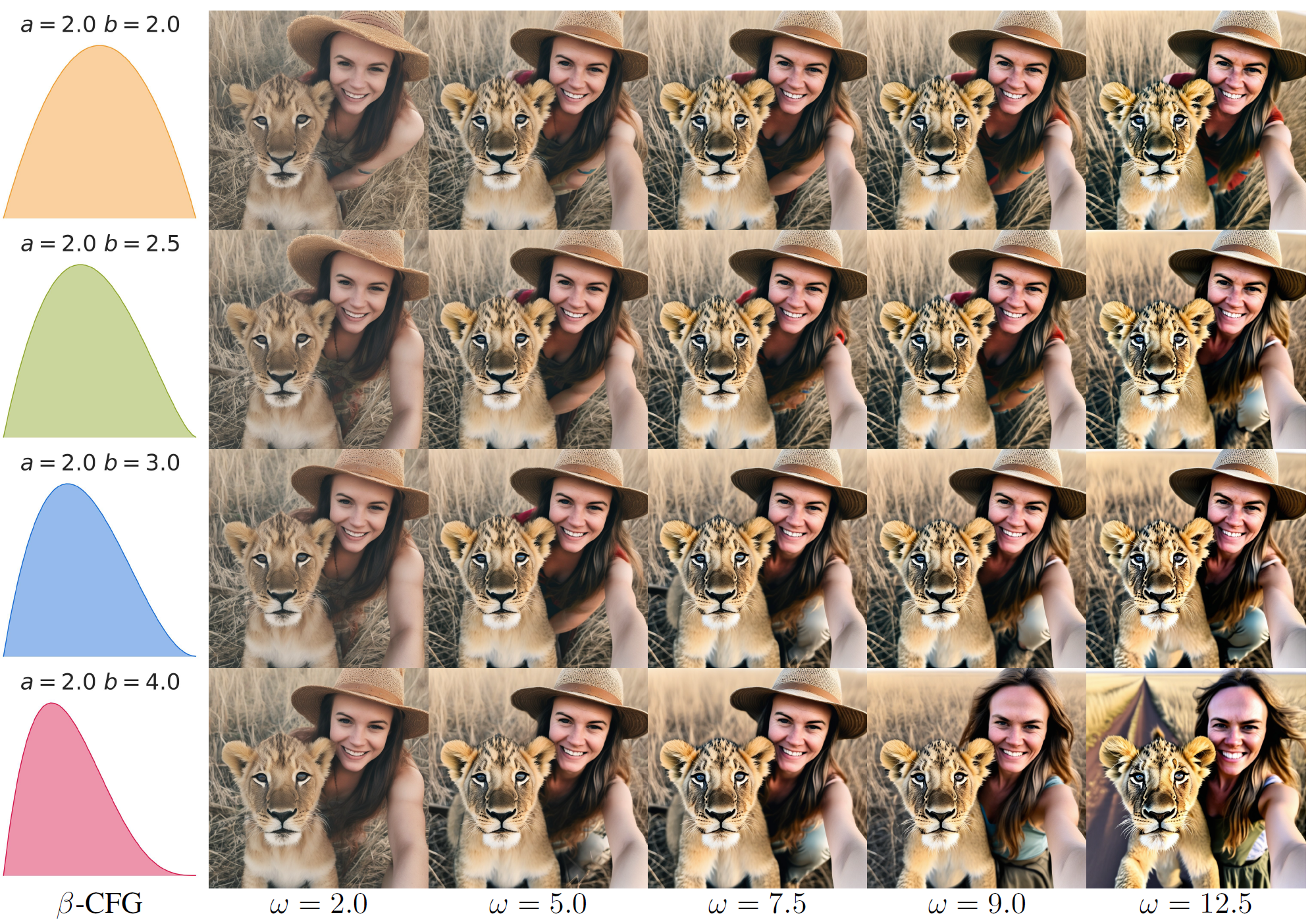}
    \vspace{-0.2cm}
    \caption{Prompt: "selfie of a woman and her lion cub on the plains"}
    \label{fig:ap3}
\end{figure*}

\begin{figure*}[!h]
    \centering
    \renewcommand{\arraystretch}{0}
    \setlength{\tabcolsep}{0pt}
\includegraphics[width=0.9\linewidth]{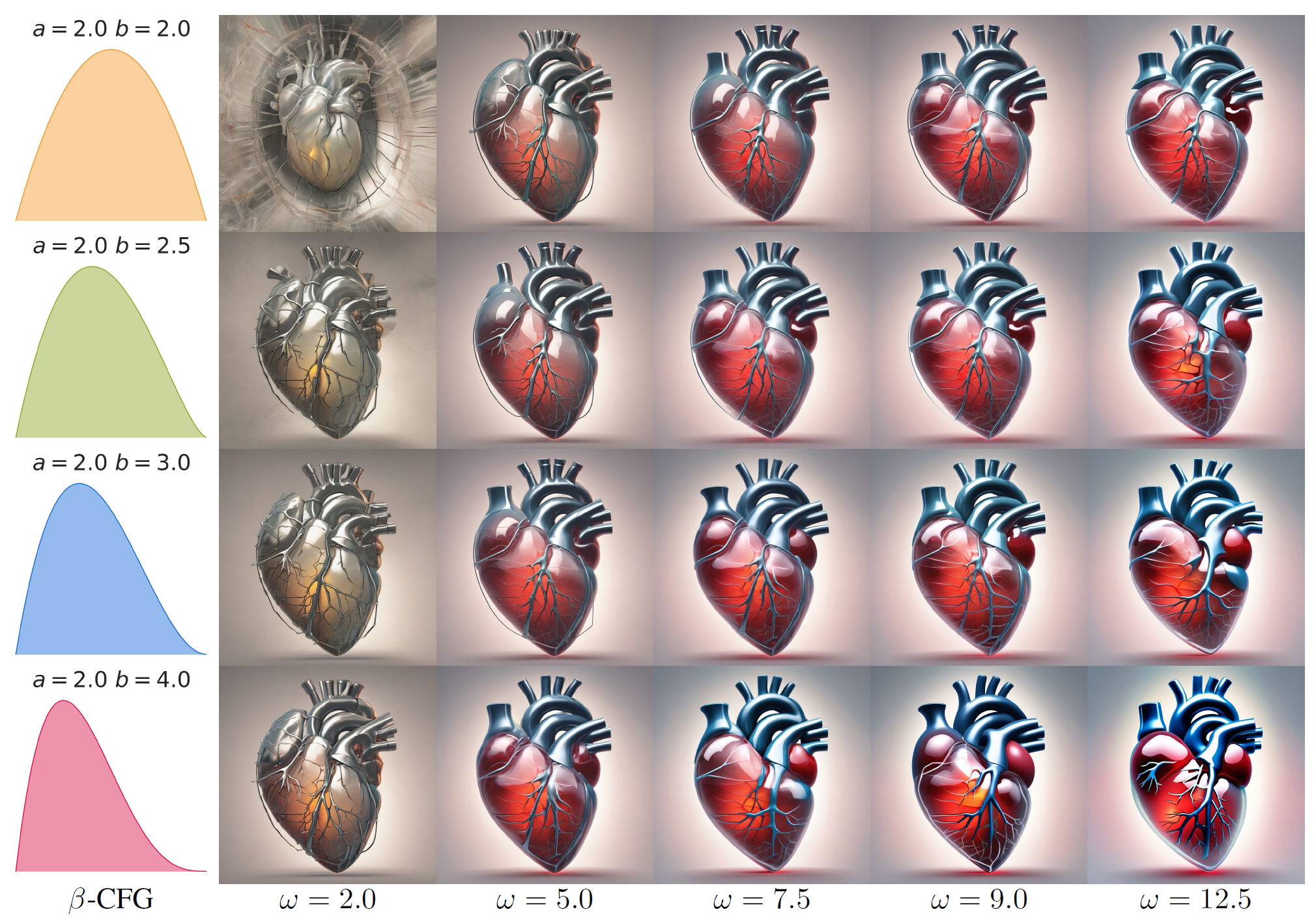}
    \vspace{-0.2cm}
    \caption{Prompt: "An illustration of a human heart made of translucent glass."}
    \label{fig:ap4}
\end{figure*}

This section presents an ablation study based on the parameters of the $\beta$-distribution. The quantitative results are listed in Tab.~\ref{tab:able} and are illustrated in Fig.~\ref{fig:ablation_cfg_beta}. Fig.~\ref{fig:ap1}, \ref{fig:ap2}, \ref{fig:ap3}, \ref{fig:ap4}.
Tab~\ref{tab:able2} displays the ablation studies' results on the $\gamma$ parameter. Figure~\ref{fig:sampling_gamma} provides a visual comparison of various $\gamma$ parameter values.

\begin{table*}[h!]
    \centering
    \begin{tabular}{ccccccccccc}
        \hline
        \multirow{2}{*}{Method} & \multicolumn{2}{c}{$\omega = 2.0, \lambda = 0.2$} & \multicolumn{2}{c}{$\omega = 5.0, \lambda = 0.4$} & \multicolumn{2}{c}{$\omega = 7.5, \lambda = 0.6$} & \multicolumn{2}{c}{$\omega = 9.0, \lambda = 0.8$} & \multicolumn{2}{c}{$\omega = 12.5, \lambda = 1.0$} \\
        & FID $\downarrow$ & CLIP $\uparrow$ & FID $\downarrow$ & CLIP $\uparrow$ & FID $\downarrow$ & CLIP $\uparrow$ & FID $\downarrow$ & CLIP $\uparrow$ & FID $\downarrow$ & CLIP $\uparrow$ \\
        \hline
        \our (2.0, 2.0)& 15.02 & 0.306
& 15.76 & 0.317
& 17.99 & 0.319
& 18.94 & 0.319
& 20.97 & 0.320
\\
\our (2.0, 2.5)& 14.18 & 0.308
& 16.80 & 0.318
& 18.99 & 0.319
& 20.24 & 0.319
& 22.24 & 0.320
\\
\our (2.0, 3.0)& 13.85 & 0.309
& 17.51 & 0.317
& 19.90 & 0.319
& 21.03 & 0.319
& 23.26 & 0.319
\\
\our (2.5, 2.0)& 17.26 & 0.303
& 15.18 & 0.316
& 16.73 & 0.318
& 17.71 & 0.319
& 19.46 & 0.320
\\
\our (2.5, 2.5)& 15.31 & 0.306
& 15.62 & 0.317
& 17.80 & 0.319
& 18.76 & 0.319
& 20.62 & 0.319
\\
\our (2.5, 3.0)& 14.58 & 0.307
& 16.36 & 0.317
& 18.63 & 0.319
& 19.67 & 0.319
& 21.84 & 0.319
\\
\our (3.0, 2.0)& 20.11 & 0.300
& 15.16 & 0.314
& 16.38 & 0.317
& 17.00 & 0.318
& 18.62 & 0.319
\\
\our (3.0, 2.5)& 17.35 & 0.303
& 15.39 & 0.316
& 16.89 & 0.318
& 17.89 & 0.318
& 19.73 & 0.319
\\
\our (3.0, 3.0)& 15.70 & 0.305
& 15.77 & 0.316
& 17.73 & 0.318
& 18.75 & 0.319
& 20.72 & 0.319
\\
        \hline
    \end{tabular}
    \caption{Ablation study of $\beta$-distribution parameters of T2I with SD v1.5}
    \label{tab:able}
\end{table*}

\begin{table*}[h!]
    \centering
    \begin{tabular}{ccccccccccc}
        \hline
        \multirow{2}{*}{Method} & \multicolumn{2}{c}{$\omega = 2.0, \lambda = 0.2$} & \multicolumn{2}{c}{$\omega = 5.0, \lambda = 0.4$} & \multicolumn{2}{c}{$\omega = 7.5, \lambda = 0.6$} & \multicolumn{2}{c}{$\omega = 9.0, \lambda = 0.8$} & \multicolumn{2}{c}{$\omega = 12.5, \lambda = 1.0$} \\
        & FID $\downarrow$ & CLIP $\uparrow$ & FID $\downarrow$ & CLIP $\uparrow$ & FID $\downarrow$ & CLIP $\uparrow$ & FID $\downarrow$ & CLIP $\uparrow$ & FID $\downarrow$ & CLIP $\uparrow$ \\
        \hline
        $\gamma= 0.0$ & 195.72 & 0.308
        & 185.55 & 0.318
        & 185.27 & 0.320
        & 185.09 & 0.320
        & 185.76 & 0.321 \\
        $\gamma= 0.25$
        & 201.32 & 0.303
        & 185.81 & 0.317
        & 185.05 & 0.319
        & 185.36 & 0.320
        & 185.15 & 0.320 \\
        $\gamma= 0.50$ 
        & 210.74 & 0.293
        & 186.57 & 0.316
        & 185.91 & 0.319
        & 184.97 & 0.319
        & 184.67 & 0.320 \\
        $\gamma= 0.75$ 
        & 229.62 & 0.272
        & 188.86 & 0.313
        & 186.19 & 0.318
        & 185.08 & 0.319
        & 184.37 & 0.320 \\
        $\gamma= 1.0$ 
        & 248.45 & 0.238
        & 192.94 & 0.309
        & 186.95 & 0.316
        & 185.12 & 0.318
        & 184.39 & 0.319
\\
        \hline
    \end{tabular}
    \caption{Ablation study of $\gamma$ of T2I with SD v1.5. The metrics were computed based on 1k prompts.}
    \label{tab:able2}
\end{table*}

\end{document}